\documentclass[Afour,sageh,times]{sagej}

\usepackage{moreverb,url}
\usepackage{enumitem}
\usepackage{amsmath,amssymb,amsfonts}
\usepackage{mathrsfs}
\usepackage{algorithmic}
\usepackage{algorithm}
\usepackage{array}
\usepackage[caption=false,font=normalsize,labelfont=sf,textfont=sf]{subfig}
\usepackage{textcomp}
\usepackage{stfloats}
\usepackage{url}
\usepackage{verbatim}
\usepackage{graphicx}
\usepackage{subfig}
\usepackage{bm}
\usepackage{amsthm}
\usepackage{appendix}
\usepackage{booktabs} 
\usepackage{multirow} 
\usepackage[colorlinks,bookmarksopen,bookmarksnumbered,citecolor=red,urlcolor=red]{hyperref}
\usepackage{tabularx}
\usepackage[table]{xcolor}
\usepackage{hyperref}

\newtheoremstyle{upright}
{3pt}
{3pt}
{\normalfont}
{}
{\bfseries}
{.}
{ }
{}

\newtheorem{theorem}{Theorem}
\newtheorem{lemma}{Lemma}

\newtheorem{remark}{Remark}
\newtheorem{assumption}{Assumption}
\newtheorem{proposition}{Proposition}

\newcommand\BibTeX{{\rmfamily B\kern-.05em \textsc{i\kern-.025em b}\kern-.08em
		T\kern-.1667em\lower.7ex\hbox{E}\kern-.125emX}}

\def\volumeyear{2016}
\begin{document}
	
	
	\title{VIP: Variation-based Iterative-learning Planning for Robotic Navigation}
	
	\author{Shuli Lv\affilnum{1}, Pengda Mao\affilnum{1}, Chen Min\affilnum{1}, Li Hong\affilnum{1}, Runxiao Liu\affilnum{1}, Shuai Wang\affilnum{2} and Quan Quan\affilnum{1,2}}
	
	\affiliation{\affilnum{1}School of Automation Science and Electrical Engineering, Beihang University, Beijing, P.R. China
		\\
		\affilnum{2}Tianmushan Laboratory, Beihang University, Hangzhou, P.R. China
	}
	
	\corrauth{Quan Quan, is with the School of Automation Science and Electrical Engineering, Beihang University, Beijing 100191, P.R.China, and also with the Tianmushan Laboratory, Beihang University,	Hangzhou 311115, P.R. China}
	
	\email{qq\_buaa@buaa.edu.cn}
	
	\begin{abstract}
		Over the past decade, autonomous robotic systems have been increasingly deployed in applications such as surveying, search and rescue, and last-mile delivery. These applications require robots to generate safe and efficient motion plans in large, complex, and obstacle-dense environments, often under limited onboard computing resources. However, conventional planning methods commonly rely on finite-dimensional trajectory parameterization or increasingly long prediction horizons, leading to rapidly growing computational costs, particularly in multi-robot scenarios. This paper presents a novel variation-based iterative-learning planning (VIP) framework for efficient motion planning of both single robots and robotic swarms. Instead of optimizing a large number of discrete trajectory variables, VIP directly updates the planning command as a continuous function in an infinite-dimensional function space. The same variation-based update can be implemented in a model-in-the-loop manner for offline planning or in a robot-in-the-loop manner between online physical executions. By avoiding the computational burden associated with horizon expansion and high-dimensional trajectory discretization, VIP maintains a per-iteration computational complexity of $\mathcal{O}(n)$, where $n$ denotes the number of spatial discretization points. Extensive simulations and real-world experiments demonstrate that the proposed framework can efficiently generate and iteratively improve motion plans for different planning objectives, robotic platforms, and swarm configurations, highlighting its effectiveness, computational efficiency, and scalability as a general planning methodology.
	\end{abstract}
	
	
	\maketitle
	
	\section{Introduction} \label{intro}
	In recent years, robots have been increasingly deployed across a wide range of applications, including delivery services, photography, and search-and-rescue operations (\cite{singamaneni2024survey}). As robotic systems move from structured settings to complex real-world environments, motion planning has become a fundamental capability for autonomous operation. A planner must generate safe, dynamically feasible, and efficient motion commands while accounting for environmental constraints. This requirement becomes particularly challenging with \emph{the limited onboard computing resources} of robotic platforms, where the number of planning variables and coupled constraints can grow rapidly. Therefore, developing a computationally efficient and scalable planning methodology for both single robots and robotic swarms is an important research problem (\cite{tordesillas2021faster}).
	
	\begin{figure*}
		\centering
		\includegraphics[width=6.8in]
		{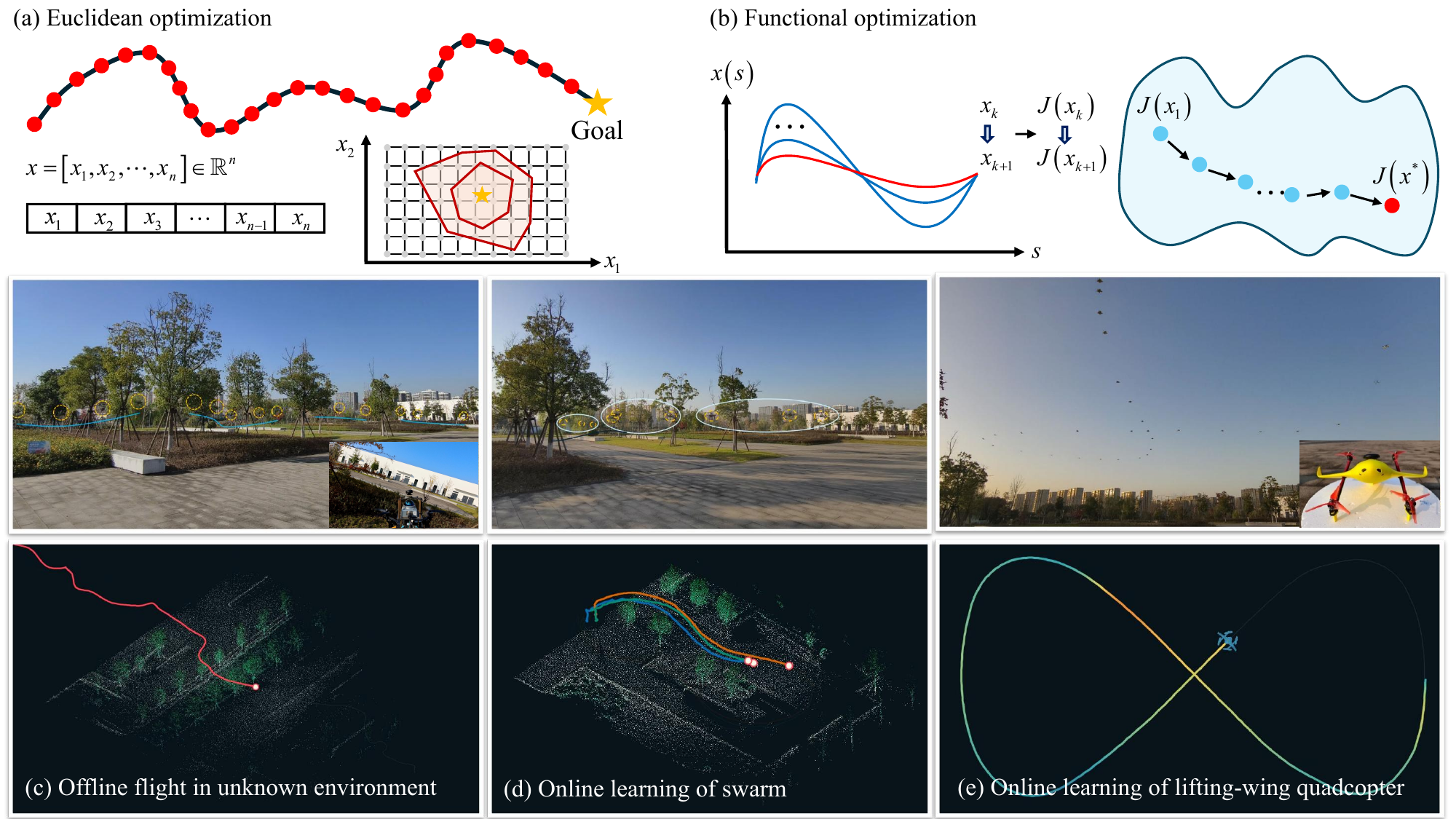}
		\caption{Overview of the proposed variation-based iterative-learning planning (VIP) framework and its real-world experimental validation. (a) Euclidean optimization, where a trajectory is parameterized by a finite set of waypoints. (b) Functional optimization adopted by VIP, where the trajectory is directly optimized in an infinite-dimensional function space. (c) Offline flight in an unknown environment. (d) Online learning of a robot swarm. (e) Online learning of a lifting-wing quadcopter.}
		\label{VIPframe}
	\end{figure*}
	
	\subsection{Motivations}
	
	To generate motion plans for a single robot, existing methods can be broadly categorized into \emph{prediction-based planning} and \emph{reactive-based planning} methods (\cite{rousseas2024reactive, florence2020integrated}). Prediction-based planners, such as model predictive control, formulate the planning problem as a multi-step optimization problem over a finite prediction horizon. The planned motion may be represented by discrete states, parameterized continuous trajectories such as polynomials or B-splines, or candidate trajectories generated through sampling and search. These formulations can explicitly incorporate system dynamics, motion constraints, and predicted environmental information. However, their computational cost generally increases with the prediction horizon, trajectory resolution, and number of decision variables. Reactive planners instead determine motion commands primarily from the current robot state and locally observed environment. Although they typically require less online computation, their planning performance may be sensitive to local environmental structures and limited look-ahead information. Consequently, obtaining high-quality motion plans with low computational cost remains difficult, especially in large and obstacle-dense environments.
	
	The computational challenge becomes more significant when extending motion planning from a single robot to robotic swarms. Multi-robot planning requires not only generating feasible motion commands for individual robots, but also accounting for inter-robot interactions, collision-avoidance constraints, and collective-motion objectives. Centralized planners generally optimize the motion of all robots in a joint state space, causing the problem dimension and computational burden to increase rapidly with the swarm size. Distributed planners alleviate centralized computation by solving local planning problems, but their performance may depend strongly on communication conditions, local coordination rules, and simplified assumptions regarding neighboring robots. Formation-based methods provide a structured way to coordinate a swarm, but fixed or weakly adaptive formations may restrict the feasible planning space and reduce the ability of the swarm to pass efficiently through cluttered environments. These limitations indicate that scalability should be addressed at the level of the planning formulation itself, rather than only through additional computational resources or distributed implementation.
	
	Beyond the limitations of individual planning algorithms, a more fundamental issue concerns how collective motion-planning problems are formulated. Existing multi-robot approaches largely follow a bottom-up philosophy, in which swarm motion emerges from predefined local coordination and control rules (\cite{crespi2008top}). Although this paradigm supports distributed implementation, it provides limited ability to directly optimize swarm-level planning objectives. Top-down approaches instead describe collective motion through concise macroscopic variables and models (\cite{Wiedemann2025physics}), providing a more systematic representation of swarm-level objectives and constraints. In particular, partial differential equation models offer a promising way to connect individual robot motion with macroscopic swarm evolution (\cite{sinigaglia2022density}). However, existing formulations are often not designed as general-purpose planning mechanisms and may still require computationally demanding numerical optimization. These observations motivate a unified planning methodology that can represent both individual and collective motion objectives, directly optimize planning commands, and retain low computational complexity as the planning dimension or swarm size increases.
	
	\subsection{Contributions}
	To address these planning challenges without relying on conventional computational scaling strategies, this paper proposes a variation-based iterative-learning planning (VIP) framework. 
	
	As shown in Fig. \ref{VIPframe}, the central philosophy of VIP is to directly optimize the planning command in an infinite-dimensional function space, rather than using a large number of point-wise decision variables in Euclidean space. From an energy-oriented perspective, different planning objectives are represented through corresponding system-energy functionals, and the planning command is iteratively improved according to the variation of the energy response. This formulation converts motion planning into a function-level iterative optimization process and avoids repeatedly solving a high-dimensional trajectory-optimization problem. To further remove the need for computationally expensive analytical gradients and accurate dynamic models, a model-free implementation is developed in which the planning command is updated using energy variations recorded from previous rollouts. Consequently, VIP provides a unified formulation for offline planning, execution-based online planning, and iterative replanning.
	
	Building upon our previous studies on time-optimal spatial iterative learning in vector fields (\cite{lv2025high}), swarm motion planning (\cite{lv2024mean}), and single-robot planning (\cite{Lv2026time}), this work establishes a VIP framework that systematically connects these previously separate formulations. Unlike our earlier studies, which focused on specific planning scenarios or individual optimization objectives, the present framework formulates different single-robot and swarm planning problems from the common perspective of energy variation. It therefore extends variation-based iterative learning from task-specific solutions to a general planning methodology that can be adapted to different systems, performance objectives, motion representations, and collective-planning settings.
	
	The main contributions are summarized as follows.
	
	(i) \textbf{\emph{A unified energy-variation planning formulation}}: A unified planning framework is established by representing different performance objectives through corresponding system-energy functionals. Rather than being restricted to a specific trajectory representation, robotic platform, or time-optimal objective, VIP provides a common formulation in which planning objectives can be incorporated through the definition of the energy functional and its variation. The same planning principle applies to both single-robot and multi-robot systems, enabling different motion-planning tasks to be addressed within a consistent mathematical framework.
	
	(ii) \textbf{\emph{Low computational complexity through functional optimization}}: VIP directly updates the planning command in an infinite-dimensional function space, avoiding the need to construct and repeatedly solve a high-dimensional trajectory-optimization problem. The variation-based update is evaluated along the planned motion with a computational complexity that grows linearly with the number of spatial discretization points. For the model-free implementation, the overall complexity after $k^*$ learning iterations is $O(k^*n)$, where $n$ is the number of discretization points. As conceptually illustrated in Fig.~\ref{VIPframe}, this function-level optimization mechanism limits the growth of decision variables and coupled constraints associated with trajectory refinement, prediction-horizon expansion, and multi-robot planning, making VIP suitable for computationally constrained onboard platforms.
	
	(iii) \textbf{\emph{Online and model-free planning based on execution history}}: A model-free planning mechanism is developed to update the planning command directly from energy variations recorded during robot-in-the-loop execution, without requiring an explicit dynamic model or computationally expensive analytical gradient evaluation. Moreover, the same variation-based update can be implemented in a model-in-the-loop mode using simulated rollouts for offline planning. By incorporating the actual closed-loop response in planning layer, the resulting plan reflects practical tracking behavior, dynamic limitations, disturbances, and modeling uncertainties, rather than relying solely on geometric paths or nominal system predictions. This execution-aware formulation extends iterative learning from conventional trajectory-tracking control to planning-command optimization and enables motion plans to be continuously improved through repeated execution. Extensive simulations and real-world experiments validate the planning effectiveness, computational efficiency, and scalability of the proposed VIP framework.

	\subsection{Paper organization}
	The remainder of this paper is organized as follows. \emph{Section \ref{relatedwork}} reviews single-robot navigation, swarm navigation, and iterative learning. \emph{Section \ref{preliminary}} introduces the geometric and spatial-domain preliminaries. Following a concrete-to-abstract order, \emph{Section \ref{problem}} first formulates the single-robot and robotic-swarm systems, then derives their common energy systems and formulates the unified optimization problem. \emph{Section \ref{IL}} develops the model-based and model-free VIP laws directly for this unified system and analyzes convergence, computational complexity, and implementation modes. \emph{Sections \ref{singlerealization} and \ref{swarmrealization}} provide task-specific realization and verification details without repeating the unified derivation. \emph{Sections \ref{simcom} and \ref{exp}} report simulations and real-world experiments. \emph{Section \ref{conclusion}} concludes the paper.
	
	\subsection{Notations}
	Unless otherwise specified, all vectors are column vectors. $\nabla_{\mathbf{z}} \triangleq \partial / \partial \mathbf{z}$ and $\Vert\mathbf{z}\Vert \triangleq \sqrt{\mathbf{z}^{\mathrm{T}}\mathbf{z}}$. The operators $\mathrm{D}_{z}$ and $\partial_z$ denote the total and partial derivatives with respect to $z$, respectively, and $\delta_z$ denotes a functional variation. For a spatial or temporal domain $\mathcal{D}$, the Hilbert space $\mathcal{L}^2(\mathcal{D})$ is equipped with the norm $\Vert f\Vert_{\mathcal{L}^2(\mathcal{D})}=(\int_{\mathcal{D}} |f|^2\mathrm{d}\zeta)^{1/2}$ and inner product $\langle f,g\rangle_{\mathcal{D}}=\int_{\mathcal{D}} fg\,\mathrm{d}\zeta$. In particular, $\mathcal{D}=(0,L)$ for spatial learning profiles and $\mathcal{D}=\mathcal{T}$ for swarm-density fields.

	\section{Related work}  \label{relatedwork}
	In this section, an overview of related studies on topics is provided, as shown in Tab. \ref{tab:method_comparison}, including single robot navigation, robotic swarm navigation, and iterative learning.
	
	\begin{table*}[t]
		\centering
		\caption{Comparison of planning paradigms for single robots and robotic swarms.}
		\label{tab:method_comparison}
		\renewcommand{\arraystretch}{1.15}
		\setlength{\tabcolsep}{5pt}
		\footnotesize
		
		\begin{tabularx}{\textwidth}{
				>{\raggedright\arraybackslash}p{2.2cm}
				>{\raggedright\arraybackslash}p{2.3cm}
				>{\raggedright\arraybackslash}X
				>{\raggedright\arraybackslash}X}
			
			\toprule
			\textbf{Paradigm}
			&
			\textbf{Subclass}
			&
			\textbf{Single robot}
			&
			\textbf{Robotic swarm}
			\\
			\midrule
			
			\multirow{3}{=}{Prediction-based planning}
			&
			Optimization-based
			&
			\textbf{Explicit optimization}; constraint handling; model dependence; high computation
			&
			Coordinated trajectories; collision avoidance; increasing \textbf{computation and communication}
			\\
			
			&
			Parameterized trajectory
			&
			Reduced variables; smooth trajectories; complex time optimization
			&
			Individual trajectory parameterization; demanding \textbf{joint optimization}
			\\
			
			&
			Sampling-based
			&
			Nonlinear objectives; flexible search; sampling dependence
			&
			Flexible motion generation; conflict resolution; limited scalability
			\\
			
			\midrule
			
			\multirow{2}{=}{Reactive-based planning}
			&
			Vector/potential field
			&
			\textbf{Fast response}; low computation; convergence-oriented
			&
			Local interactions; distributed coordination; limited global optimization
			\\
			
			&
			CBF/flocking
			&
			Safety constraints; reactive collision avoidance; online optimization
			&
			Collision avoidance; connectivity; cooperation; dense-swarm complexity
			\\
			
			\midrule
			
			\multirow{3}{=}{Iterative learning}
			&
			Model-based ILC
			&
			Function-space update; model information; \textbf{tracking-level learning}
			&
			Repetitive collective tracking; predefined references; model dependence
			\\
			
			&
			Model-free ILC
			&
			Execution-data reuse; model independence; \textbf{tracking-level learning}
			&
			Data-driven collective tracking; predefined collective motion
			\\
			
			&
			\textbf{Proposed VIP}
			&
			\textbf{Planning-level learning};
			\textbf{low complexity}; \textbf{online and model-free}
			&
			\textbf{Macroscopic learning};
			microscopic coordination
			\\
			
			\bottomrule
		\end{tabularx}
	\end{table*}

	\begin{table}[t]
		\centering
		\caption{Bottom-up and top-down swarm navigation paradigms.}
		\label{tab:swarm_paradigms}
		\renewcommand{\arraystretch}{1.12}
		\setlength{\tabcolsep}{3.5pt}
		\scriptsize
		
		\begin{tabularx}{\columnwidth}{
				>{\raggedright\arraybackslash}p{1.25cm}
				>{\raggedright\arraybackslash}X
				>{\raggedright\arraybackslash}X}
			
			\toprule
			\textbf{Paradigm}
			&
			\textbf{Principle}
			&
			\textbf{Main trade-off}
			\\
			\midrule
			
			Bottom-up
			&
			Generates collective motion through individual planning, control, or local interactions
			&
			Explicit individual coordination, but increasing computation and communication
			\\
			
			Top-down
			&
			Describes the swarm using shared regions, macroscopic variables, or density evolution
			&
			High scalability, but limited direct control of individual motion performance
			\\
			
			\bottomrule
		\end{tabularx}
	\end{table}

	\subsection{Single robot motion planning}
	In the field of robotic navigation, various methodologies have been developed to optimize motion primitives. Motion planning generally entails solving an optimal control problem to minimize a predefined cost function subject to certain constraints. Prediction-based planning involves identifying the optimal sequence of control commands at each sampling interval. A prevalent optimization-based method, MPC, addressing trajectory planning by utilizing precise model information, is often a choice for aggressive maneuvers (\cite{nguyen2021model}), such as nonlinear MPC (NMPC) (\cite{sun2022comparative}), model predictive contouring control (MPCC) (\cite{liniger2015optimization}), and perception-aware MPC (PAMPC) (\cite{falanga2018pampc}). However, achieving superior outcomes often requires accurate modeling, and optimizing nonlinear models also imposes a significant computational load (\cite{song2023reaching}). As a result, it is typically challenging to apply these methods in real time for longer-horizon optimization.
	
	To improve optimization efficiency and prediction horizon, polynomial (\cite{richter2016polynomial, tordesillas2021faster}) and B-spline (\cite{zhou2020ego}) representations are employed, leveraging differential flatness (\cite{mueller2015computationally}). These methods transform the optimization variable from discrete control inputs to a parameterized trajectory. The widely recognized minimum snap (\cite{mellinger2011minimum, mellinger2012trajectory}) method is known to reduce the aggressiveness of control inputs (\cite{foehn2021time}). Nevertheless, the transformation to minimize flight time, as discussed in certain studies, introduces complexity to the objective function and may lead to deviations from the original constraints (\cite{han2021fast}). Consequently, the task is often formulated as a constrained optimization problem, whose conversion into an unconstrained form may increase complexity and potentially produce solutions that violate the original constraints \cite{icsleyen2022low}.
	
	Given the considerable computational burden, many methods alleviate the challenge by limiting the solution space or approximating the original problem (\cite{karaman2011sampling, webb2013kinodynamic}). For example, sampling-based planning (\cite{orthey2023sampling}), such as the dynamic window approach (DWA) (\cite{fox1997dynamic}), enhances computational efficiency by sampling and evaluating simulated trajectories within a velocity search space. Model predictive path integral (MPPI) control samples a batch of forward rollouts and updates the control sequence through cost-weighted averaging, making it applicable to nonlinear dynamics and non-smooth objectives (\cite{williams2017model}). Along this line, sampling-based model predictive control has become an important tool for real-time motion generation. To ensure smooth control inputs and prevent deviations from the planned path, sampling-based planning is frequently combined with state-to-state trajectories, such as polynomials, representing a common strategy.
	
	In the above prediction-based methods, adding extra waypoints (higher resolution) typically introduces more constraints or an expanded search space, increasing computational demands and making solutions harder to find. An alternative approach is provided by reactive planning (\cite{rousseas2024reactive, florence2020integrated}). As a form of reactive planning, VF-based planning emerges as a distinctive methodology (\cite{goncalves2010vector}). VF-based planning retains the benefits of low computational burden, ensuring optimal use of environmental and search information for safe navigation (\cite{rubi2020survey}). However, traditional VF methods primarily emphasize path convergence rather than rapid, efficient performance. The control barrier function (CBF) method (\cite{ames2016control, ames2019control}) combines the VF-based and optimization-based methods, avoids collisions by introducing state constraints, and is more suitable for precise control and scenarios with complex constraints. The CBF is utilized to simultaneously achieve control objectives, collision avoidance, and connectivity maintenance; however, it requires more computational resources.
	
	\subsection{Robotic swarm motion planning}
	From a single robot to a swarm, prediction-based methods generate collision-free trajectories with higher-order continuity for each robot, either in a centralized or distributed fashion, and then guide each robot to follow its corresponding trajectory (\cite{toumieh2022decentralized, park2025decentralized}). To reach a specified target point, each robot in the swarm system should first find a discrete geometric path in the global map before locally optimizing the path to produce a feasible trajectory that avoids obstacles and other robots’ trajectories (\cite{zhou2022swarm, tordesillas2021mader}). If multi-robot trajectory planning is centralized, a designated central node computes trajectories for all robots and then broadcasts them via wireless communication; the maximum number of robots that can be handled is constrained by the central node's computational power. To reduce this burden, an optimal virtual tube (\cite{mao2023optimal}) is proposed to guide the swarm within a shared collision-free tube, but scalability is still limited by communication and coordination overhead. In contrast, in a distributed  fashion, each robot shares its planned trajectory with its neighbors via wireless communication, and the maximum number of robots that can be accommodated is limited by their communication capabilities (\cite{tordesillas2021mader}).
	
	Reactive-based methods are well-suited for managing a large robotic swarm. These methods typically employ straightforward controllers that react promptly to obstacles or other robots, enabling fast, responsive operation in dynamic environments with low computational and communication resource requirements. From a single robot to a swarm, these methods need to ensure the swarm can avoid collisions and cooperate. Popular examples include the potential field method (\cite{khatib1986real, gonccalves2020stable}) and flocking-based method (\cite{vasarhelyi2018optimized}). For a robotic swarm, CBF is used to simultaneously achieve control objectives, collision avoidance among robots in the swarm and with obstacles, and connectivity maintenance (\cite{glotfelter2017nonsmooth}), while ensuring that the feasible solution set for the QP problem is nonempty.
	The research area of multi-robot systems places considerable emphasis on formation control (\cite{alonso2017multi}). The leader-follower approach is the most common method in formation control, including the displacement-based method (\cite{chen2020distributed}), distance-based method (\cite{bae2020distributed}), and bearing-based method (\cite{zhao2019bearing}). However, scalability, adaptability, and efficiency present significant challenges. For instance, when the number of robots is scaled up, the physical size of the formation becomes too large for practical use.
	
	To enable scalable deployment and coordinated global configuration of large robotic swarms, as shown in Tab. \ref{tab:swarm_paradigms}, a top-down design paradigm has attracted increasing attention. Rather than explicitly planning or controlling the motion of individual robots, these approaches describe collective behaviors at the macroscopic level using partial differential equations (PDEs) (\cite{krishnan2018distributed, sinigaglia2022density}). In particular, mean-field PDE models provide a systematic framework for characterizing the evolution of swarm density and establishing a connection between individual robot dynamics and emergent collective behavior (\cite{elamvazhuthi2018bilinear}). By formulating coordination objectives in terms of spatial density distributions, PDE-based methods can reduce dependence on swarm size, thereby improving scalability for large-scale deployment.
	
	\subsection{Iterative learning}
	Iterative learning control (ILC) is a control methodology designed to improve the performance of systems that repetitively execute a task over multiple iterations (\cite{bristow2006survey, sun2002iterative}). ILC provides an infinite-dimensional optimization perspective from a functional standpoint, allowing iterative solutions to be computed over the entire navigation space rather than optimizing in a discrete or local space. A number of ILC design methods have been proposed and can generally be divided into two categories: model-based ILC design and model-free ILC design, depending on whether information on system dynamics is used. Model-based ILC design uses an explicit (though not necessarily accurate) system model to design the input updating law (\cite{gao2026robust}); examples include gradient-based ILC (\cite{owens2016norm}), norm-optimal ILC (\cite{ratcliffe2006norm}), predictive-optimal ILC (\cite{amann1998predictive}), and so on. On the other hand, model-free design directly updates the system input without using model information (\cite{janssens2012data}). Data from previous iterations is exploited, and the gained experience is used to improve the performance of subsequent executions by updating the feedforward input signal (\cite{xu2011survey}). With this, the algorithm compensates for unmodeled system dynamics, repetitive noise, or parametric uncertainties (\cite{zhang2022model}). 
	
	Despite these advantages, directly applying conventional ILC to robotic navigation is not straightforward. Classical ILC typically assumes a repetitive task with a given reference, whereas navigation requires the robot to determine its motion under environmental constraints (\cite{zhang2026user}). Moreover, most ILC methods focus on reducing tracking errors rather than optimizing planning-level objectives such as traversal time or motion efficiency, and when applied to navigation, they usually rely on a preplanned trajectory that is subsequently tracked by the controller (\cite{purwin2011performing,Zhong2026initial}); in this case, the improvement mainly occurs at the execution layer, while the traversal-time optimization remains separated from the learning process. These challenges motivate the proposed VIP framework, which extends iterative learning from trajectory tracking to planning-oriented navigation by updating the planning command while maintaining computational efficiency.

	\section{Preliminaries}  \label{preliminary}
	\subsection{Definitions in robotic navigation} \label{secDef}
	In the field of mobile robotics, various approaches have been proposed for planning feasible paths and optimizing motion primitives. The classic navigation framework for the robot typically comprises three components: perception, planning, and control.
	
	From the perception layer, the space occupied by obstacles in the environment is obtained, and defined as $\Omega \subset \mathbb{R}^3$. The traversable area is defined as	$\mathcal{C} = \mathcal{S} - \Omega$, $\mathcal{C}\subset \mathbb{R}^3$, which indicates the difference set from the whole space $\mathcal{S}\subset \mathbb{R}^3$ to $\Omega \subset \mathbb{R}^3$. In the navigation mission, the robot needs to move from the starting point $\mathbf{b}_\mathrm{p}\in \mathbb{R}^3$ to the ending point $\mathbf{g}_\mathrm{p}\in \mathbb{R}^3$ without collision as soon as possible.
	Considering the obstacle-free configuration space $\mathcal{C} \subset \mathbb{R}^3$,
	with a path planner, a feasible smooth path $\mathcal{V} \subset \mathcal{C}$ from start to end is obtained. Let $\mathbf{\gamma}\left(s\right) : \mathbb{R}_{\geq 0} \rightarrow \mathbb{R}^3$ be a parametric representation for $\mathcal{V}$, with $s \in \left[0,L\right]$ being the arc length, and $L$ denoting the path length. 
	
	For a robot at any position $\mathbf{p}$, as shown in Fig. \ref{Vtc}, the parameters of the closest points
	on the curve is defined as $l \left(\mathbf{p}\right): \mathbb{R}^3\rightarrow  \mathbb{R}_{\geq 0}$, where the multivalued $l$ is discussed in \cite{rezende2021constructive}. The closest point on $\mathbf{\gamma}$ is defined through the mapping $\mathbf{m}: \mathbf{\mathbf{p}} \in \mathbb{R}^3 \rightarrow \mathbf{\gamma} \left(l\right)$, that
	\begin{equation}
		\begin{split}
			l\left(\mathbf{p}\right) &= \mathop{\arg \min}\limits_{s\in \left[0,L\right]} \Vert \mathbf{p}- \mathbf{\gamma}\left(s\right) \Vert^2 \\
			\mathbf{m}\left(\mathbf{p}\right) &= \mathbf{\gamma} \left(l\right),
		\end{split}
		\label{ilmap}
	\end{equation}
	where the corresponding arc length of $\mathbf{\gamma}\left(l\right)$ from $\mathbf{\gamma}\left(0\right)$ to $\mathbf{\gamma}\left(l\right)$ is denoted as $l$. This mapping gives the robot a way to index the spatial coordinate $l$ along the path. 
	The tangent vector of the closest point on the path is defined through
	\begin{equation}
		\mathbf{t}_{\mathrm{c}}\left(l\right)  = \mathrm{D}_l \mathbf{\gamma}\left(l\right),
	\end{equation}
	which indicates the unit tangent direction (\cite{rezende2021constructive}) at the location $l$ along $\mathbf{\gamma}$. Then the tangent direction of the robot is defined as
	\begin{equation}
		\mathbf{t}_{\mathrm{c}}\left(\mathbf{p}\right) = \mathbf{t}_{\mathrm{c}}\left(l\left(\mathbf{p}\right)\right).
	\end{equation}
	More details about the properties can be found in \cite{rezende2021constructive, quan2023distributed}. An example of the definitions above is also illustrated in Fig. \ref{area} and Fig. \ref{Vtc}.
	
	\subsection{Virtual tube for navigation}\label{vtt}
	In the previous work, the virtual tube was proposed for \emph{planning} (\cite{mao2023optimal}) and \emph{passing-through control} (\cite{gao2025distributed}). Restricting robots within the safety zone can simplify control, requiring only that no collisions occur between robots and the boundary of the virtual tube.  
	
	\begin{figure}[htbp]
		\centering
		\includegraphics[width=2.0in]
		{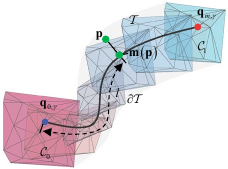}
		\caption{Modified by (\cite{mao2023optimal}). Definition of the virtual tube for spatial navigation. The tube is constructed from the terminal cross-sections $\mathcal{C}_0$ and $\mathcal{C}_1$, the generator curve $\gamma$, intermediate cross-sections $\mathcal{C}_l$, providing a safe traversal space for subsequent single-robot and swarm navigation.}
		\label{area}
	\end{figure}

	The related definitions of the virtual tube for planning are drawn to propose the following content. As shown in Fig.~\ref{area}, the virtual tube is defined as a set in an $n$-dimensional space represented by a 4-tuple $\left(\mathcal{C}_0,\mathcal{C}_1,\mathbf{f},\mathbf{h}\right)$ where
	$\mathcal{C}_0,\mathcal{C}_1$ are terminals and disjoint bounded convex subsets, $\mathbf{f}$ is a diffeomorphism that $\mathcal{C}_0\rightarrow\mathcal{C}_1$, and there is a set of order pairs $\mathcal{P}=\left\{\left(\mathbf{q}_0,\mathbf{q}_m\right) \vert\mathbf{q}_0\in\mathcal{C}_0,
	\mathbf{q}_m=\mathbf{f}\left(\mathbf{q}_0\right)\in\mathcal{C}_1 \right\}$. As for any $\mathbf{q}_0\in\mathcal{C}_0$ and the fixed diffeomorphism $\mathbf{f}$, $\mathbf{h}$ denotes the map  $\mathcal{P}\times\left[0,L\right]\rightarrow\mathcal{T}$, so the virtual tube is defined as 
	\begin{equation}
		\mathcal{T} = \left\{\mathbf{h}\left(\mathbf{q}_0,\mathbf{q}_m,l\right) \vert \left(\mathbf{q}_0,\mathbf{q}_m\right)\in\mathcal{P},l\in\left[0,L\right] \right\}
	\end{equation}
	where $l$ denotes curve parameter in (\cite{mao2023optimal}).
	Given an order pair $\left(\mathbf{q}_0,\mathbf{q}_m\right)$, $\mathbf{h}\left(\mathbf{q}_0,\mathbf{q}_m,l\right)$ denotes the path with $\mathbf{q}_0$ as the starting point and $\mathbf{q}_m$ as the end point, so the generator curve $\mathbf{\gamma}\left(l\right)$ starting at $\mathbf{q}_{0,\mathbf{\gamma}}$ is also denoted as $\mathbf{h}\left(\mathbf{q}_{0,\mathbf{\gamma}},\mathbf{q}_{m,\mathbf{\gamma}},l\right)$, where $l$ denotes the arc length in the following content. Then, the cross section $\mathcal{C}_l$ at $l$ is defined as
	\begin{equation}
		\mathcal{C}_l = \left\{ \mathbf{h}\left(\mathbf{q}_0,\mathbf{q}_m,l\right) \vert \left(\mathbf{q}_0,\mathbf{q}_m\right)\in\mathcal{P}\right\}. \label{crosssec}
	\end{equation}
	The surface of $\mathcal{T}$ is the boundary, defined as $\partial \mathcal{T}$. To plan a virtual tube in obstacle environments, the Tube-RRT* algorithm \cite{mao2025tube} is proposed as a implement, which is utilized in the simulations and experiments.
	
	\subsection{Transformation from Time Domain to Space Domain}
	The conversion between the time coordinate $t$ and space coordinate $l$ (\cite{xu2008spatial}) is denoted as 
	\begin{equation}
		\mathrm{D}_t = \frac{\mathrm{d}}{\mathrm{d} t} =  \frac{\mathrm{d}}{\mathrm{d} l}  \frac{\mathrm{d} l}{\mathrm{d} t} = v \mathrm{D}_l,
		\label{nablal}
	\end{equation}
	where $v=\mathrm{D}_t l$ denotes the progression speed along the spatial coordinate. For the single robot, $l$ is the closest-point coordinate on the path and $v$ is the executable tangential speed. For the swarm, $l$ is the effective macroscopic progression coordinate associated with the common traversal field. Since $l(t)=\int_0^t v(s)\,\mathrm{d}s$, the time-to-space transformation is well defined whenever $v>0$. This positivity is guaranteed by selecting a strictly positive traversal command and restricting the analysis to the bounded operating set specified below. Therefore, $l(t)$ is strictly increasing and admits a global inverse, and any time-domain signal can be expressed as a function of $l$.

	\section{Problem formulations} \label{problem}
	This section introduces the two navigation formulations before extracting the common energy systems analyzed in \emph{Section~\ref{IL}}. The single-robot formulation is expressed by a finite-dimensional path error, whereas the robotic-swarm formulation is expressed by a distribution error. Starting from the concrete closed-loop systems makes the origin and physical meaning of the common model explicit.
	
	For both formulations, the executable velocity is generated by the radial saturation mapping
	\begin{equation}
		\begin{split}
			\mathrm{sat}(\mathbf{x},v_{\mathrm{m}})&=
			\begin{cases}
				\mathbf{x}, & \Vert\mathbf{x}\Vert\leq v_{\mathrm{m}},\\
				v_{\mathrm{m}}\mathbf{x}/\Vert\mathbf{x}\Vert, & \Vert\mathbf{x}\Vert>v_{\mathrm{m}},
			\end{cases}\\
			\kappa_v(\mathbf{x})&=
			\begin{cases}
				1, & \Vert\mathbf{x}\Vert\leq v_{\mathrm{m}},\\
				v_{\mathrm{m}}/\Vert\mathbf{x}\Vert, & \Vert\mathbf{x}\Vert>v_{\mathrm{m}},
			\end{cases}
		\end{split}
		\label{5}
	\end{equation}
	where $v_{\mathrm{m}}>0$ is the executable velocity limit, and $		\mathrm{sat}(\mathbf{x},v_{\mathrm{m}})=\kappa_v (\mathbf{x}) \mathbf{x}$.
	
	\subsection{Single-robot formulation} \label{singleIL}
	\begin{figure}[htbp]
		\centering
		\includegraphics[width=3.1in]{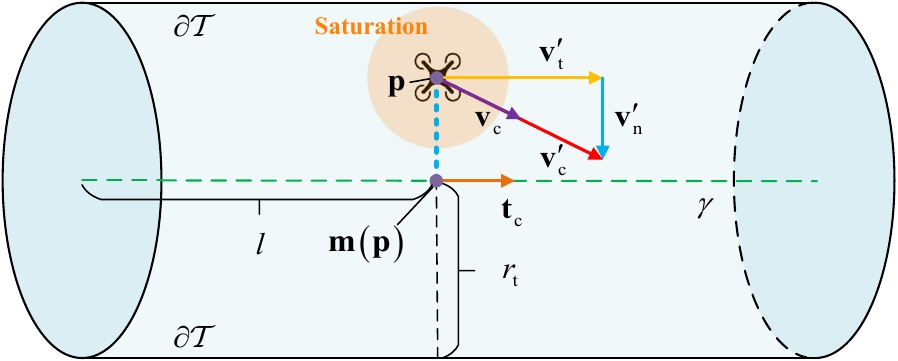}
		\caption{Geometric decomposition of the single-robot command. The nominal component regulates path error, the traversal component advances the robot along the tangent direction, and saturation produces the executable command.}
		\label{Vtc}
	\end{figure}
	
	Consider a robot whose position and velocity satisfy
	\begin{equation}
		\dot{\mathbf{p}}=\mathbf{v},
		\dot{\mathbf{v}}=\tau\left(\mathbf{v}_{\mathrm{c}}-\mathbf{v}\right),
		\label{veduc}
	\end{equation}
	where $\tau>0$ represents the closed-loop velocity response. With the path and closest-point map defined in \emph{Section~\ref{secDef}} and Fig. \ref{Vtc}, define
	\begin{equation}
		\mathbf{e}_{\mathrm{p}}=\mathbf{m}(\mathbf{p})-\mathbf{p},
		V_{\mathrm{s}}=\frac{1}{2}e_{\mathrm{p}}^2,
		e_{\mathrm{p}}=\Vert\mathbf{e}_{\mathrm{p}}\Vert.
		\label{singleenergy}
	\end{equation}
	Here $V_{\mathrm{s}}$ is a function measuring deviation from the geometric path rather than mechanical energy.
	
	\begin{assumption} \label{assumvtrack}
		The low-level loop satisfies $\mathbf{v}=\mathbf{v}_{\mathrm{c}}+\delta\mathbf{v}$ with $\Vert\delta\mathbf{v}\Vert\leq\varepsilon$ on the considered operating set.
	\end{assumption}
	
	The controller for navigating from start to the end is designed as
	\begin{equation}
		\begin{split}
			\mathbf{v}_{\mathrm{c}}(\mathbf{p}) &=\mathrm{sat}\left(\mathbf{v}^{\prime}_{\mathrm{c}}(\mathbf{p}),v_{\mathrm{m}}\right),\\
			\mathbf{v}^{\prime}_{\mathrm{c}} (\mathbf{p})&=\mathbf{v}^{\prime}_{\mathrm{n}}(\mathbf{p})+\mathbf{v}^{\prime}_{\mathrm{t}}(\mathbf{p}),\\
			\mathbf{v}^{\prime}_{\mathrm{n}}(\mathbf{p})&=-k_{\mathrm{p}}V_{\mathrm{s}}\nabla_{\mathbf{p}}e_{\mathrm{p}}(\mathbf{p}),\\
			\mathbf{v}^{\prime}_{\mathrm{t}}(\mathbf{p})&=v_{\mathrm{t}} (l(\mathbf{p}))\mathbf{t}_{\mathrm{c}} (\mathbf{p}).
		\end{split}
		\label{singlevc}
	\end{equation}
	Here $k_{\mathrm{p}}>0$ is the gain, the nominal component $\mathbf{v}^{\prime}_{\mathrm{n}}$ corrects the path error, whereas $\mathbf{v}^{\prime}_{\mathrm{t}}$ determines the progression rate along the path, and $v_{\mathrm{t}} = \Vert \mathbf{v}^{\prime}_{\mathrm{t}} \Vert$ is the traversal-speed profile. Let $v_{\mathrm{n}}=\Vert\mathbf{v}^{\prime}_{\mathrm{n}}\Vert$. Using $\mathbf{e}_{\mathrm{p}}^{\mathrm{T}}\mathbf{t}_{\mathrm{c}}=0$ and allowing a bounded velocity-execution error, the resulting dynamics is
	\begin{equation}
		\dot V_{\mathrm{s}}
		=-\lambda_{\mathrm{s}}\kappa_v v_{\mathrm{n}}^{3/2}
		+\varepsilon_{\mathrm{s}}v_{\mathrm{n}}^{1/2},
		\label{dotVsatleq}
	\end{equation}
	where $\lambda_{\mathrm{s}}>0$ and $\varepsilon_{\mathrm{s}}$ is bounded. Thus, increasing $v_{\mathrm{t}}$ may reduce $\kappa_v$ under active saturation and leave less command authority for path-error correction. \emph{Appendix~\ref{app:singleverification}} gives the complete calculation.
	
	\subsection{Robotic-swarm formulation} \label{swarmIL}
	Let $\mathcal{T}\subset\mathbb{R}^{3}$ be a bounded, connected virtual-tube domain with a Lipschitz boundary. The collective state of the swarm is represented by a normalized density $\rho(t,\mathbf{p})$, satisfying
	\begin{equation}
		\rho\geq0,
		\qquad
		\int_{\mathcal{T}}\rho(t,\mathbf{p})\,\mathrm{d}\mathbf{p}=1.
		\label{eq:swarm_density_normalization}
	\end{equation}
	For the microscopic motion $\dot{\mathbf{p}}_i=\mathbf{v}_{\mathrm{c}}(t,\mathbf{p}_i)$, $i=1,\ldots,M$, the corresponding macroscopic density evolves according to the continuity equation (\cite{krishnan2018distributed,lv2024mean})
	\begin{equation}
		\begin{split}
			\frac{\partial\rho}{\partial t}
			&=-\nabla_{\mathbf{p}}\cdot\left(\rho\mathbf{v}_{\mathrm{c}}\right),
			\mathbf{p}\in\mathcal{T},\\
			\mathbf{n}^{\mathrm{T}}\rho\mathbf{v}_{\mathrm{c}}&=0,
			\mathbf{p}\in\partial\mathcal{T}.
		\end{split}
		\label{swarmpde}
	\end{equation}
	Here, $\mathbf{n}$ denotes the unit inner normal to the boundary $\partial\mathcal{T}$. During the analyzed traversal interval, the density support remains away from the terminal cross sections, so the boundary term generated by integration by parts vanishes without prohibiting forward motion. The Wasserstein-space interpretation of (\ref{swarmpde}) and the finite-swarm density reconstruction are provided in \emph{Appendix~\ref{app:swarmverification}}.
	
	Let $\rho_{\mathrm{d}}(t,\mathbf{p})$ be a continuously differentiable desired density satisfying
	\begin{equation}
		\rho_{\mathrm{d}}\geq0,
		\qquad
		\int_{\mathcal{T}}\rho_{\mathrm{d}}(t,\mathbf{p})\,\mathrm{d}\mathbf{p}=1.
		\label{eq:desired_density_normalization}
	\end{equation}
	Its construction from the occupied cross sections of the virtual tube, followed by a feasible forward shift along the tube, is given in \emph{Appendix~\ref{app:swarmverification}}. The shift provides a feasible mechanism for limiting aggregate antagonism between the density-regulation and traversal fields; it does not impose a pointwise alignment requirement. Define the density error and the associated energy as
	\begin{equation}
		\phi(t,\mathbf{p})
		=\rho(t,\mathbf{p})-\rho_{\mathrm{d}}(t,\mathbf{p}),
		V_{\mathrm{m}}
		=\frac{1}{2}\int_{\mathcal{T}}\phi^{2}(t,\mathbf{p})\,\mathrm{d}\mathbf{p}.
		\label{swarmenergy}
	\end{equation}
	Because both densities are normalized, $\int_{\mathcal{T}}\phi\,\mathrm{d}\mathbf{p}=0$. The nominal regulation component and the traversal component are selected as
	\begin{equation}
		\mathbf{v}^{\prime}_{\mathrm{n}}
		\triangleq\mathbf{v}^{\prime}_{\phi}
		=-k_{\rho}\frac{\nabla_{\mathbf{p}}\phi}{\rho},
		\mathbf{v}^{\prime}_{\mathrm{t}}
		=v_{\mathrm{t}}\mathbf{t}_{\mathrm{c}},
		\qquad k_{\rho}>0.
		\label{vphicom}
	\end{equation}
	The analysis is restricted to an operating set on which $\rho$ is positive over the swarm support; in implementation, $\rho$ in the denominator can be replaced by $\rho+\delta_{\rho}$ with a small $\delta_{\rho}>0$.
	
	For a stationary $\rho_{\mathrm{d}}$, the nominal density-regulation field is exponentially dissipative:
	\begin{equation}
		\dot V_{\mathrm{m}}
		\leq-\frac{2k_{\rho}}{C_{\mathrm{P}}^{2}}V_{\mathrm{m}},
		\label{eq:swarm_nominal_energy_identity}
	\end{equation}
	where $C_{\mathrm{P}}>0$ depends only on $\mathcal{T}$. The exact identity and its Poincar\'e--Wirtinger derivation are given in \emph{Appendix~\ref{app:swarmverification}}.
	
	To regulate the density while moving the swarm forward, apply the pointwise saturated composite field
	\begin{equation}
		\begin{split}
			\mathbf{v}_{\mathrm{c}}(t,\mathbf{p})
			&=\kappa_v(t,\mathbf{p})
			\left(\mathbf{v}^{\prime}_{\mathrm{n}}+\mathbf{v}^{\prime}_{\mathrm{t}}\right),\\
			\kappa_v(t,\mathbf{p})
			&=\min\left\{1,
			\frac{v_{\mathrm{m}}}
			{\Vert\mathbf{v}^{\prime}_{\mathrm{n}}+\mathbf{v}^{\prime}_{\mathrm{t}}\Vert}
			\right\}.
		\end{split}
		\label{dswarmc}
	\end{equation}
	Let $v_{\mathrm{n}}\triangleq k_{\rho}\Vert\nabla_{\mathbf{p}}\phi\Vert_{\mathcal{L}^{2}(\mathcal{T})}$. Direct differentiation of $V_{\mathrm{m}}$ along (\ref{swarmpde}) and (\ref{dswarmc}) yields
	\begin{equation}
		\dot V_{\mathrm{m}}
		=-\frac{1}{k_{\rho}}\bar{\kappa}_{v}v_{\mathrm{n}}^{2}
		+\frac{\beta_{\rho}}{k_{\rho}}v_{\mathrm{t}}v_{\mathrm{n}}
		+\varepsilon_{1,\mathrm{m}},
		\label{dotVswarm}
	\end{equation}
	where $\bar{\kappa}_{v}\in(0,1]$ represents the effect of velocity saturation, $\beta_{\rho}$ characterizes the interaction between forward traversal and density regulation, and $\varepsilon_{1,\mathrm{m}}$ accounts for variation of the desired density. These coefficients are bounded on the considered compact operating set. Their exact integral definitions, bounds, and the complete calculation are given in \emph{Appendix~\ref{app:swarmverification}}.
	
	\subsection{Common energy system}
	\label{sec:commonenergysystem}
	Although the two formulations evolve on different state spaces, their regulation performance is characterized by the scalar energies defined in (\ref{singleenergy}) and (\ref{swarmenergy}). Then the unified energy
	\begin{equation}
		V\triangleq
		\begin{cases}
			V_{\mathrm{s}}, & \text{for the single robot}\\
			V_{\mathrm{m}}, & \text{for the robotic swarm}
		\end{cases}
		\label{eq:common_energy_definition}
	\end{equation}
	is introduced. By construction, $V$ is quadratic in the corresponding regulation error and satisfies $V\geq0$ with $V=0\Longleftrightarrow
			\mathbf{e}_{\mathrm{p}}=\mathbf{0}$ or $\phi=0$
	, and $V>0$ otherwise. Hence, $V$ is positive definite with respect to the corresponding regulation error. The term \emph{energy} is used in this Lyapunov sense rather than in the sense of mechanical energy.
	
	From above, equations~(\ref{singlevc}) and (\ref{dswarmc}) share the command decomposition
	\begin{equation}
		\mathbf{v}^{\prime}_{\mathrm{c}}
		=\mathbf{v}^{\prime}_{\mathrm{n}}+\mathbf{v}^{\prime}_{\mathrm{t}},
		\qquad
		\mathbf{v}_{\mathrm{c}}
		=\mathrm{sat}\left(\mathbf{v}^{\prime}_{\mathrm{c}},v_{\mathrm{m}}\right),
		\label{compositecommand}
	\end{equation}
	where $\mathbf{v}^{\prime}_{\mathrm{n}}$ dissipates the energy by regulating the corresponding error, whereas $\mathbf{v}^{\prime}_{\mathrm{t}}$ generates the overall navigation motion.
	
	\begin{proposition}
		\label{prop:commonenergy}
		For the single-robot and robotic-swarm formulations, the energy $V$ satisfies
		\begin{equation}
			\dot V
			=-\lambda_1\kappa_E v_{\mathrm{n}}^{p}
			+\varepsilon v_{\mathrm{t}}^{o}v_{\mathrm{n}}^{q}
			+\varepsilon_1,
			\label{Vmodel2}
		\end{equation}
		where $\lambda_1>0$, $p>q\geq0$, $o\geq0$, $0<\kappa_E\leq1$, and $\varepsilon$ and $\varepsilon_1$ are bounded on a compact operating set.
	\end{proposition}
	\begin{proof}
		See \emph{Appendix~\ref{app:proof-common-energy}}.
	\end{proof}
	
	Equation~(\ref{Vmodel2}) consists of a nominal energy-dissipation term and bounded coupling and residual terms. In the nominal case $\varepsilon=\varepsilon_1=0$,
	\begin{equation}
		\dot V=-\lambda_1\kappa_Ev_{\mathrm{n}}^{p}\leq0.
		\label{eq:nominal_common_energy_decay}
	\end{equation}
	Moreover, under the path-geometry and normalized-density assumptions stated above, $v_{\mathrm{n}}=0$ if and only if $V=0$. Consequently,
	\begin{equation}
		V>0\ \Longrightarrow\ \dot V<0,
		V=0\ \Longrightarrow\ \dot V=0,
		\label{eq:common_energy_lyapunov_property}
	\end{equation}
	so $V$ is a Lyapunov function for the single-robot error dynamics and a Lyapunov function for the swarm error dynamics. Under bounded nonideal effects, (\ref{Vmodel2}) gives the corresponding perturbed energy dynamics.
	
	On the considered compact operating set, boundedness of the unsaturated composite command ensures that there exists $\underline{\kappa}_E>0$ such that
	\begin{equation}
		0<\underline{\kappa}_E\leq\kappa_E\leq1.
		\label{eq:kappaE_lower_bound}
	\end{equation}
	If $|\varepsilon|\leq\bar{\varepsilon}$, $|\varepsilon_1|\leq\bar{\varepsilon}_1$, and $v_{\mathrm{t}}\leq\bar v_{\mathrm{t}}$, then
	\begin{equation}
		\dot V
		\leq
		-\lambda_1\underline{\kappa}_E v_{\mathrm{n}}^{p}
		+\bar{\varepsilon}\bar v_{\mathrm{t}}^{o}v_{\mathrm{n}}^{q}
		+\bar{\varepsilon}_1.
		\label{eq:common_energy_drift_bound}
	\end{equation}
	Since $p>q$, the dissipative term dominates the lower-order coupling term for sufficiently large $v_{\mathrm{n}}$, yielding negative energy drift outside a bounded neighborhood determined by $\bar{\varepsilon}$ and $\bar{\varepsilon}_1$.
	
	The spatial movement formulated by VIP is generated by the executable traversal component. Accordingly, both formulations use
	\begin{equation}
		v=\kappa_E v_{\mathrm{t}}.
		\label{eq:v_vt_relation}
	\end{equation}
	For the single robot, this relation follows by projecting the saturated command onto $\mathbf{t}_{\mathrm{c}}$, since the nominal regulation component is normal to the path. For the swarm, $v$ denotes the effective macroscopic progression rate associated with the same energy-weighted saturation coefficient appearing in (\ref{dotVswarm}). Thus, the energy dynamics and the spatial coordinate use the same effective command scaling.
	
	By designing $v_{\mathrm{t}}\geq\underline{v}_{\mathrm{t}}>0$, (\ref{eq:kappaE_lower_bound}) and (\ref{eq:v_vt_relation}) give
	\begin{equation}
		v\geq\underline{\kappa}_E\underline{v}_{\mathrm{t}}>0,
		\label{eq:positive_progression}
	\end{equation}
	which guarantees the invertibility required by the time-to-space transformation in (\ref{nablal}). VIP is subsequently developed directly from the common energy dynamics (\ref{Vmodel2}).
	
	\subsection{Regularity and saturation properties}
	\label{sec:saturationproperties}
	The following derivatives are understood almost everywhere because the radial saturation map is piecewise continuously differentiable. For the swarm, differentiation with respect to $V$ follows the radial scalarization of the current regulation field specified in \emph{Appendix~\ref{app:proof-saturation-derivatives}}.
	
	\begin{proposition}
		\label{prop:saturation_derivatives}
		Suppose that the nominal-command magnitude is nondecreasing with the energy. For the single robot, use the normal--tangent command decomposition in (\ref{singlevc}). For the swarm, construct the desired density by the forward-shift design in \emph{Appendix~\ref{app:desired-density-design}}, with the shift selected according to (\ref{eq:aggregate_acute_design}). Then
		\begin{equation}
			\frac{\partial\kappa_E}{\partial V}\leq0,
			\frac{\partial\kappa_E}{\partial v_{\mathrm{t}}}\leq0,
			\frac{\partial(\kappa_Ev_{\mathrm{t}})}{\partial v_{\mathrm{t}}}\geq0.
			\label{eq:saturation_monotonicity}
		\end{equation}
		The inequalities are strict whenever the corresponding regulation or traversal component is nonzero on a set of nonzero weighted measure.
	\end{proposition}
	\begin{proof}
		See \emph{Appendix~\ref{app:proof-saturation-derivatives}}.
	\end{proof}
	
	\begin{remark}
		For the single robot, the required signs follow from the orthogonal command construction and therefore introduce no additional geometric condition. For the swarm, they follow from the integral acute-angle criterion used to select the forward shift of the desired density. This criterion constrains only the weighted aggregate interaction between the two fields; local opposing motions remain admissible during density redistribution.  This design aims to ensure that, during the optimization process, all variables can be optimized by optimizing only one variable. In other words, it achieves this by saturating and harmonizing the relationships between variables.
		\label{remark:sat}
	\end{remark}
	
	\subsection{Problem formulation}
	
	By combining \emph{Proposition~\ref{prop:commonenergy}}, the saturation relation in \emph{Remark~\ref{remark:sat}}, and (\ref{eq:v_vt_relation}), the traversal time
	$T=\int_{0}^{L}v^{-1}\mathrm{d}l$ can be expressed in the spatial domain. Accordingly, the navigation problem is formulated in the following unified integral form:
	\begin{equation}
		\begin{aligned}
			\min_{v_{\mathrm{t}}(l),\,l\in[0,L]}\quad
			J &= \int_{0}^{L}
			\left( \frac{1}{\kappa_E v_{\mathrm{t}}} +k_V V\right)\mathrm{d}l,\\
			\mathrm{s.t.}\quad
			&(\ref{Vmodel2}),\qquad
			\Vert\mathbf{v}_{\mathrm{c}}\Vert\leq v_{\mathrm{m}},
		\end{aligned}
		\label{opcommon}
	\end{equation}
	where $k_V>0$ is a weighting coefficient that balances traversal efficiency and navigation safety. 
	
	The term $1/(\kappa_E v_{\mathrm{t}})$ represents the local traversal-time cost per unit path length; minimizing its spatial integral encourages a larger admissible tangential velocity and, consequently, a shorter overall traversal time. The term $k_V V$ penalizes the accumulated energy-like state deviation along the route, thereby discouraging excessive deviation from the desired navigation state and improving the safety and regularity of the resulting motion. Meanwhile, through \emph{Proposition~\ref{prop:commonenergy}}, the original system dynamics are equivalently represented by the spatial energy model (\ref{Vmodel2}), allowing the navigation performance and system evolution to be optimized within a unified energy-based formulation.
	

	\section{VIP: variation-based iterative-learning planning} \label{IL}
	This section develops VIP in two layers. The model-based variation law provides a theoretical reference direction in the function space, whereas the model-free law approximates that direction from the energy profile recorded during a rollout. The latter is the practical mechanism used to achieve \emph{low computational cost} and \emph{online model-free learning} across heterogeneous tasks.
	
	\subsection{Preliminary}
	
	Firstly, the following lemmas are presented (\cite{debnath2005introduction, rafajlowicz2018iterative}), which are well known.
	
	For a given $v_{\mathrm{t}}\in\mathcal{L}^{2}(0,L)$, define the Hamiltonian $H$of $J$ as
	\begin{equation}
		H\left(V,v_{\mathrm{t}},\psi\right)
		=
		\frac{1}{\kappa_Ev_{\mathrm{t}}}
		+k_VV
		+\psi \mathrm{D}_lV,
		\label{Hamiltonian}
	\end{equation}
	where $\mathrm{D}_l V =   \dot{V} / v$
	follows from \eqref{nablal}. The corresponding adjoint variable $\psi$ is determined by
	\begin{equation}
		\mathrm{D}_l\psi
		=
		-\frac{\partial H}{\partial V},
		\psi(L)=0.
		\label{functionalC}
	\end{equation}
	
	\begin{lemma}\label{gateaux}
		The Gateaux differential of $J$ at $v_{\mathrm{t}} \in \mathcal{L}^2\left(0,L\right)$ in the direction $\overline{v}\in \mathcal{L}^2\left(0,L\right)$ has the following form:
		\begin{equation}
			\frac{\mathrm{d}J\left(v_{\mathrm{t}} +\epsilon\overline{v}\right)}{\mathrm{d} \epsilon}\vert_{\epsilon=0} = \int_{0}^{L} F\left(V,v_{\mathrm{t}},\psi \right) \overline{v} \left(l\right) \mathrm{d} l.
		\end{equation}
		If the Frechet derivative of $J$ exists, then $F$ in this case is equal to
		\begin{equation}
			F\left(V,v_{\mathrm{t}},\psi\right) = \frac{\partial H\left(V,v_{\mathrm{t}},\psi\right)}{\partial v_{\mathrm{t}}}.
		\end{equation}
	\end{lemma}
	
	\begin{lemma}\label{descent}
		Suppose that $J$ is Fréchet differentiable at $v_{\mathrm{t}}$ and that
		$F\left(V,v_{\mathrm{t}},\psi\right)$ is its $\mathcal{L}^{2}$-gradient. If
		$F\left(V,v_{\mathrm{t}},\psi\right)\not\equiv 0$, then
		\begin{equation}
			\overline{v} = -F\left(V,v_{\mathrm{t}},\psi\right)
		\end{equation}
		is a descent direction of $J$ at $v_{\mathrm{t}}$. If
		$F\left(V,v_{\mathrm{t}},\psi\right)\equiv 0$, then $v_{\mathrm{t}}$ satisfies the first-order stationarity condition.
	\end{lemma}
	
\begin{figure}[htbp]
		\centering
		\includegraphics[width=3.0in]{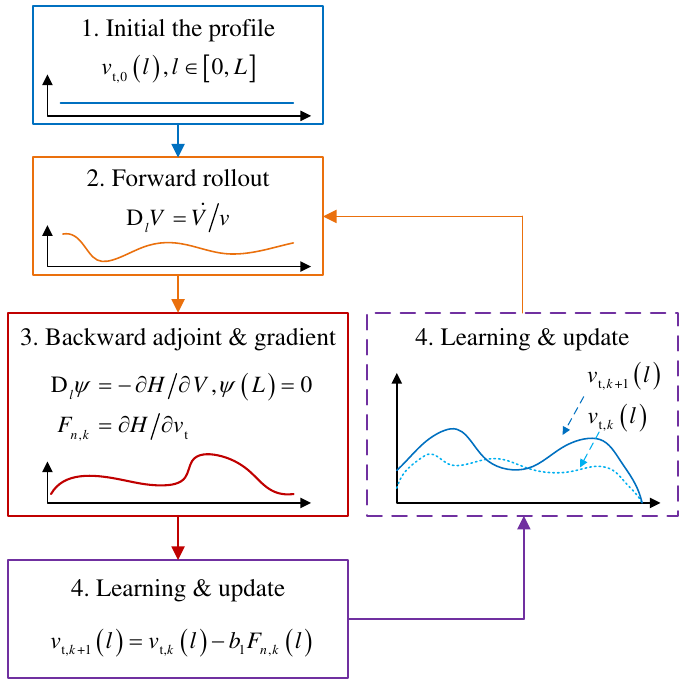}
		\caption{Workflow of model-based optimization. The profile is initialized, propagated forward, corrected with the backward adjoint and gradient, and iteratively updated until convergence.}
		\label{variation}
	\end{figure}

	The optimization process is illustrated in Fig. \ref{variation}, where $F$ acts as variation gradient flow to optimize $J$ from a functional perspective.
	
	\subsection{Model-based VIP}
	\subsubsection{Variation} \label{modelIL}
	Using (\ref{eq:v_vt_relation}), the time-to-space transformation gives
	\begin{equation}
		\mathrm{D}_lV
		=\frac{-\lambda_1\kappa_Ev_{\mathrm{n}}^p
			+\varepsilon v_{\mathrm{t}}^ov_{\mathrm{n}}^q
			+\varepsilon_1}{\kappa_Ev_{\mathrm{t}}}.
		\label{eq:spatial_common_energy}
	\end{equation}
	In the following partial-derivative calculation, $\lambda_1,p,q$, and $o$ are fixed  parameters, $v_{\mathrm{n}}=v_{\mathrm{n}}(V)$, and $\kappa_E=\kappa_E(V,v_{\mathrm{t}},l)$. The bounded coefficients $\varepsilon(l)$ and $\varepsilon_1(l)$ are treated as exogenous spatial profiles during one rollout. This is the same dependence convention used in the model-free sign analysis below.
	
	By introducing the \emph{Hamiltonian} $H$ with (\ref{eq:spatial_common_energy}), there exists
	\begin{equation}
		H=
		\underbrace{\frac{1}{\kappa_Ev_{\mathrm{t}}}+k_VV}_{H_1}
		+\psi(l)
		\underbrace{\frac{-\lambda_1\kappa_Ev_{\mathrm{n}}^p
				+\varepsilon v_{\mathrm{t}}^ov_{\mathrm{n}}^q
				+\varepsilon_1}{\kappa_Ev_{\mathrm{t}}}}_{H_2},
		\label{eq:common_hamiltonian}
	\end{equation}
	and the optimal condition is written as
	\begin{equation} \label{psit}
		\begin{split}
			\mathrm{D}_l\psi\left(l\right) &= -\frac{\partial H}{\partial V}= -\frac{\partial H_1}{\partial V} - \psi\left(l\right)\frac{\partial H_2}{\partial V} \\
			&= W\left(l\right) + \psi\left(l\right) P\left(l\right)
		\end{split}
	\end{equation}
	with $W\left(l\right) = -\partial H_1 / \partial V$ and $P\left(l\right) = - \partial H_2 / \partial V$.
	
	The Gateaux differential is then calculated
	\begin{equation}
		\begin{split} \label{gde}
			F_n \left(l\right) &= \frac{\partial H}{\partial v_{\mathrm{t}}} =  \frac{\partial H_1}{\partial v_{\mathrm{t}}} + \psi\left(l\right)\frac{\partial H_2}{\partial v_{\mathrm{t}}} \\
			&= M\left(l\right) + \psi\left(l\right) N\left(l\right)
		\end{split}
	\end{equation}
	with $M\left(l\right) = \partial H_1 / \partial v_{\mathrm{t}}$ and $N\left(l\right) = \partial H_2 / \partial v_{\mathrm{t}}$.
	
	Let $\{v_{\mathrm{t},k}\}$ denote the sequence of commands generated through iterative learning, where $k \geq 0$ is the learning-iteration index. For a given iterate $v_{\mathrm{t},k}$, applying \emph{Lemma~\ref{gateaux}} along the direction $\overline{v}=-F_{n,k}$ gives
	\begin{equation}
		\frac{\mathrm{d}
			J\left(v_{\mathrm{t},k}-\epsilon F_{n,k}\right)}
		{\mathrm{d}\epsilon}
		\vert_{\epsilon=0}
		=
		-\int_{0}^{L}F_{n,k}^{2}(l)\mathrm{d}l
		\leq0.
		\label{eq:descent_variation}
	\end{equation}
	Thus, by \emph{Lemma~\ref{descent}}, $-F_{n,k}$ provides a local steepest-descent direction of $J$. Taking a finite step along this direction yields the learning law
	\begin{equation}
		v_{{\mathrm{t}},k+1}(l)	= v_{{\mathrm{t}},k}(l) -b_1 F_{n,k}(l),
		\label{eq:learning_law}
	\end{equation}
	where $b_1>0$ is the learning step size and $k$ denotes the learning-iteration index. So from (\ref{gde}) there exists
	\begin{equation} \label{il1law}
		\begin{split}
			v_{\mathrm{t},k+1}\left(l\right)  = v_{\mathrm{t},k} \left(l\right)  - b_1  \left(M_k\left(l\right) + \psi_k\left(l\right) N_k\left(l\right) \right).
		\end{split}
	\end{equation}
	By solving (\ref{psit}), the adjoint variable is obtained as
	\begin{equation} \label{psidata}
		\psi_k\left(l\right) = e^{\int_{0}^{l} P_k\left(s\right) \mathrm{d} s} \left(\int_{0}^{l} W_k\left(s\right) e^{-{\int_{0}^{s} P_k\left(\sigma\right) \mathrm{d} \sigma}} \mathrm{d}s + b_2\right)
	\end{equation}
	where $b_2 = \psi \left(0\right)>0$ is selected so as to ensure $\psi\left(L\right) = 0$. After finding such $b_2$ and substituting it into (\ref{psidata}), there exists the ``reverse'' expression
	\begin{equation} \label{psireverse}
		\psi_k\left(l\right) = -e^{\int_{0}^{l} P_k\left(s\right) \mathrm{d} s} \int_{l}^{L} W_k\left(s\right) e^{-{\int_{0}^{s} P_k\left(\sigma\right) \mathrm{d} \sigma}} \mathrm{d}s.
	\end{equation}

	\subsubsection{Convergence}
	\begin{theorem} \label{theoremmodelbased}
		Assume that $J$ is bounded from below and that its Fr\'echet derivative with respect to $v_{\mathrm{t}}$ is $F_m$-Lipschitz continuous on the considered neighborhood. For the model-based IL law (\ref{il1law}), if $0<b_1<2/F_m$, then $\{J(v_{\mathrm{t},k})\}$ is monotonically non-increasing and convergent, and $\lim_{k\rightarrow\infty}\Vert F_{n,k}\Vert_{\mathcal{L}^2(0,L)}=0$.
		
	\end{theorem}
	\begin{proof}
		See \emph{Appendix \ref{prooftheoremmodelbased}}. 
	\end{proof}
	
	\subsection{Model-free VIP}
	
	The model-based law gives the exact functional-gradient direction, but its implementation requires $\kappa_E$, the energy modeling, their derivatives, and the adjoint variable. The model-free design therefore uses the model-based result only to determine how the update direction should change with the recorded energy. No analytical derivative is evaluated during implementation.
	
	\subsubsection{Approximate learning law}\label{freeIL}
	Using $v=\kappa_Ev_{\mathrm{t}}$, the two Hamiltonian terms in (\ref{eq:common_hamiltonian}) can be written directly as
	\begin{equation}
		\begin{split}
			H_1
			&=\frac{1}{\kappa_Ev_{\mathrm{t}}}+k_VV,\\
			H_2
			&=-\frac{\lambda_1v_{\mathrm{n}}^p}{v_{\mathrm{t}}}
			+\frac{\varepsilon v_{\mathrm{t}}^{o-1}v_{\mathrm{n}}^q}{\kappa_E}
			+\frac{\varepsilon_1}{\kappa_Ev_{\mathrm{t}}}.
		\end{split}
		\label{eq:H12_modelfree}
	\end{equation}
	This form is valid for both formulations: $\kappa_E=\kappa_v$ for the single robot and $\kappa_E=\bar{\kappa}_v$ for the swarm. During one rollout, $v_{\mathrm{n}}=v_{\mathrm{n}}(V)$, while $\varepsilon$ and $\varepsilon_1$ are treated as bounded spatial profiles, consistently with the model-based calculation.
	
	The direct traversal-time contribution to the gradient in (\ref{gde}) is
	\begin{equation}
		\begin{split}
			M(l)
			&=\frac{\partial H_1}{\partial v_{\mathrm{t}}}\\
			&=-\frac{1}{(\kappa_Ev_{\mathrm{t}})^2}
			\frac{\partial(\kappa_Ev_{\mathrm{t}})}{\partial v_{\mathrm{t}}}\leq0,
		\end{split}
		\label{Ml}
	\end{equation}
	where the inequality follows from \emph{Proposition~\ref{prop:saturation_derivatives}}. Therefore, increasing the traversal command can only increase or maintain the executable traversal speed, and consequently can only decrease or maintain the local traversal-time cost $1/v$.
	
	Differentiating $H_2$ with respect to $v_{\mathrm{t}}$ gives
	\begin{equation}
		\begin{split}
			N(l)
			&=\frac{\partial H_2}{\partial v_{\mathrm{t}}}\\
			&=\frac{\lambda_1v_{\mathrm{n}}^p}{v_{\mathrm{t}}^2}
			+\varepsilon v_{\mathrm{n}}^q
			\left[
			\frac{(o-1)v_{\mathrm{t}}^{o-2}}{\kappa_E}
			-\frac{v_{\mathrm{t}}^{o-1}}{\kappa_E^2}
			\frac{\partial\kappa_E}{\partial v_{\mathrm{t}}}
			\right]\\
			&\quad
			-\frac{\varepsilon_1}{(\kappa_Ev_{\mathrm{t}})^2}
			\frac{\partial(\kappa_Ev_{\mathrm{t}})}{\partial v_{\mathrm{t}}}.
		\end{split}
		\label{Nl}
	\end{equation}
	The first term in (\ref{Nl}) is positive and proportional to $v_{\mathrm{n}}^p$. The remaining terms contain either $v_{\mathrm{n}}^q$ or the bounded residual $\varepsilon_1$. Since $p>q$, $v_{\mathrm{t}}$ and $\kappa_E$ are bounded away from zero by (\ref{eq:kappaE_lower_bound})--(\ref{eq:positive_progression}), and the derivatives of $\kappa_E$ are bounded on the compact operating set, the positive regulation term becomes dominant as $v_{\mathrm{n}}$ increases.
	
	To determine the sign of the adjoint variable, note that
	\begin{equation}
		\begin{split}
			W(l)
			&=-\frac{\partial H_1}{\partial V}\\
			&=\frac{1}{\kappa_E^2v_{\mathrm{t}}}
			\frac{\partial\kappa_E}{\partial V}-k_V\\
			&\leq-k_V<0,
		\end{split}
		\label{Ql}
	\end{equation}
	where $\partial\kappa_E/\partial V\leq0$ follows from \emph{Proposition~\ref{prop:saturation_derivatives}}: pointwise orthogonality gives the result for the single robot, while the integral acute-angle design in \emph{Appendix~\ref{app:desired-density-design}} gives it for the swarm. Therefore, the original requirement $k_V>0$ is sufficient; no additional lower bound on the objective weight is introduced. The coefficient $P(l)=-\partial H_2/\partial V$ remains bounded on the same compact operating set; its sign is not required. From the reverse adjoint expression (\ref{psireverse}), $W(l)<0$ and $\psi(L)=0$ imply
	\begin{equation}
		\psi(l)\geq0,
		\qquad l\in[0,L].
		\label{eq:psi_nonnegative}
	\end{equation}
	
	The sign transition of the exact model-based direction now follows without introducing task-specific sensitivity variables. From (\ref{gde}),
	\begin{equation}
		F_n(l)=M(l)+\psi(l)N(l).
	\end{equation}
	When the energy is small, the regulation command $v_{\mathrm{n}}$ is small and the nonpositive traversal term $M$ favors increasing $v_{\mathrm{t}}$ whenever additional progression authority is available. As the energy increases, $v_{\mathrm{n}}$ increases and the positive $v_{\mathrm{n}}^p$ term in $N$ eventually dominates the lower-order and bounded terms. Since $\psi\geq0$, the exact gradient can consequently change from negative to positive. This gives the same acceleration--deceleration principle for both $V_{\mathrm{s}}$ and $V_{\mathrm{m}}$: accelerate when the recorded energy is small and decelerate when regulation requires more command authority.
	
	The approximate model-free learning law is therefore designed as
	\begin{equation}
		\begin{split}\label{approximateILlaw}
			v_{\mathrm{t},k+1}
			=v_{\mathrm{t},k}-b_3g\left(\zeta_k\right),
		\end{split}
	\end{equation}
	where $b_3>0$ and $\zeta_k\triangleq k_{\mathrm{e}}V_k$, with $k_{\mathrm{e}}>0$ mapping the  energy to a common dimensionless coordinate. The scaling coefficient $k_{\mathrm{e}}$ is distinct from the single-robot regulation gain $k_{\mathrm{p}}$ in (\ref{singlevc}). The function $g(\zeta_k)$ increases $v_{\mathrm{t}}$ when the normalized energy coordinate is small, decreases it when the coordinate is large, and leaves it unchanged inside the admissible band.
	
	Specifically, $g(\zeta)$ is designed to satisfy
	\begin{equation}
		\begin{split}
			&g(\zeta)=0,
			\qquad \zeta\in[\zeta_{\min},\zeta_{\max}],\\
			&0\leq \mathrm{D}_{\zeta}g\leq\beta,
			\qquad
			\mathrm{D}_{\zeta}g=0
			\iff \zeta\in[\zeta_{\min},\zeta_{\max}],
		\end{split}
		\label{ILlawestimate}
	\end{equation}
	where $\zeta_{\min},\zeta_{\max},\beta\in\mathbb{R}_{+}$ are design constants. Because $v_{\mathrm{n}}$ is nondecreasing with respect to $V$, the same energy band represents the sign-transition region of the model-based direction after scaling. The condition $v_{\mathrm{n}}\geq v_{\mathrm{m}}$ at $\zeta=\zeta_{\min}$ can be used to place the lower boundary near the onset of saturation. Thus, the recorded-energy law preserves the qualitative direction derived from (\ref{Ml})--(\ref{Ql}) without evaluating $\kappa_E$, its derivatives, or the adjoint variable online.
	
	\begin{remark}
		Recalling the objective interaction represented by (\ref{Vmodel2}), the proposed model-free law also accommodates the non-conflicting case $\partial\dot V/\partial v_{\mathrm{t}}\leq0$. If a larger traversal command improves energy dissipation, the recorded $V_k$ remains small and the learning law does not introduce an unnecessary deceleration correction. The traversal command can therefore increase toward the admissible limit while the energy continues to decrease. Therefore, the proposed learning law is not limited by the design limitations of \emph{Proposition \ref{prop:saturation_derivatives}}. This design is merely for the purpose of simplifying the analysis.
	\end{remark}
	
	The model-based calculation establishes the sign change that motivates the update, whereas the practical model-free implementation requires only the spatial energy profile $V_k(l)$ recorded during the preceding rollout.
	
	\subsubsection{Convergence}
	With the design of the model-free learning law (\ref{ILlawestimate}), the approximate convergence of the learned traversal command can be stated as follows.
	
	\begin{theorem}
		\label{theoremnear}
		Assume that $J$ is bounded from below and that its Fr\'echet derivative with respect to $v_{\mathrm{t}}$ is Lipschitz continuous with constant $F_m>0$. Suppose that the zero point of the model-based descent direction is contained in the admissible energy band $[\zeta_{\min},\zeta_{\max}]$. Then, for a sufficiently small learning gain $b_3>0$, the model-free learning law (\ref{approximateILlaw}) generates a monotonically non-increasing cost sequence $\{J(v_{\mathrm{t},k})\}$. Moreover, $
		\lim_{k\rightarrow\infty}\Vert g_k\Vert_{\mathcal{L}^2(0,L)}=0 .$
		Therefore, the learned energy approaches the admissible band determined by $k_{\mathrm{e}}V_k(l)\in[\zeta_{\min},\zeta_{\max}]$. 
	\end{theorem}
	\begin{proof}
		See \emph{Appendix \ref{prooftheoremnear}}. 
	\end{proof}
	
	\begin{figure}[htbp]
		\centering
		\includegraphics[width=3.3in]
		{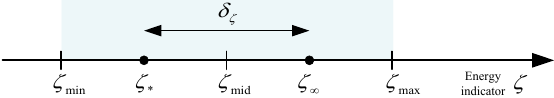}
		\caption{Admissible energy band and approximation offset in the model-free learning law.}
		\label{modelfree}
	\end{figure}
	
	\begin{remark}
		As shown in Fig. \ref{modelfree}, the parameters $\zeta_{\min}$ and $\zeta_{\max}$ in (\ref{ILlawestimate}) determine the admissible energy band of the model-free learning law. Since the exact zero point of the model-based descent direction is generally unavailable, the designed band may deviate from the ideal stationary point. Let $\zeta_{*}$ denote the zero point of the model-based descent direction, namely,
		\begin{equation}
			F_{n,k}(\zeta_{*})=0,
		\end{equation}
		and let $\zeta_{\mathrm{mid}} = \left(\zeta_{\min}+\zeta_{\max}\right)/2$ denote the center of the designed admissible band.
		If the learned energy converges to a point $\zeta_{\infty}\in[\zeta_{\min},\zeta_{\max}]$, then it can be written as
		\begin{equation}
			\zeta_{\infty}=\zeta_{*}+\delta_{\zeta} ,
		\end{equation}
		where $\delta_{\zeta}$ denotes the offset between the learned solution and the zero point of the model-based descent direction. The offset satisfies
		\begin{equation}
			|\delta_{\zeta}|
			\leq |\zeta_{\mathrm{mid}}-\zeta_{*}|+\frac{\zeta_{\max}-\zeta_{\min}}{2}.
		\end{equation}
		Therefore, the approximation error is determined by both the center offset $|\zeta_{\mathrm{mid}}-\zeta_{*}|$ and the band width $\zeta_{\max}-\zeta_{\min}$.
		
		Let $\widehat J(\zeta)$ denote the restriction of $J$ to the local one-parameter family indexed by the normalized energy coordinate. Since the Fr\'echet derivative is Lipschitz continuous and $F_{n,k}(\zeta_{*})=0$, the local cost variation satisfies
		\begin{equation}
			\widehat J(\zeta_{\infty})-\widehat J(\zeta_{*})
			\leq \frac{F_m}{2}
			\Vert \delta_{\zeta} \Vert^2_{\mathcal{L}^2(0,L)} .
		\end{equation}
		Hence, a smaller band width and a more accurate band center lead to a smaller sub-optimality gap. In practice, $[\zeta_{\min},\zeta_{\max}]$ provides a tunable compromise: a narrow band improves approximation accuracy, whereas a wider band improves robustness to modeling uncertainty, measurement noise, and iteration fluctuations.
	\end{remark}
	
	\subsubsection{Time complexity} \label{Complexity}
	
	The computational efficiency of the proposed model-free VIP mainly stems from its data-driven and pointwise update structure. After each trial, only the energy-related quantities distributed along the spatial coordinate are recorded, based on which the traversal command is updated. At each discretized point, the update requires only the evaluation of the scalar function $g(\cdot)$ and an algebraic correction of $v_{\mathrm{t},k}$. Therefore, no adjoint equation, system Jacobian, lifted input--output model, matrix inversion, or constrained optimization problem is required.
	
	In contrast, conventional trajectory optimization methods usually parameterize the trajectory using finite-dimensional variables. For example, polynomial-based methods optimize polynomial coefficients or waypoints subject to boundary, continuity, dynamic-feasibility, and collision-avoidance constraints. MPCC optimizes predicted states and control inputs over a finite horizon while considering system dynamics, contouring errors, and input constraints. These coupled optimization problems generally require repeated gradient evaluations, matrix factorizations, or numerical iterations, and their computational burden increases with the number of decision variables and constraints. Depending on the problem formulation and numerical solver, the resulting complexity can exhibit superlinear growth, such as $\mathcal{O} (n^2)$ or higher in some cases, with $n$ path segments.
	
	VIP instead optimizes the traversal command directly in the function space and updates its spatial profile using the recorded energy information. It therefore avoids repeatedly solving coupled optimization problems. Since each spatial point requires only a constant number of scalar evaluations and algebraic operations, the computational complexity of one learning iteration is $\mathcal{O}(n)$. The memory complexity is also $\mathcal{O}(n)$, as only the traversal-command profile and the recorded energy-related quantities need to be stored. If the learning process terminates after at most $k^*$ iterations, the total update complexity is $\mathcal{O}(k^*n)$.
	
	Thus, the computational advantage of VIP arises from replacing the coupled optimization of finite-dimensional trajectory variables with pointwise scalar evaluations and explicit algebraic updates, making it suitable for computation-constrained implementations. This analysis considers only the computational overhead of the learning update. The execution time of a physical experiment or simulation rollout is determined by the plant or simulator and is not included in $\mathcal{O}(k^*n)$.

	\begin{figure*}[htbp]
		\centering
		\includegraphics[width=6.7in]{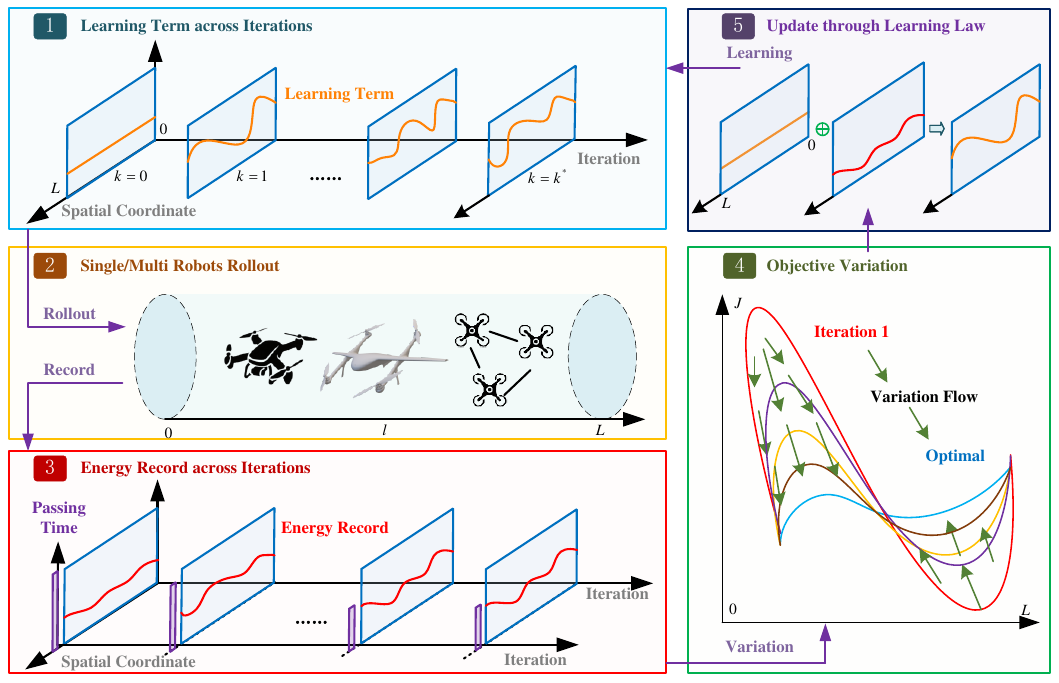}
		\caption{Structure of VIP and its two rollout modes. The energy may describe path tracking error or swarm-distribution error. In the model-in-the-loop mode, rollouts are generated by a nominal simulator before deployment. In the robot-in-the-loop mode, the energy profile is recorded during a physical traversal and the command is updated before the next repeated traversal. Both modes use the same spatial model-free update, whose role is to determine how fast the robot or swarm traverses each path location.}
		\label{structure}
	\end{figure*}
	
	\subsection{Implementation modes of VIP} \label{vipimplementation}
	Fig.~\ref{structure} summarizes a common rollout--record--evaluate--update structure. The traversal profile $v_{\mathrm{t},k}(l)$ is discretized over $l\in[0,L]$ during the $k$th rollout. The geometric planner or virtual tube determines where motion is feasible, while VIP learns how fast the system should traverse each location.
	
	Two rollout modes are distinguished. In the \emph{model-in-the-loop offline mode}, a simulator or prediction model generates the energy profile $V_k(l)$ before physical execution. The model is used only as a rollout generator; the update itself remains the model-free law in (\ref{approximateILlaw}). This mode is suitable when repeated physical trials are expensive.
	
	In the \emph{robot-in-the-loop online mode}, the physical robot or swarm completes one traversal, records $V_k(l)$ and $T_k$, and updates $v_{\mathrm{t},k+1}(l)$ before the next repeated traversal. Thus, ``online'' refers to learning from successive real executions rather than to an adaptation occurring continuously within one traversal.
	
	Spatial recording aligns rollouts of different durations at the same path locations. At each of the $n$ discretization points, the model-free update evaluates one scalar function $g(\cdot)$ and performs one algebraic correction. The same implementation is therefore retained for path tracking error, swarm distribution error, or another energy satisfying \emph{Section~\ref{problem}}.
	
	\section{Single-robot realization and model-free implementation} \label{singlerealization}
	The single-robot model, energy, and composite controller have already been introduced in Subsection~\ref{singleIL}. This section only provides the execution condition and the resulting model-free implementation; it does not repeat the unified system derivation.
	
	\subsection{Energy system verification}

	Under \emph{Assumption~\ref{assumvtrack}}, the path-normal/tangent orthogonality yields the energy model (\ref{dotVsatleq}), which is the single-robot instance of the unified system (\ref{Vmodel2}) with
	\begin{equation}
		p=\frac{3}{2},
		\qquad q=\frac{1}{2},
		\qquad o=0.
	\end{equation}
	The complete derivative calculation is given in \emph{Appendix~\ref{app:singleverification}}. The physical interpretation is immediate: increasing $v_{\mathrm{t}}$ can reduce $\kappa_v$ under saturation and therefore weaken path-error correction.
	
	\subsection{Model-free VIP law}
	For the single-robot task, let $V_k=V_{\mathrm{s},k}$ in the generic law of Subsection~\ref{freeIL}. The traversal profile is updated by
	\begin{equation}
		v_{\mathrm{t},k+1}(l)
		=v_{\mathrm{t},k}(l)-b_3g\left(k_{\mathrm{e}}V_{\mathrm{s},k}(l)\right).
		\label{singlemodellaw}
	\end{equation}
	Here $k_{\mathrm{e}}$ is the scaling coefficient defined in Subsection~\ref{freeIL}, whereas $k_{\mathrm{p}}$ remains the single-robot regulation gain in (\ref{singlevc}).
	When the measured path energy is below the admissible band, the update increases traversal speed; when the energy exceeds the band, it decreases traversal speed. The law uses the measured $V_{\mathrm{s},k}(l)$ and does not require $\tau$, $\lambda_{\mathrm{s}}$, or $\varepsilon_{\mathrm{s}}$ for online implementation.
	
	A derivative-enhanced implementation may use $g(\zeta_{\mathrm{d},k})$, where $\zeta_{\mathrm{d},k}=k_{\mathrm{e}}V_{\mathrm{s},k}+k_{\mathrm{d}}\dot V_{\mathrm{s},k}$, and $k_{\mathrm{d}}>0$. \emph{Proposition~\ref{prop:saturation_derivatives}} gives the explicit local saturation sensitivity
	\begin{equation}
		\gamma_1
		\triangleq
		\frac{\partial\left(\kappa_v v_{\mathrm{n}}^{3/2}\right)}{\partial v_{\mathrm{t}}}
		=-\frac{\kappa_vv_{\mathrm{n}}^{3/2}v_{\mathrm{t}}}
		{v_{\mathrm{n}}^2+v_{\mathrm{t}}^2}<0
		\label{gamma6definition}
	\end{equation}
	in the active-saturation region with $v_{\mathrm{n}},v_{\mathrm{t}}>0$. The following result gives a compact boundedness statement; the detailed mean-value expansion is retained in the appendix.
	
	\begin{theorem} \label{contractive mapping}
		Suppose that \emph{Assumption~\ref{assumvtrack}} holds, the controller is given by (\ref{singlevc}), and $g(\zeta)\geq k_g\zeta-\zeta_{\mathrm{th}}$ for some $k_g,\zeta_{\mathrm{th}}>0$. If the derivative-enhanced update satisfies, uniformly on the considered compact active-saturation set,
		\begin{equation}
			-1<b_3k_gk_{\mathrm{d}}\lambda_{\mathrm{s}}\gamma_1<0,
		\end{equation}
		then $v_{\mathrm{t},k}$ is uniformly ultimately bounded as $k\rightarrow\infty$.
	\end{theorem}
	\begin{proof}
		See \emph{Appendix~\ref{prooftheoremIL}}.
	\end{proof}
	
	\begin{remark}
		The VIP update acts on the desired velocity profile and does not require the full robot dynamics. A dynamic platform is included whenever its low-level controller guarantees \emph{Assumption~\ref{assumvtrack}}. \emph{Appendix~\ref{app:singleverification}} gives a standard Lagrangian-system controller as one sufficient construction. In the offline mode, a dynamic model can generate prediction rollouts; in the online mode, the same update is driven directly by recorded physical rollouts. Thus, the dynamic extension is treated as an implementation condition rather than as another independent VIP formulation.
	\end{remark}
	
	\section{Robotic-swarm model-free implementation and microscopic realization} \label{swarmrealization}
	The macroscopic density system, energy, and composite velocity field have been established in \emph{Subsection~\ref{swarmIL}}, while the KDE realization and desired-density construction are detailed in \emph{Appendix~\ref{app:swarmverification}}. This section therefore retains only the model-free traversal update and the connection between the macroscopic fields and the microscopic distributed controller.
	
	\subsection{Model-free VIP law}
	Using $V_k=V_{\mathrm{m},k}$ in the generic update of Subsection~\ref{freeIL} gives
	\begin{equation}
		v_{\mathrm{t},k+1}(l)
		=v_{\mathrm{t},k}(l)-b_3g\left(k_{\mathrm{e}}V_{\mathrm{m},k}(l)\right).
		\label{swarmmodellaw}
	\end{equation}
	The same $k_{\mathrm{e}}$ notation is used to scale the swarm energy $V_{\mathrm{m}}$ to the admissible band; its numerical value may be selected according to the swarm-density tolerance.
	The parameter mapping in \emph{Proposition~\ref{prop:commonenergy}} is
	\begin{equation}
		p=2,
		\qquad q=1,
		\qquad o=1.
	\end{equation}
	The update uses the density-error energy $V_{\mathrm{m},k}$ computed from the KDE density in (\ref{KDE}) and does not require online identification of $k_{\rho}$, $\beta_{\rho}$, or $\partial\rho_{\mathrm{d}}/\partial t$.
	
	\subsection{Physical interpretation}
	The low-altitude air city transport model in \cite{safadi2023macroscopic} connects microscopic robot interactions with a macroscopic fundamental diagram relating flow, density, and speed. The diagram indicates that increasing swarm density generally reduces achievable traversal speed. Therefore, the swarm exhibits the same competition represented by (\ref{Vmodel2}): the common traversal command should increase when the density error is small and decrease when additional distribution-regulation authority is needed.
	
	\begin{figure}[htbp]
		\centering
		\includegraphics[width=2.2in]{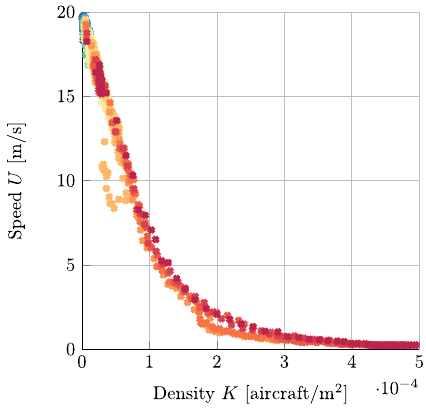}
		\caption{Qualitative inverse relationship between swarm density and traversal speed revealed by the macroscopic fundamental diagram (\cite{safadi2023macroscopic}).}
		\label{spDT}
	\end{figure}
	
	\subsection{Relationship with a distributed virtual tube controller} \label{subsec:classicVTC}
	A distributed vector-field controller for guiding a robotic swarm through a curved virtual tube is developed in \cite{quan2023distributed,gao2025distributed}. Its robot-avoidance and tube-keeping components are
	\begin{equation}
		\begin{split}
			\mathbf{u}_{\mathrm{rep},i}
			&=-\sum_{j\in\mathcal{N}}
			\frac{\partial V_{\mathrm{rep},ij}}{\partial\Vert\widetilde{\mathbf{p}}_{ij}\Vert}
			\frac{\widetilde{\mathbf{p}}_{ij}}{\Vert\widetilde{\mathbf{p}}_{ij}\Vert},\\
			\mathbf{u}_{\mathrm{t},i}
			&=\frac{\partial V_{\mathrm{t},i}}{\partial d_{\mathrm{t},i}}
			\frac{\mathbf{e}_{\mathrm{p}}(\mathbf{p}_i)}
			{\Vert\mathbf{e}_{\mathrm{p}}(\mathbf{p}_i)\Vert},
		\end{split}
	\end{equation}
	where $\widetilde{\mathbf{p}}_{ij}=\mathbf{p}_i-\mathbf{p}_j$, $\mathcal{N}$ is the neighbor set, and $d_{\mathrm{t},i}$ is the distance margin to the tube boundary. The forward component is $\mathbf{u}_{1,i}=v_{\mathrm{m}}\mathbf{t}_{\mathrm{c}}$, which is parallel to the macroscopic traversal term $\mathbf{v}^{\prime}_{\mathrm{t}}$ in (\ref{dswarmc}).
	
	Differentiating (\ref{KDE}) gives
	\begin{equation}
		\nabla_{\mathbf{p}}\rho(\mathbf{p})
		=-\frac{1}{Mh^5}
		\sum_{j\in\mathcal{N}}
		K\left(\frac{\mathbf{p}-\mathbf{p}_j(t)}{h}\right)
		(\mathbf{p}-\mathbf{p}_j).
	\end{equation}
	Thus, at robot $i$, the diffusion component $\mathbf{v}^{\prime}_{\phi}$ has the same local repulsive structure as $\mathbf{u}_{\mathrm{rep},i}$. The macro--micro correspondence is therefore: (i) $\mathbf{v}^{\prime}_{\mathrm{t}}$ determines the learned common forward speed, corresponding to $\mathbf{u}_{1,i}$; (ii) $\mathbf{v}^{\prime}_{\phi}$ regulates density, corresponding to inter-robot dispersion; and (iii) $\mathbf{u}_{\mathrm{t},i}$ enforces the local tube boundary. VIP modifies only the common traversal magnitude, while the established microscopic controller maintains local safety and distribution.
	
	\section{Simulations and comparisons} \label{simcom}
	To demonstrate the effectiveness and capabilities of the proposed method, this section presents simulations and comparisons for the two energy systems introduced in \emph{Subsections~\ref{singleIL}} and \emph{\ref{swarmIL}}. The corresponding implementation details are given in \emph{Sections~\ref{singlerealization}} and \emph{\ref{swarmrealization}}. The unknown-environment scenarios in \emph{Subsection~\ref{singleNav}} evaluate the single-robot case, whereas \emph{Subsection~\ref{swarmVTNav}} evaluates swarm navigation in a virtual tube. Together, they assess computational efficiency, traversal-time improvement, and bounded error across different operational scales.
	
	\subsection{Single robot autonomous offline planning in unknown environments} \label{singleNav}
	\subsubsection{Simulations in complex environments}
	
	Simulations are conducted in two random forest environments, denoted as Map 1 and Map 2. The proposed method is evaluated from the start point to the goal point on a workstation equipped with an Intel Core i7-12700F CPU and 16 GB RAM. The algorithm is implemented in C++. Each map has a size of $40 \mathrm{m}\times 20 \mathrm{m}\times 5 \mathrm{m}$, and the maximum speed is set to $5 \mathrm{m/s}$. Tube-RRT* (\cite{mao2025tube}) is first used to generate a feasible path or virtual tube, and VIP is then applied to learn the traversal-speed profile along the spatial coordinate.
	
\begin{figure}[htbp]
		\centering
		\includegraphics[width=3.1in]
		{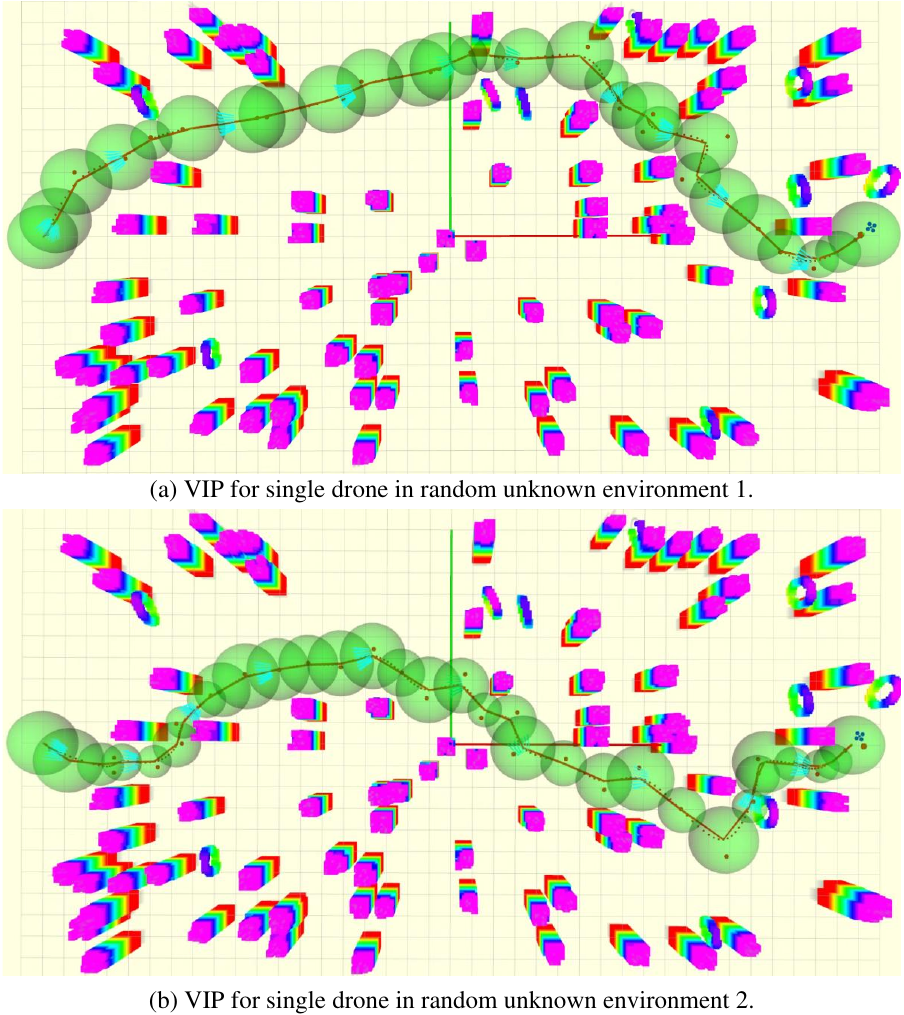}
		\caption{VIP-based single-drone navigation in two random forest environments.}
		\label{dplan}
	\end{figure}

	The simulation results are shown in Fig.~\ref{dplan}. In both maps, the robot successfully navigates through obstacle-dense forest-like environments. The generated path or tube provides spatial feasibility, while VIP further adjusts the traversal command according to the local geometric and tracking requirements. In relatively open and smooth regions, the learned traversal command increases the speed to reduce traversal time. In narrow passages or turning regions, the command becomes more conservative, leaving more control authority for path convergence and safety regulation. This behavior is consistent with the energy--traversal trade-off discussed in Subsection~\ref{singleIL}: rapid traversal is encouraged when the tracking energy remains small, whereas deceleration is introduced when large deviation may degrade the stability of path following.
	
	The computational results further verify the lightweight property of the proposed method. The planning time is $0.28 \mathrm{ms}$ for 50 waypoints and $0.52 \mathrm{ms}$ for 100 waypoints, indicating an approximately linear increase with the number of path points. The average speed reaches $2.98 \mathrm{m/s}$ under the maximum-speed constraint of $5 \mathrm{m/s}$. These results are consistent with the complexity analysis in \emph{Section~\ref{Complexity}}: VIP updates the traversal profile along the spatial coordinate and avoids solving a long-horizon nonlinear optimization problem at each planning step, making it suitable for real-time navigation with dense waypoint information. The related code is open-source and publicly available\footnote{\url{https://github.com/lyushuli/VIP}}.
	
	\subsubsection{Comparisons} \label{compareSingleNav}
\begin{figure*}[htbp]
		\centering
		\includegraphics[width=7in]
		{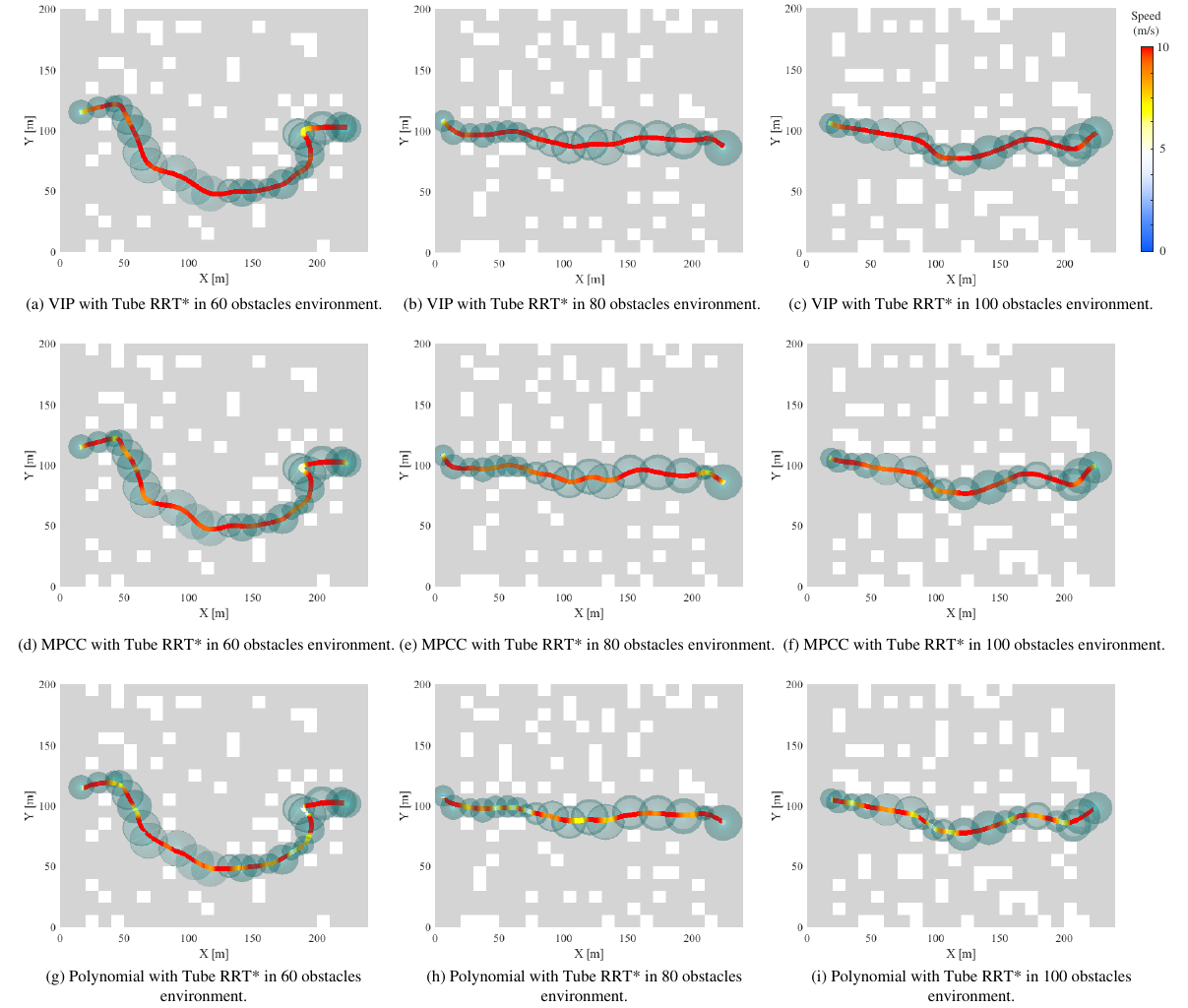}
		\caption{Numerical comparisons of single robot autonomous navigation in unknown environments.}
		\label{singletube}
	\end{figure*}

	After obtaining the smooth path from a path planner, comparison simulations are conducted utilizing various methods in this subsection. In the obstacle environments shown in Fig. \ref{singletube}, Tube-RRT* is used to plan the virtual tube. Considering both the path length and the volume of openings, the path found by Tube-RRT* algorithm pass through large openings, which are suitable for constructing the virtual tube and for safe navigation. The comparison methods include MPCC (\cite{romero2022model}) and the polynomial technique with the combination of time-optimal time allocation (\cite{teissing2024real}) and minimum snap trajectory (\cite{mellinger2011minimum}). All methods are implemented in MATLAB, and the simulations are run on Intel(R) Core(TM) i9-10900 CPU @ 2.80 GHz. The optimization problems with MPCC and polynomial are solved with a QP solver Quadprog. 
	
	The comparisons results are shown in Fig. \ref{singletube} and Tab. \ref{tab:performance}, unknown scenarios with 60, 80, and 100 random obstacles in a 230m $\times$ 200m $\times$ 30m environment are generated, and the maximum speed is set to 10m/s. For traversal time, the proposed VIP demonstrates consistent competitiveness, showing 1.31-1.57\% improvements over MPCC and Polynomial methods in the 60-obstacle scenario. While its advantage varies across scenarios (ranging from -1.31\% to 1.37\% compared to MPCC), it maintains superior performance against Polynomial (up to 6.80\% improvement). More strikingly, VIP exhibits exceptional performance in computational efficiency, achieving 47.4-87.7\% reductions in average replanning time across all test scenarios compared to both benchmark algorithms. The Polynomial method's higher replanning time stems from its computationally intensive time-allocation optimization and QP solving. This suggests VIP's particular strength in real-time applications where rapid replanning is crucial. Especially with the expansion of the number of planning points and the scope, VIP will better demonstrate its advantage in linear complexity. These results collectively demonstrate that VIP offers a favorable balance between path efficiency (traversal time) and computational performance (replanning time), making it particularly suitable for dynamic environments requiring frequent trajectory updates.
	
	\begin{table*}[htbp]
		\centering
		\caption{Performance comparison of different algorithms (VIP results in bold with relative change)}
		\label{tab:performance}
		\begin{tabular}{ccccccccc}
			\toprule
			\multirow{2}{*}{ \begin{tabular}{@{}c@{}} \textbf{Obstacle} \\ \textbf{Number}\end{tabular} } & 
			\multicolumn{4}{c}{\textbf{Traversal Time (s)}} & 
			\multicolumn{4}{c}{\textbf{Avg. Replanning Time (ms)}} \\
			\cmidrule(lr){2-5} \cmidrule(lr){6-9}
			& \textbf{MPCC} & \textbf{Polynomial} & \textbf{VIP} & \textbf{vs. Others} & \textbf{MPCC} & \textbf{Polynomial} & \textbf{VIP} & \textbf{vs. Others} \\
			\midrule
			60 & 34.30 & 34.39 & \textbf{33.85} & \begin{tabular}{@{}c@{}}-1.31\% vs. MPCC \\ -1.57\% vs. Poly\end{tabular} & 5.32 & 17.66 & \textbf{2.37} & \begin{tabular}{@{}c@{}}-55.5\% vs. MPCC \\ -86.6\% vs. Poly\end{tabular}\\
			80 & 25.55 & 27.79 & \textbf{25.90} & \begin{tabular}{@{}c@{}}+1.37\% vs. MPCC \\ -6.80\% vs. Poly\end{tabular} & 4.98 & 21.32 & \textbf{2.62} & \begin{tabular}{@{}c@{}}-47.4\% vs. MPCC \\ -87.7\% vs. Poly\end{tabular}\\
			100 & 25.05 & 26.77 & \textbf{25.25} & \begin{tabular}{@{}c@{}}+0.80\% vs. MPCC \\ -5.68\% vs. Poly\end{tabular} & 5.44 & 19.81 & \textbf{2.45} & \begin{tabular}{@{}c@{}}-55.0\% vs. MPCC \\ -87.6\% vs. Poly\end{tabular}\\
			\bottomrule
		\end{tabular}
	\end{table*}
	
	\subsection{Swarm online learning in the virtual tube} \label{swarmVTNav}
	All swarm simulations in this subsection are implemented in MATLAB and executed on the same Intel(R) Core(TM) i9-10900 CPU @ 2.80 GHz platform. The purpose is to verify whether the proposed learning strategy can be extended from a single robot to a robotic swarm without optimizing the trajectory of each robot individually. In the swarm setting, VIP regulates the macroscopic traversal speed along the virtual tube, while the virtual-tube controller maintains the microscopic distribution and boundary constraints.
	
	\subsubsection{Enhance effectiveness in 2-D virtual tube}
	In this simulation, VIP is used to improve the planning efficiency of a robotic swarm controlled by (\cite{quan2023distributed, gao2025distributed}), as introduced in \emph{Section \ref{subsec:classicVTC}}. The proposed method uses the recorded data from previous iterations to refine the swarm traversal command. The virtual tube is generated in a $110 \mathrm{m}\times 20 \mathrm{m}$ environment, the swarm consists of 30 homogeneous agents, and the maximum speed of each robot is set to $6 \mathrm{m/s}$.
	
\begin{figure*}[htbp]
		\centering
		\includegraphics[width=7in]
		{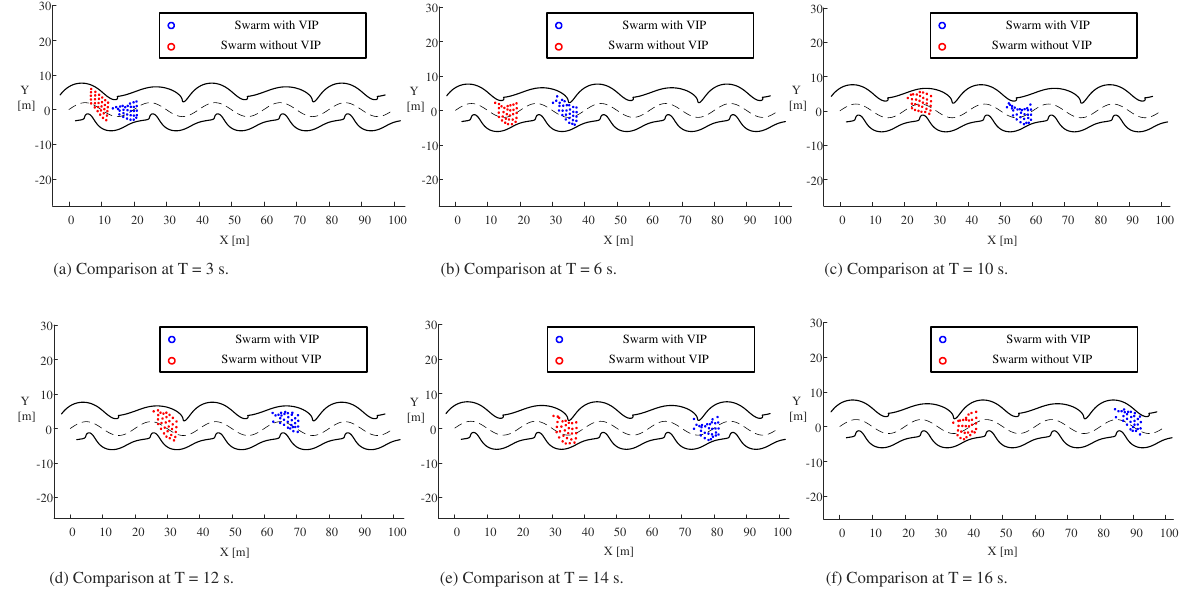}
		\caption{Numerical comparison of robot swarm navigation in a virtual tube.}
		\label{2dswarmtube}
	\end{figure*}

	The comparison results are shown in Fig. \ref{2dswarmtube}. Without VIP, the swarm follows the virtual tube with a relatively conservative traversal profile. With VIP, the swarm moves faster along the tube while maintaining a bounded spatial distribution. The snapshots at different time instants show that the learned command increases forward progress without destroying the tube constraint. This indicates that the macroscopic traversal-speed learning does not conflict with the microscopic distribution regulation.
	
\begin{figure}[htbp]
		\centering
		\includegraphics[width=2.3in]
		{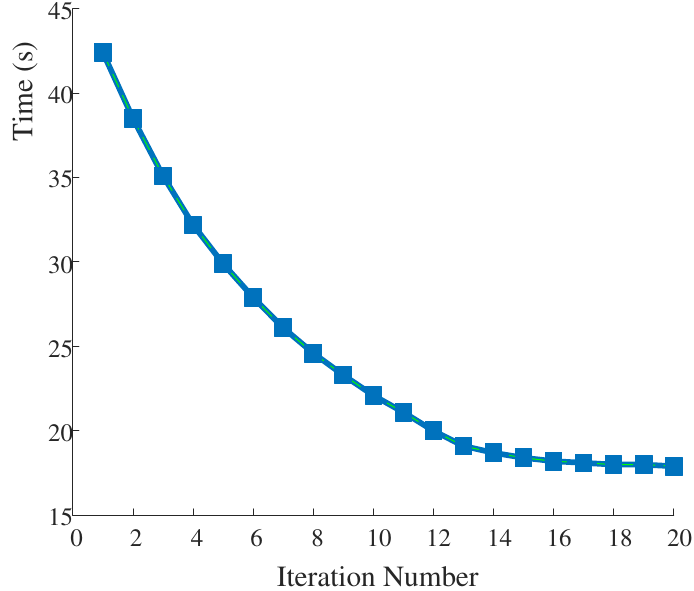}
		\caption{Passing-time descent process of swarm navigation under VIP. The curve shows that the learned common traversal command progressively reduces traversal time and then settles as the model-free IL update converges.}
		\label{swarmILT}
	\end{figure}

	The time descent process is shown in Fig. \ref{swarmILT}. The traversal time decreases over learning iterations and then tends to stabilize, consistent with the convergence behavior of the model-free learning law. From the density-based formulation in \emph{Subsection \ref{swarmIL}}, the swarm energy describes the mismatch between the current density and the desired distribution. VIP improves efficiency by adjusting the common traversal component, while the distribution-related component continues to regulate the swarm configuration. Therefore, the simulation illustrates the intended division of roles: the virtual tube ensures spatial safety and maintains distribution, whereas VIP learns to traverse the tube more efficiently.
	
	\subsubsection{Enhance effectiveness in 3-D complex environment}
\begin{figure*}[htbp]
		\centering
		\includegraphics[width=7in]
		{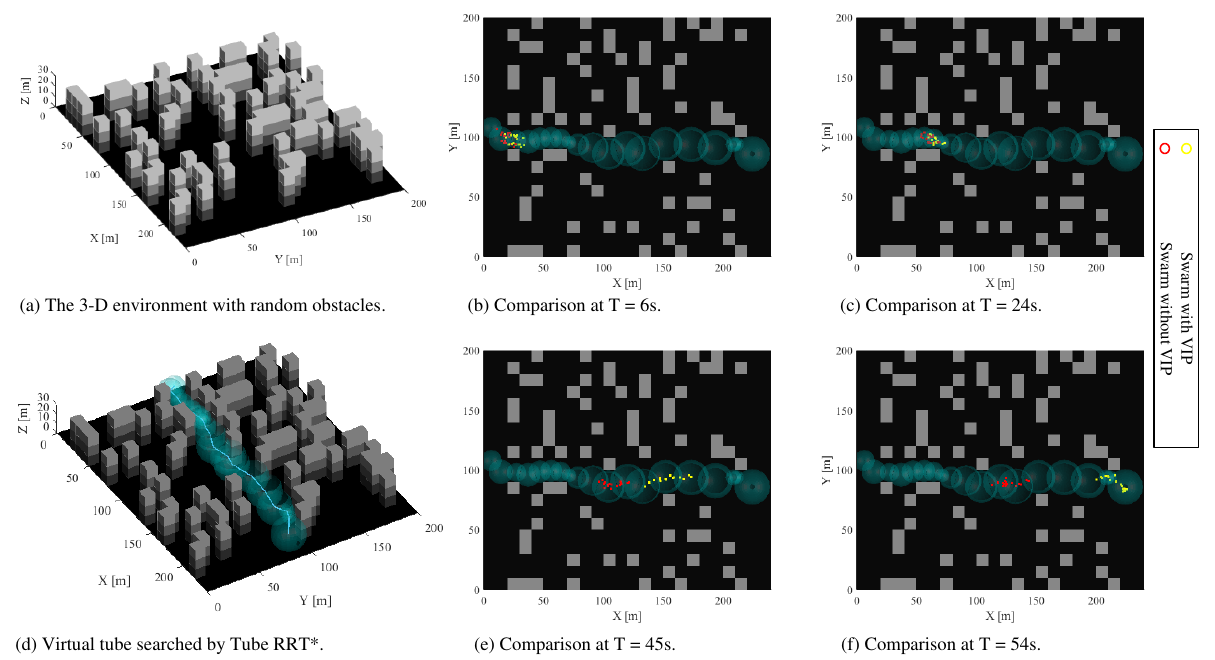}
		\caption{Three-dimensional swarm navigation in an obstacle environment. Tube-RRT* constructs a virtual tube through random obstacles, and VIP improves swarm traversal efficiency while the virtual-tube controller preserves safety and distribution regulation.}
		\label{3dswarm}
	\end{figure*}

	The proposed framework is further evaluated in a 3-D obstacle environment. The distributed controller (\cite{quan2023distributed, gao2025distributed}) is implemented in a 3-D form, and VIP is used to learn the traversal-speed profile of the swarm. As shown in Fig. \ref{3dswarm}(a), random obstacles are generated in a $230 \mathrm{m}\times 200 \mathrm{m}\times 30 \mathrm{m}$ environment, which is consistent with the single-robot comparison setting in \emph{Section \ref{compareSingleNav}}. Tube-RRT* is used for tube planning, as illustrated in Fig. \ref{3dswarm}(d). The swarm size is set to 20, and the maximum speed of each robot is set to $8 \mathrm{m/s}$.
	
	The results in Fig. \ref{3dswarm}(b)--(c) and Fig. \ref{3dswarm}(e)--(f) show that the swarm can pass through the planned 3-D virtual tube while maintaining a feasible collective distribution. The learned profile improves the overall traversal speed, while the tube controller confines the agents inside the safe corridor. Compared with the 2-D case, the 3-D environment introduces more complex spatial curvature and obstacle distribution. The successful execution in this setting supports the scalability of the VIP framework from planar tube navigation to spatial tube navigation.
	
	The single-robot and swarm simulations together demonstrate the role of VIP as a general learning layer for navigation efficiency improvement. For single robots, the energy term is defined by tracking deviation. For robotic swarms, the energy term is defined by density-distribution error. Although the physical meanings of the energy functions are different, both cases share the same learning logic: the traversal command is increased where the energy remains acceptable and reduced where rapid traversal may lead to large deviations. This consistency confirms the unified formulation developed in the theoretical sections.
	
	\section{Experiments} \label{exp}
	To further validate the deployability of the proposed VIP framework, real-world experiments are conducted on three robotic platforms and scenarios: a LiDAR-based quadcopter flying in an unknown outdoor environment, a lifting-wing quadcopter, and a multi-robot swarm flying in a virtual tube. The experiments are designed to examine whether the learned spatial traversal command can be executed on physical platforms under onboard sensing, localization, actuation, communication, and environmental constraints.
	
	The simulation results in \emph{Section \ref{simcom}} mainly verify the computational and algorithmic properties of VIP. The experiments in this section further examine the feasibility of its implementation. In all experiments, VIP is used as a planning or traversal-command learning layer. This setting is consistent with the structure in Fig. \ref{structure}, where the learning term is updated from recorded process data and then applied to the next execution.
	
	\subsection{Offline flight in unknown environment}
	\subsubsection{Experiment settings}
\begin{figure}[htbp]
		\centering
		\includegraphics[width=3.1in]{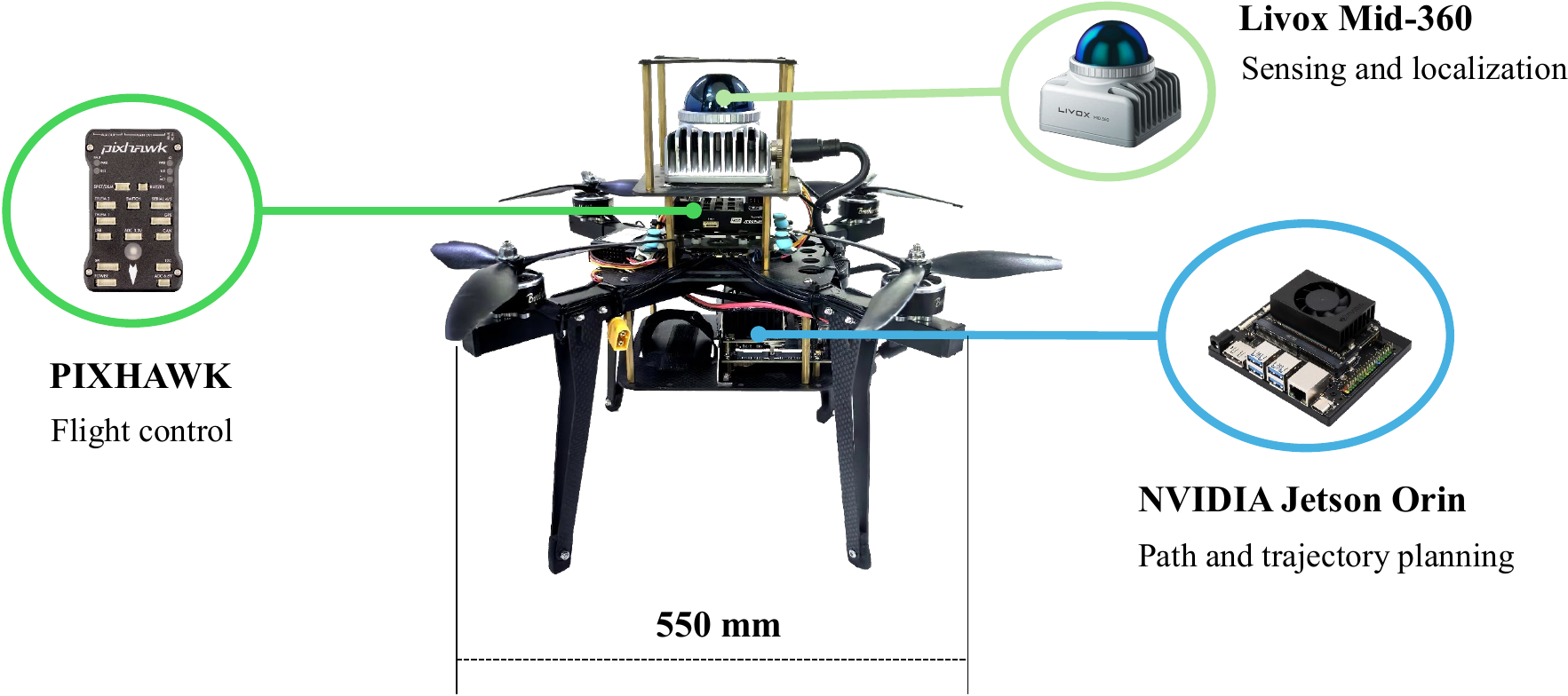}
		\caption{Experimental platform for LiDAR-based drone navigation.}
		\label{quad_exp}
	\end{figure}

	To validate the performance of the proposed method in an unknown outdoor environment, a LiDAR-based quadcopter platform is used, as shown in Fig. \ref{quad_exp}. The platform is equipped with an onboard NVIDIA Jetson Orin computer and a Livox Mid-360 LiDAR. The LiDAR point cloud is used for mapping and localization with Fast-LIO (\cite{xu2022fast}). Based on the reconstructed map, a collision-free path is generated by Tube RRT* (\cite{mao2025tube}) and then transformed into a spatial reference for VIP.
	
	In this experiment, the perception, planning, and control modules run in an integrated onboard pipeline. The LiDAR and Fast-LIO modules provide the local map and pose estimation. The planner generates the spatial path in the reconstructed environment. VIP then determines the traversal command along the path. This setting is closer to practical autonomous navigation than to offline simulation because it involves sensing noise, localization drift, computational delays, and actuation constraints. Furthermore, a replanning mechanism is implemented to adapt to long-distance, unknown environments, and the algorithm's real-time performance and low computational complexity are verified.
	
	\subsubsection{Experimental results}
	\begin{figure*}[htbp]
		\centering
		\includegraphics[width=6.8in]{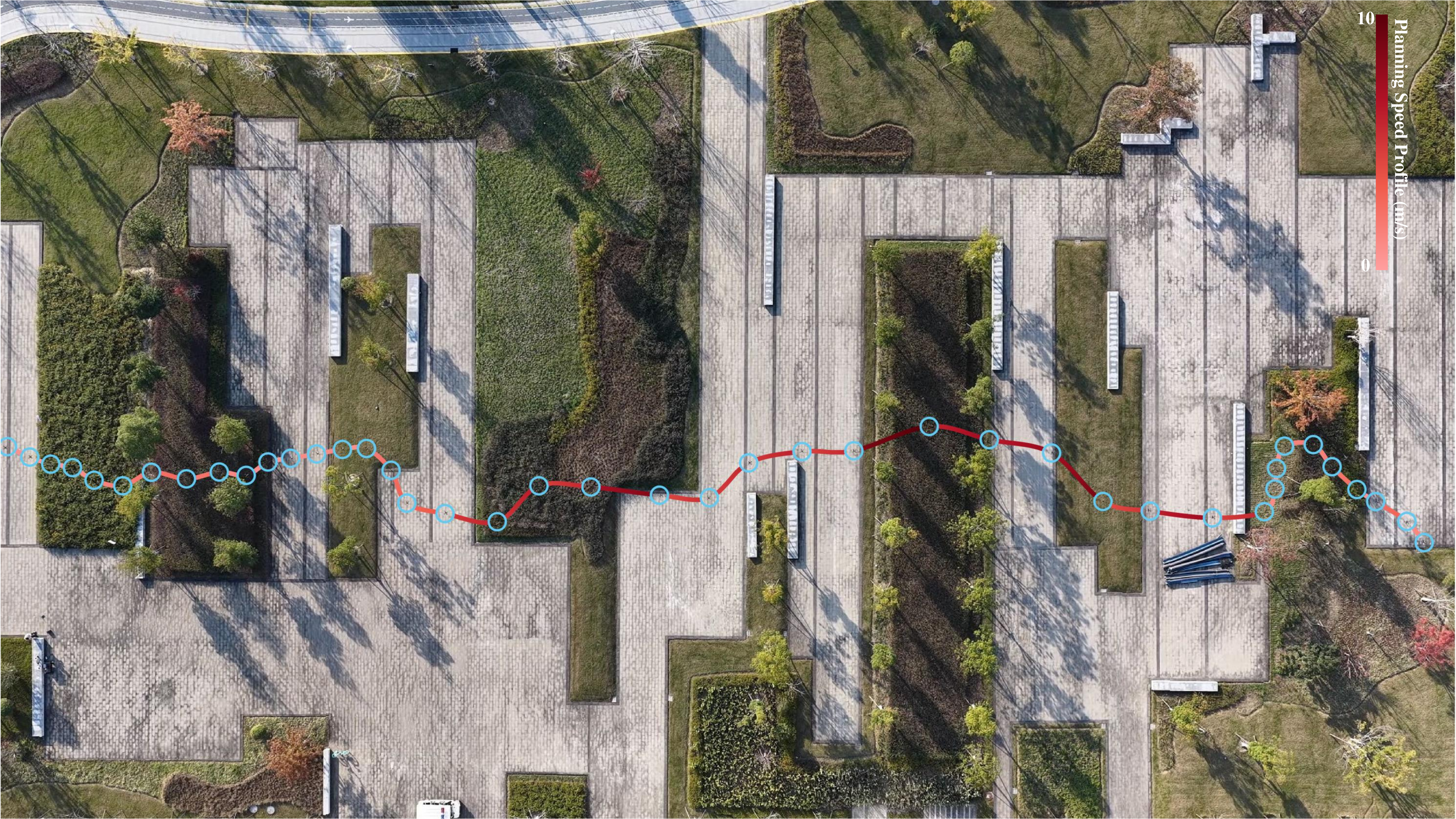}
		\caption{Aerial view of single-drone fast navigation in an unknown outdoor environment. The recorded flight trajectory covers approximately $150 \mathrm{m}$ and verifies real-flight execution of the VIP traversal-speed profile under onboard perception and localization.}
		\label{single_nav}
	\end{figure*}
	
\begin{figure*}[htbp]
		\centering
		\includegraphics[width=6.9in]{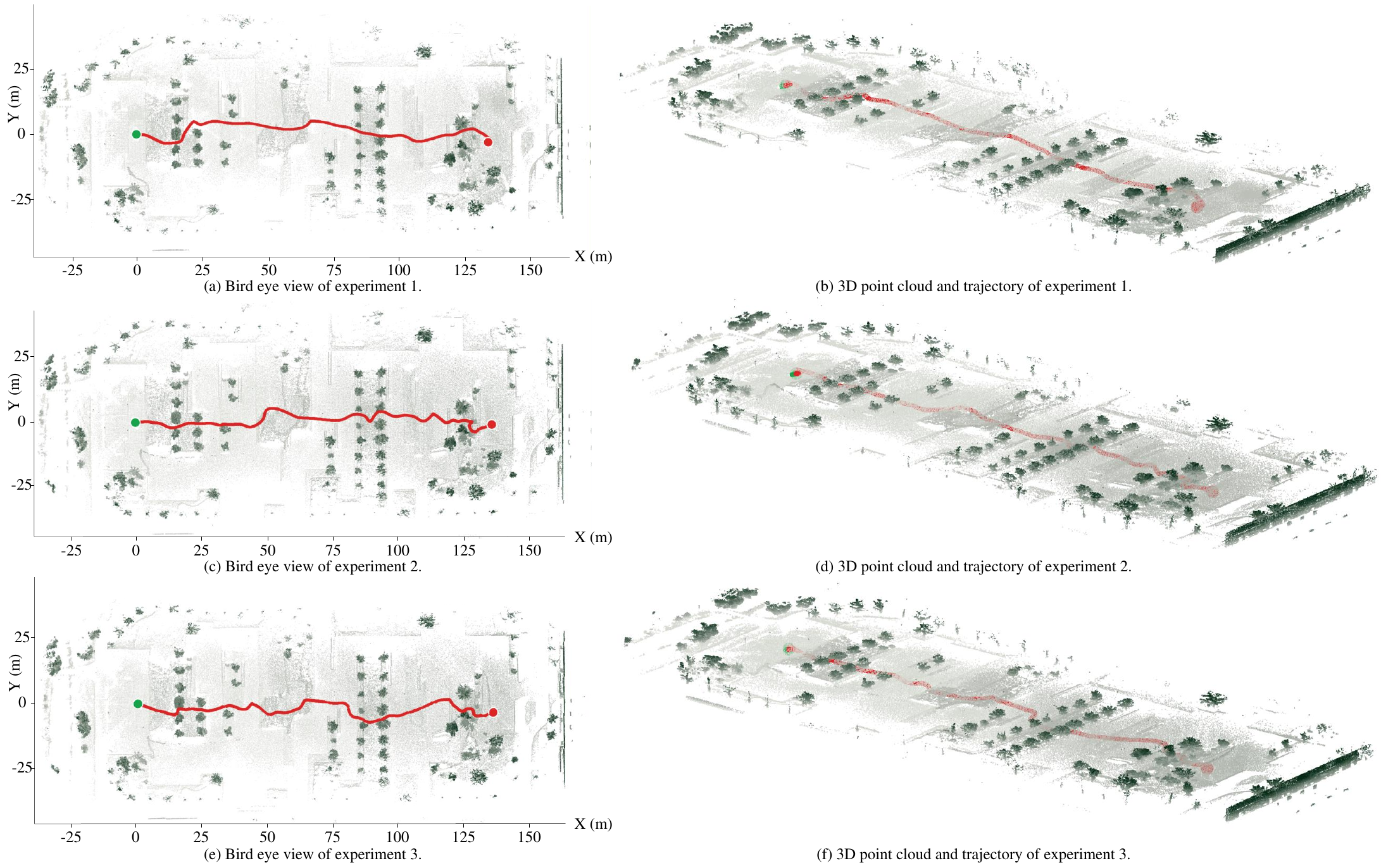}
		\caption{Repeated single-drone navigation trials with reconstructed point clouds and recorded trajectories. The bird's-eye and 3-D views show repeatable fast traversal in an unknown outdoor environment using onboard LiDAR perception and VIP-based traversal-command learning.}
		\label{single_cloud}
	\end{figure*}

	The outdoor navigation result is shown in Fig.~\ref{single_nav}, where the drone traverses an unknown outdoor field with a total flight distance of approximately $150 \mathrm{m}$. Fig.~\ref{single_cloud} presents three repeated experiments with the reconstructed point clouds and the corresponding recorded trajectories. The environment contains trees, buildings, and open corridors, which impose practical requirements on perception reliability, planning quality, replanning frequency, and trajectory execution.
	
	The experimental results show that the drone can pass through the outdoor environments while maintaining feasible and continuous motion. During the flight tests, the drone achieves a maximum speed of about $10 \mathrm{m/s}$, demonstrating that the proposed framework can support fast navigation on a real aerial platform. This result is not only related to the maximum velocity command, but also to the compatibility between the planned spatial path, the learned traversal-speed profile, and the onboard tracking capability. If the traversal command is too aggressive in narrow or highly curved regions, the accumulated tracking error may lead to unstable path following or unsafe motion. The successful high-speed flight therefore indicates that VIP can adjust the traversal command according to the local navigation difficulty rather than simply increasing speed along the entire path.
	
	In the onboard implementation, the RRT*-based planner continuously performs accumulated sampling and local sampling for path replanning, while VIP rapidly updates the traversal-speed profile along the spatial coordinate. In this pipeline, RRT* mainly provides spatial feasibility by searching for a collision-free path in the reconstructed environment, whereas VIP further improves traversal efficiency by learning how fast the drone should move along the generated path. Therefore, the two modules have different roles: the sampling-based planner determines where to fly, and VIP determines how fast to traverse the resulting spatial path under the energy-traversal trade-off. With the onboard computing platform, VIP achieves a computation time of $0.93 \mathrm{ms}$ for 100 path points, thereby verifying its lightweight nature and real-time responsiveness.
	
	The computation result also supports the complexity analysis of the proposed method. Since VIP updates the traversal-speed profile directly along the spatial coordinate, its computation mainly scales with the number of path points rather than with a long prediction horizon or a large nonlinear programming problem. The ability to replan within $1 \mathrm{ms}$ shows that VIP can handle refined and dense spatial path information generated by the onboard planner. This is important for unknown outdoor navigation, where the path may be frequently updated due to newly observed obstacles, local map changes, and localization uncertainty.
	
	These results indicate that VIP can be executed efficiently together with onboard LiDAR perception and real-time localization. Compared with the simulations in \emph{Section~\ref{singleNav}}, this experiment further verifies that the method can tolerate practical sensing noise, localization error, replanning delay, and onboard computational constraints. The real-flight result also shows that the learned traversal command remains executable after being integrated into the complete perception--planning--control pipeline, rather than being valid only in offline simulation.
	
	The repeated trials in Fig.~\ref{single_cloud} further demonstrate the repeatability and implementation robustness of the proposed method. The spatial planning is generated in real time from the reconstructed environment, and the traversal command is updated according to the real-time information. This supports the use of VIP as a lightweight traversal-command learning layer for fast autonomous navigation in unknown outdoor scenarios.
	
	\subsection{Application on online learning with a Lifting-Wing Quadcopter}
	
	\subsubsection{Experimental Settings}
	
\begin{figure}[htbp]
		\centering
		\includegraphics[width=2.5in]{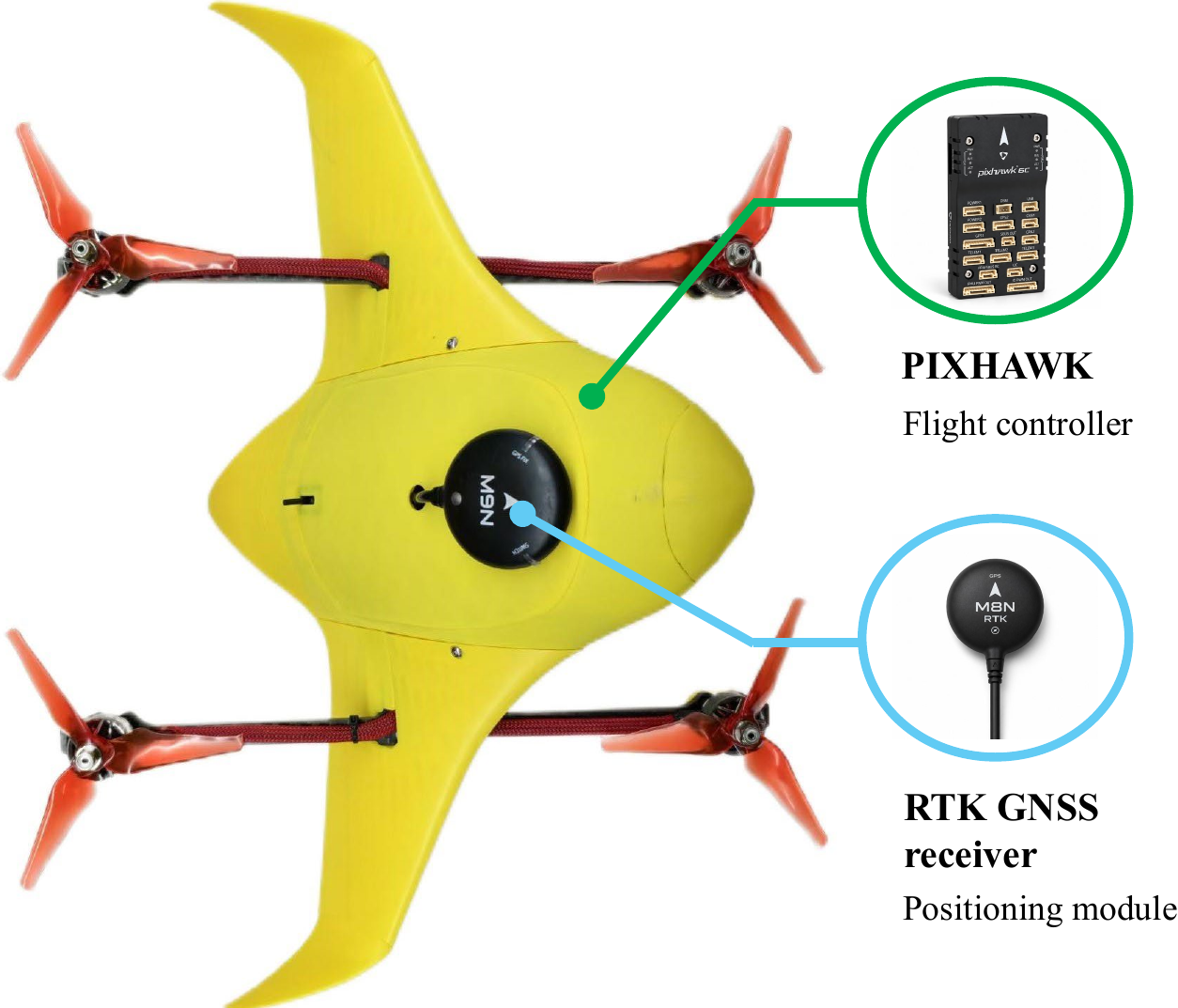}
		\caption{Experimental platform of the lifting-wing quadcopter.}
		\label{lift}
	\end{figure}

	The experimental platform is a lifting-wing quadcopter, as shown in Fig.~\ref{lift}. The control and allocation modules are designed in Simulink, automatically converted into C/C++ code, and integrated into the Pixhawk/PX4 autopilot through the Pixhawk Support Package toolbox. Different from a conventional multirotor, the lifting-wing quadcopter is influenced by both rotor-generated thrust and lifting-wing aerodynamic effects during forward flight (\cite{quan2025lifting}). Therefore, this platform provides a practical test for evaluating whether the learned traversal command can be executed on a physical aerial system with more complex motion characteristics.
	
	The experiment is conducted in an outdoor environment, as shown in Fig.~\ref{liftwing}. The plant is commanded to follow a predefined spatial path consisting of long straight segments and large turning maneuvers. During flight, the VIP-generated traversal command is sent to the low-level flight controller, while the onboard controller tracks the corresponding velocity or position references. Before the outdoor flight test, hardware-in-the-loop simulation is used to perform iterative validation and obtain a converged traversal-speed profile. Then, two additional iterations are conducted in real flight to further adapt the learned command to the physical platform. This experiment is used to verify the continuous executability of the spatial-domain command generated by VIP on a real lifting-wing aerial platform.
	
	\subsubsection{Experimental Results}
	
\begin{figure*}[htbp]
		\centering
		\includegraphics[width=6.4in]{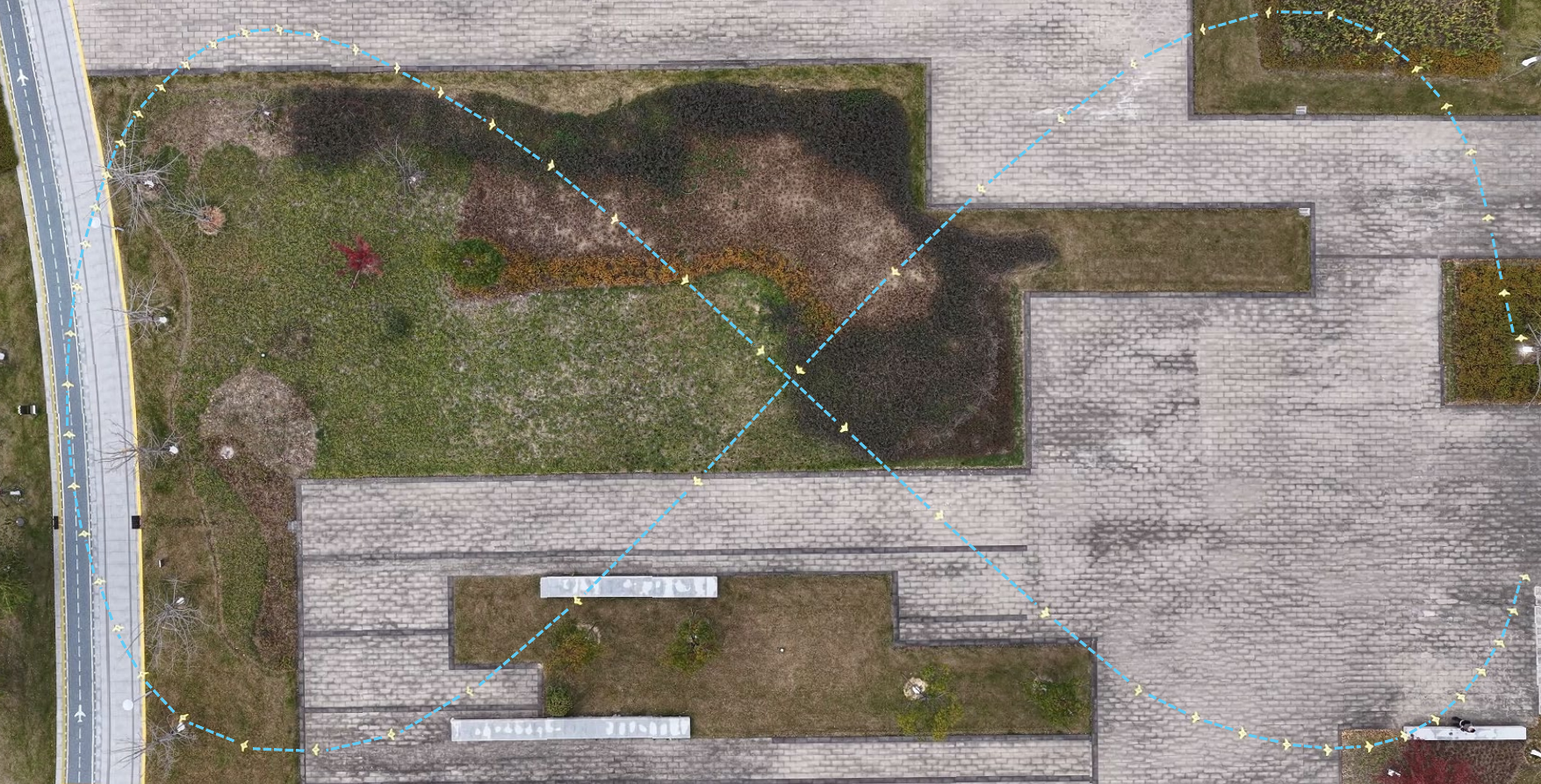}
		\caption{Outdoor flight trajectory of the lifting-wing quadcopter. The recorded trajectory demonstrates continuous execution of the VIP traversal command through straight, turning, and crossing regions on a physical platform.}
		\label{liftwing}
	\end{figure*}

\begin{figure}[htbp]
		\centering
		\includegraphics[width=3.3in]{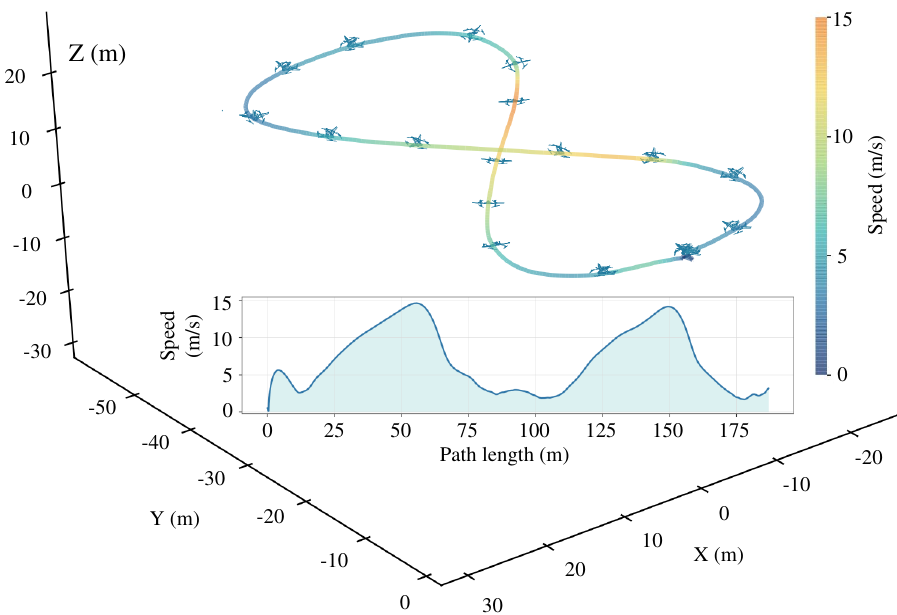}
		\caption{Visualization of the lifting-wing quadcopter flight experiment. The image sequence illustrates the real-flight execution process and verifies the compatibility between the learned traversal command and the onboard flight-control architecture.}
		\label{liftvis}
	\end{figure}

	The flight result is shown in Fig.~\ref{liftwing}, and the flight visualization is shown in Fig.~\ref{liftvis}. The recorded trajectory shows that the lifting-wing quadcopter successfully completes the planned path with continuous motion. This platform is selected to provide a more challenging validation case for the proposed model-free VIP. Since the model-free learning law updates the traversal command based on recorded execution data rather than relying on an explicit analytical model of the vehicle dynamics, the lifting-wing quadcopter is well-suited for testing its applicability to complex physical systems. The platform passes through straight, turning, and crossing regions without obvious interruption, indicating that the learned traversal command can be accepted by the onboard controller and transformed into feasible physical motion.
	
\begin{figure}[htbp]
		\centering
		\includegraphics[width=3.3in]{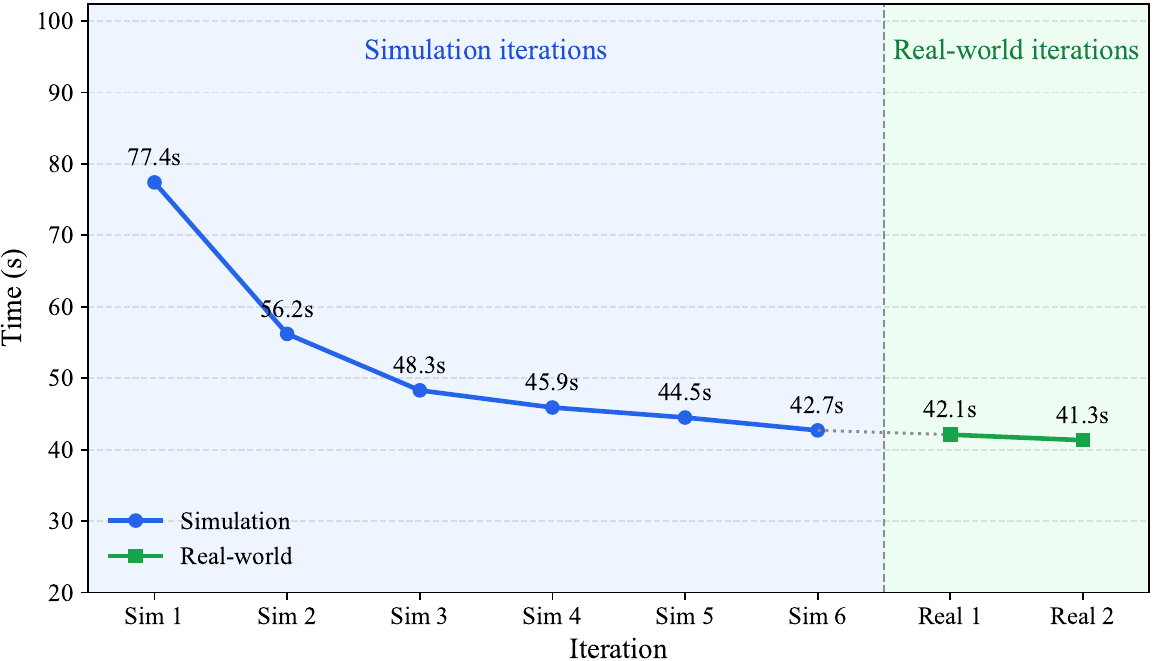}
		\caption{Iteration-wise traversal time reduction for the lifting-wing quadcopter experiment. The traversal time is first reduced through hardware-in-the-loop simulation iterations and is then further refined in real-world flight iterations, demonstrating the transferability and executability of the learned VIP traversal command on the physical platform.}
		\label{liftt}
	\end{figure}

	The iteration-wise result in Fig.~\ref{liftt} further illustrates the learning process of VIP on the lifting-wing quadcopter. During the hardware-in-the-loop simulation stage, the traversal time decreases from $77.4 \mathrm{s}$ to $42.7 \mathrm{s}$, indicating that the traversal-speed profile is progressively improved prior to physical deployment. After transferring the learned command to the real platform, two additional real-world iterations further reduce the traversal time from $42.1 \mathrm{s}$ to $41.3 \mathrm{s}$. This result indicates that simulation iterations can provide an effective initialization for VIP, while a small number of physical rollouts can compensate for platform-specific effects such as aerodynamic uncertainty, actuation delay, and tracking mismatch.
	
	The experiment also demonstrates the practical effect of the energy-traversal trade-off. In straight segments, the vehicle can maintain a relatively high traversal speed because the path-following error is easier to regulate. In turning regions, the trajectory becomes more sensitive to velocity and attitude tracking errors, and the traversal command is correspondingly more conservative to preserve sufficient control authority. This behavior is consistent with the design intuition of VIP: traversal speed increases when tracking energy remains acceptable and decreases when aggressive motion may degrade path-following stability. The successful flight indicates that VIP can be integrated into an existing flight-control architecture as an outer-loop planning and traversal-command learning module.

	\subsection{Swarm online learning in a virtual tube}
	\subsubsection{Experimental Settings}
	
	The swarm experiment is conducted with three homogeneous quadcopters in an outdoor environment. The hardware configuration of each quadcopter is the same as that used in the single-drone experiment. A virtual tube is constructed along the planned traversal corridor, and each robot is guided by its corresponding virtual-tube controller. VIP is used to regulate the swarm's common traversal motion along the tube, while the virtual-tube controller maintains the relative distribution, inter-robot spacing, and boundary constraints.
	
	This experiment is designed to verify the extension of VIP from single-robot navigation to multi-robot flight. The energy modeling of the swarm experiment follows the formulation used in the simulation, in which the energy term describes the swarm's distribution deviation within the virtual tube. Unlike methods that optimize a separate trajectory for each robot, the proposed framework operates on the common traversal component at the swarm level. The microscopic controller then regulates local interactions among robots and maintains the tube constraint. This division is consistent with the macro--micro connection discussed in Subsection~\ref{subsec:classicVTC}: the macroscopic learning term determines the swarm's forward progress, while the microscopic interaction terms maintain local distribution and safety.

	\begin{figure*}[htbp]
		\centering
		\includegraphics[width=6.8in]{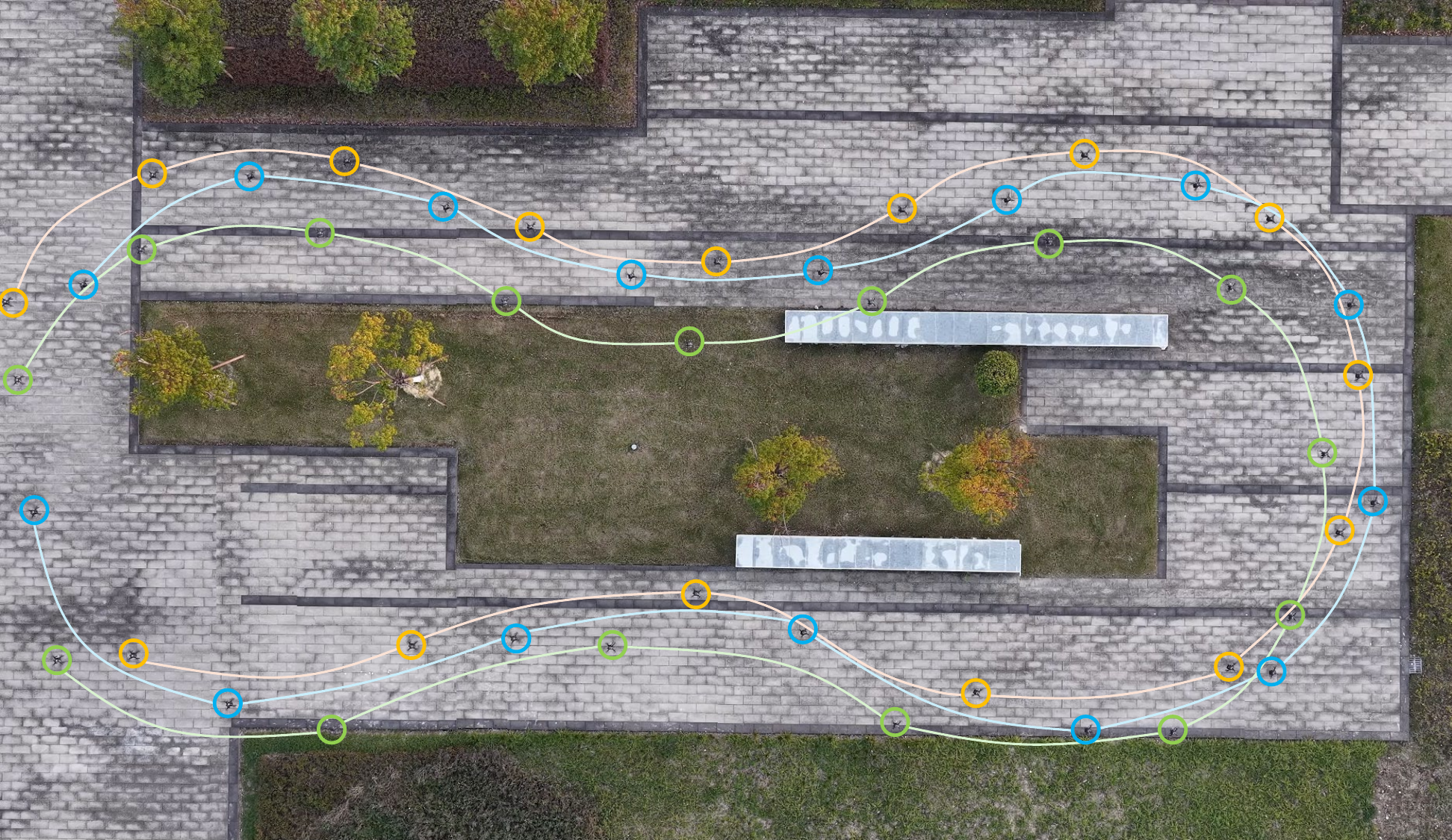}
		\caption{Image sequence of three-quadcopter swarm flight inside an outdoor virtual tube. The snapshots show coordinated forward traversal while maintaining bounded relative configuration and virtual-tube safety constraints.}
		\label{swarm}
	\end{figure*}
	
	\subsubsection{Experimental Results}
\begin{figure}[htbp]
		\centering
		\includegraphics[width=3.3in]{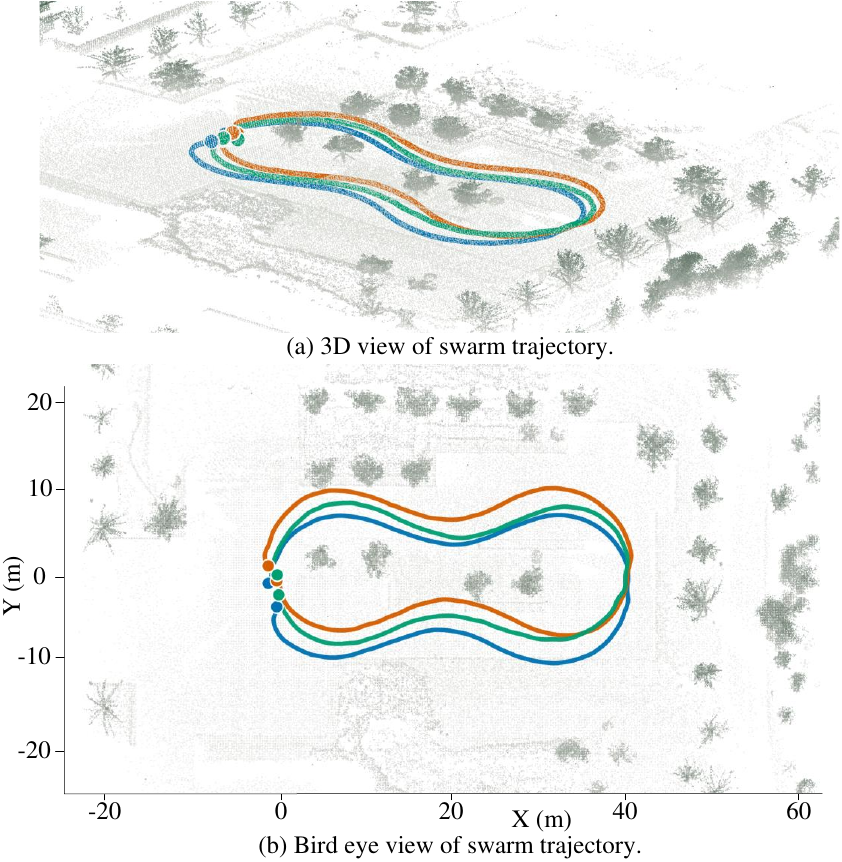}
		\caption{Recorded trajectories of the three-quadcopter swarm in the reconstructed point cloud and bird's-eye view. The results demonstrate virtual-tube-constrained swarm motion and the execution of the learned common traversal command in real flight.}
		\label{swarm_cloud}
	\end{figure}

	The swarm flight process is shown in Fig.~\ref{swarm}. The overlaid snapshots illustrate that the three quadcopters move forward along the outdoor virtual tube while maintaining a bounded relative configuration. The corresponding 3-D and bird's-eye trajectories are shown in Fig.~\ref{swarm_cloud}. 
	
	The initial traversal speed of the swarm is set to $0.5 \mathrm{m/s}$. After five iterations, the maximum traversal speed reaches $2 \mathrm{m/s}$. This result shows that VIP can progressively increase the common traversal command as long as the swarm distribution energy remains within an acceptable range. At the same time, the virtual-tube controller maintains the local distribution and boundary constraints, preventing the increased traversal speed from directly causing unsafe divergence among robots.
	
\begin{figure}[htbp]
		\centering
		\includegraphics[width=3.2in]{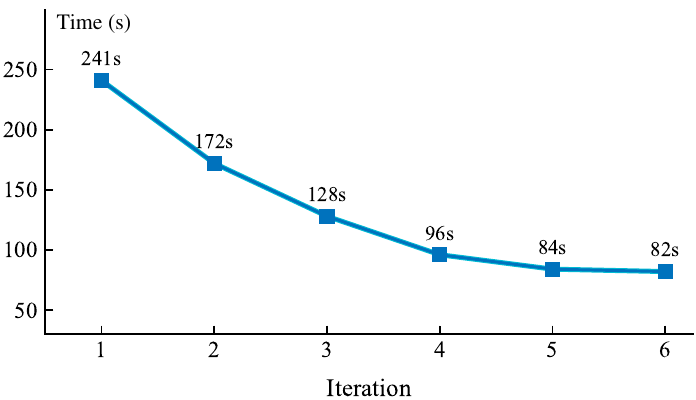}
		\caption{Iteration-wise traversal time reduction for the three-quadcopter swarm experiment. The common traversal command learned by VIP reduces the swarm traversal time across iterations, while the virtual-tube controller maintains inter-robot spacing, distribution regulation, and boundary constraints during real flight.}
		\label{swarmt}
	\end{figure}

	The iteration-wise traversal time is plotted in Fig.~\ref{swarmt}. The traversal time decreases from $241 \mathrm{s}$ in the first rollout to $82 \mathrm{s}$ after six iterations, confirming that VIP progressively improves the swarm-level traversal command through repeated executions. This reduction is achieved without optimizing independent trajectories for all quadcopters. Instead, VIP updates the macroscopic traversal component shared by the swarm, whereas the virtual-tube controller preserves microscopic distribution regulation and collision-safe motion. Therefore, Fig.~\ref{swarmt} directly supports the proposed macro--micro structure for improving swarm traversal efficiency with low computational burden.
	
	These results provide practical evidence for extending VIP to multi-robot systems. In the single-drone experiment, the energy term corresponds to path-following deviation. In the swarm experiment, the energy term represents the deviation in distribution within the virtual tube. Although the physical meanings of the energy functions differ, the learning mechanism remains the same: VIP adjusts the traversal command based on the recorded energy variation. Therefore, the experiment demonstrates that the proposed framework can accommodate different energy definitions while preserving a unified traversal-command learning structure.
	
	The swarm experiment also illustrates the benefit of separating macroscopic traversal learning from microscopic interaction control. VIP does not need to optimize each quadcopter's individual trajectory. Instead, it learns a common traversal profile for the swarm, while the virtual-tube controller handles local spacing and boundary regulation. This structure reduces the burden of multi-robot trajectory optimization and provides a practical implementation path for enhancing traversal efficiency at the swarm level. Together with the single-drone experiment, the result supports the central claim of this paper: VIP can serve as a unified outer-loop learning layer for improving traversal efficiency in both single-robot and robotic-swarm navigation.
	
	\section{Conclusion} \label{conclusion}
	This paper established VIP as a unified planning-level iterative-learning framework with three core capabilities. First, a unified energy-variation planning formulation represents task-dependent performance objectives through a common energy function, enabling the same spatial learning principle to address single-robot path following and robotic-swarm density regulation. Second, function-level planning optimization yields an $\mathcal{O}(n)$ computational complexity per model-free update without requiring online adjoint solution or horizon-based optimization. Third, online and model-free planning is realized through execution feedback: the same update law uses simulated rollouts in the model-in-the-loop mode and recorded physical rollouts in the robot-in-the-loop mode. The theoretical analysis establishes descent and neighborhood-approximation properties, while simulations and real-world experiments demonstrate reduced traversal time, bounded task error, and applicability across different robotic platforms and swarm configurations. Future work will investigate within-rollout adaptation, dynamic environments, explicit uncertainty-aware safety constraints, and larger robot swarms with fully onboard learning.

	\section*{Appendix}
	\appendices 
	\section{Detailed derivation for the single-robot formulation} \label{app:singleverification}
	This appendix collects the calculations omitted from \emph{Subsection~\ref{singleIL}} and \emph{Section~\ref{singlerealization}}, so that the main text can retain the concrete-to-abstract sequence and present only the energy result needed by VIP.
	For compactness, write $V=V_{\mathrm{s}}$ throughout this appendix.
	
	\begin{lemma} \label{lemmaerr}
		For $\mathbf{e}_{\mathrm{p}}=\mathbf{m}(\mathbf{p})-\mathbf{p}$, there hold $\nabla_{\mathbf{p}}e_{\mathrm{p}}=-\mathbf{e}_{\mathrm{p}}/e_{\mathrm{p}}$ and $\mathbf{t}_{\mathrm{c}}^{\mathrm{T}}\mathbf{e}_{\mathrm{p}}=0$ (\cite{goncalves2010vector}).
	\end{lemma}
	
	First consider perfect velocity execution, $\dot{\mathbf{p}}=\mathbf{v}_{\mathrm{c}}$, and only the nominal command in (\ref{singlevc}). Since $\Vert\mathbf{v}^{\prime}_{\mathrm{n}}\Vert=\lambda_2V$ with $\lambda_2=k_{\mathrm{p}}$, \emph{Lemma~\ref{lemmaerr}} gives
	\begin{equation}
		\begin{split}
			\dot V
			&=\mathbf{e}_{\mathrm{p}}^{\mathrm{T}}
			\left(\mathbf{t}_{\mathrm{c}}\dot l-\dot{\mathbf{p}}\right)
			=-\mathbf{e}_{\mathrm{p}}^{\mathrm{T}}\mathbf{v}^{\prime}_{\mathrm{n}}\\
			&=-\sqrt{2}\lambda_2 V^{3/2}
			=-\lambda_{\mathrm{s}}\Vert\mathbf{v}^{\prime}_{\mathrm{n}}\Vert^{3/2},
		\end{split}
		\label{dotVsingle}
	\end{equation}
	for an appropriate $\lambda_{\mathrm{s}}>0$. This establishes the nominal single-robot dissipation term with $p=3/2$.
	
	For the velocity dynamics (\ref{veduc}), \emph{Assumption~\ref{assumvtrack}} gives $\mathbf{v}=\mathbf{v}_{\mathrm{c}}+\delta\mathbf{v}$. Before saturation,
	\begin{equation}
		\begin{split}
			\dot V
			&=-\mathbf{e}_{\mathrm{p}}^{\mathrm{T}}
			\left(\mathbf{v}^{\prime}_{\mathrm{n}}+\delta\mathbf{v}\right)\\
			&=-\sqrt{2}\lambda_2 V^{3/2}+\varepsilon_{\mathrm{s}}\sqrt{2V}
			=-\lambda_{\mathrm{s}}v_{\mathrm{n}}^{3/2}+\varepsilon_{\mathrm{s}}v_{\mathrm{n}}^{1/2},
		\end{split}
		\label{dotVdouble}
	\end{equation}
	where bounded factors have been absorbed into $\varepsilon_{\mathrm{s}}$. Under the saturated composite command (\ref{singlevc}), the nominal term is multiplied by $\kappa_v$ and the tangent term vanishes in the direct inner product because $\mathbf{e}_{\mathrm{p}}^{\mathrm{T}}\mathbf{t}_{\mathrm{c}}=0$. Therefore,
	\begin{equation}
		\dot V
		=-\lambda_{\mathrm{s}}\kappa_vv_{\mathrm{n}}^{3/2}
		+\varepsilon_{\mathrm{s}}v_{\mathrm{n}}^{1/2},
	\end{equation}
	which is Eq.~(\ref{dotVsatleq}) and yields the unified energy form (\ref{Vmodel2}) with $p=3/2$, $q=1/2$, and $o=0$.
	
	\section{Supplementary material for the robotic-swarm formulation} \label{app:swarmverification}
	This appendix collects the Wasserstein-space interpretation, KDE realization, desired-density construction, and integral calculations omitted from Subsection~\ref{swarmIL}. The symbols $\rho$, $\rho_{\mathrm{d}}$, $\phi$, $V_{\mathrm{m}}$, $\mathbf{v}^{\prime}_{\mathrm{n}}$, $\mathbf{v}^{\prime}_{\mathrm{t}}$, and $v_{\mathrm{n}}$ retain exactly the definitions given in the main text.
	
	\subsection{Wasserstein-space interpretation}
	Let $\mathcal{P}_{2}(\mathcal{T})$ be the set of probability measures on $\mathcal{T}$ with finite second moments,
	\begin{equation}
		\mathcal{P}_{2}(\mathcal{T})
		\triangleq
		\left\{
		\mu:\int_{\mathcal{T}}\Vert\mathbf{x}\Vert^{2}\,\mathrm{d}\mu(\mathbf{x})<\infty
		\right\}.
		\label{eq:P2_space}
	\end{equation}
	For $\mu,\eta\in\mathcal{P}_{2}(\mathcal{T})$, the 2-Wasserstein distance is
	\begin{equation}
		\mathcal{W}_{2}(\mu,\eta)
		\triangleq
		\left(
		\inf_{\gamma\in\Upsilon(\mu,\eta)}
		\int_{\mathcal{T}\times\mathcal{T}}
		\Vert\mathbf{x}-\mathbf{y}\Vert^{2}\,\mathrm{d}\gamma(\mathbf{x},\mathbf{y})
		\right)^{1/2},
		\label{eq:W2_distance}
	\end{equation}
	where $\Upsilon(\mu,\eta)$ denotes the set of transport plans with marginals $\mu$ and $\eta$. This representation allows the swarm state to be described by a probability density independently of the number of robots.
	
	For a sufficiently regular functional $\mathcal{E}:\mathcal{P}_{2}(\mathcal{T})\rightarrow\mathbb{R}$, its formal Wasserstein gradient-flow velocity is
	\begin{equation}
		\mathbf{v}_{\mathcal{E}}
		=-\nabla_{\mathbf{p}}\frac{\delta\mathcal{E}}{\delta\rho}.
		\label{eq:wasserstein_gradient_field}
	\end{equation}
	Selecting the negative differential entropy
	\begin{equation}
		\mathcal{E}(\rho)
		=\int_{\mathcal{T}}\rho(\mathbf{p})\log\rho(\mathbf{p})\,\mathrm{d}\mathbf{p}
		\label{eq:negative_entropy}
	\end{equation}
	gives
	\begin{equation}
		\mathbf{v}_{\mathcal{E}}
		=-\nabla_{\mathbf{p}}\log\rho
		=-\frac{\nabla_{\mathbf{p}}\rho}{\rho}.
		\label{eq:entropy_velocity}
	\end{equation}
	Substitution into the continuity equation (\ref{swarmpde}) yields
	\begin{equation}
		\frac{\partial\rho}{\partial t}=\Delta\rho,
		\label{eq:heat_equation}
	\end{equation}
	which explains the diffusion interpretation of the nominal swarm-distribution field.
	
	\subsection{KDE realization and desired-density construction}\label{app:desired-density-design}
	For a finite swarm, the density is reconstructed from the robot positions using KDE (\cite{zheng2021transporting}):
	\begin{equation}
		\rho_M(t,\mathbf{p})
		=\frac{1}{Mh^{3}}
		\sum_{i=1}^{M}
		K\left(\frac{\mathbf{p}-\mathbf{p}_{i}(t)}{h}\right),
		\label{KDE}
	\end{equation}
	where $h>0$ is the bandwidth and the Gaussian kernel is
	\begin{equation}
		K(\mathbf{x})
		=\frac{1}{(2\pi)^{3/2}}
		\exp\left(-\frac{1}{2}\mathbf{x}^{\mathrm{T}}\mathbf{x}\right).
		\label{eq:gaussian_kernel}
	\end{equation}
	Under standard density-regularity and bandwidth conditions, including $h=h_M\rightarrow0$ and $Mh_M^{3}\rightarrow\infty$ as $M\rightarrow\infty$, $\rho_M$ consistently approximates the underlying density. When kernel mass is truncated by the tube boundary, the estimator is renormalized over $\mathcal{T}$. In the main text, $\rho$ denotes either the macroscopic density or this finite-swarm KDE realization.
	
	\begin{figure}[htbp]
		\centering
		\includegraphics[width=3.3in]{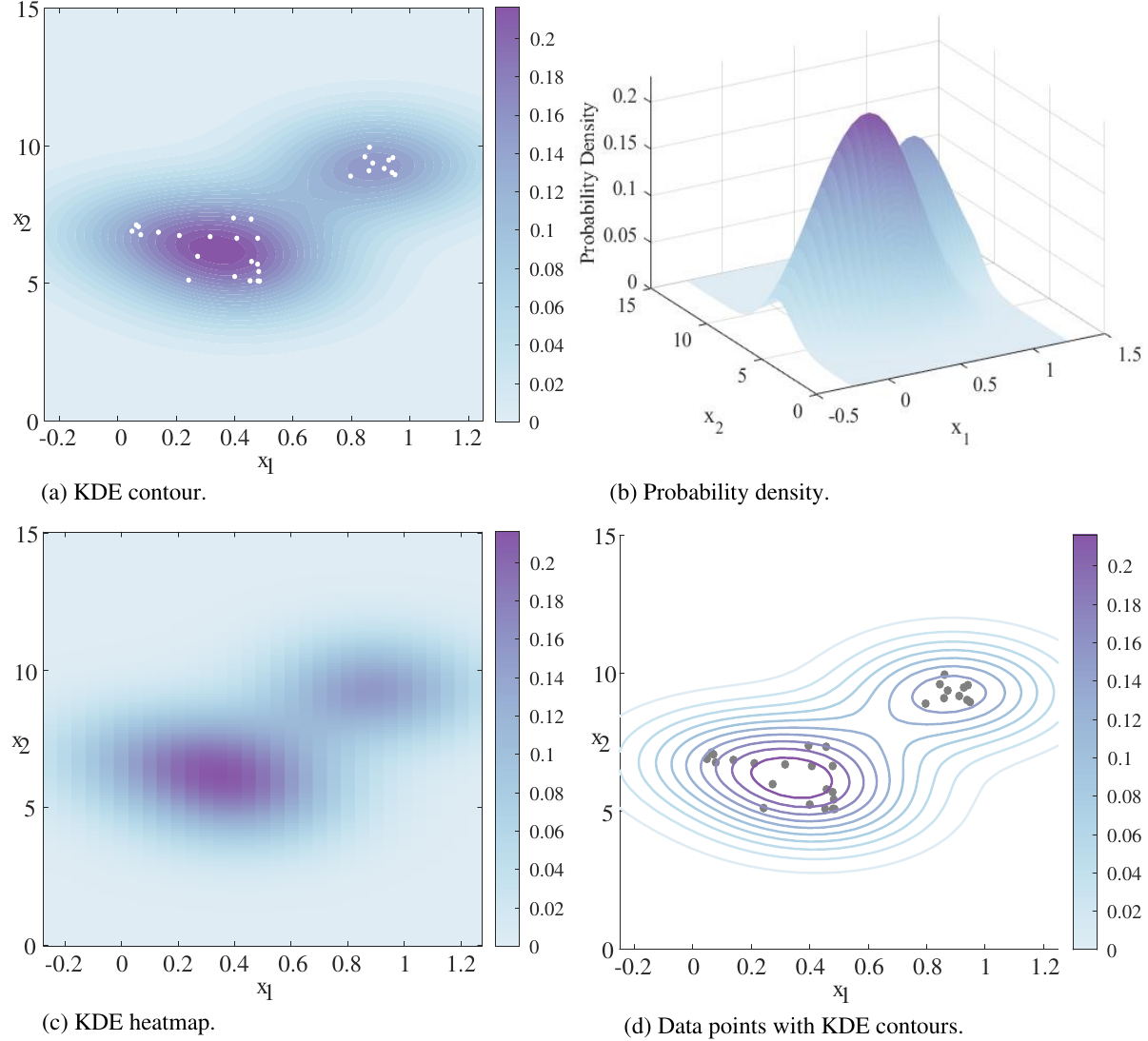}
		\caption{Kernel-density reconstruction and diffusion-gradient field for swarm-distribution regulation.}
		\label{kdensity}
	\end{figure}
	
	Let $l_i=l(\mathbf{p}_i)$ be the spatial coordinate of robot $i$, and define
	\begin{equation}
		l_{\mathrm{s}}=\min_i l_i,
		l_{\mathrm{b}}=\max_i l_i,
		\mathbf{p}_{\mathrm{m}}=\frac{1}{M}\sum_{i=1}^{M}\mathbf{p}_i,
		l_{\mathrm{m}}=l(\mathbf{p}_{\mathrm{m}}).
		\label{eq:swarm_spatial_bounds}
	\end{equation}
	The occupied region is defined rigorously as the union of tube cross sections,
	\begin{equation}
		\mathcal{O}(t)
		=\bigcup_{l\in[l_{\mathrm{s}},l_{\mathrm{b}}]}\mathcal{C}_{l}.
		\label{eq:occupied_region}
	\end{equation}
	First construct a center-preserving base support without a backward extension,
	\begin{equation}
		l_{\mathrm{d},0}^{+}
		=\max\left\{l_{\mathrm{b}},2l_{\mathrm{m}}-l_{\mathrm{s}}\right\},
		\mathcal{O}_{\mathrm{d},0}(t)
		=\bigcup_{l\in[l_{\mathrm{s}},l_{\mathrm{d},0}^{+}]}\mathcal{C}_{l}.
		\label{eq:base_desired_occupied_region}
	\end{equation}
	Let $\rho_{\mathrm{d},0}$ be a continuously differentiable normalized approximation of the uniform density on $\mathcal{O}_{\mathrm{d},0}$. For a feasible shift $\Delta_l\in[0,\Delta_{\max}]$, define the forward tube map $\mathcal{S}_{\Delta_l}$ by preserving the normalized cross-sectional coordinate of each point and increasing its longitudinal coordinate from $l$ to $l+\Delta_l$. The final desired support and density are
	\begin{equation}
		\mathcal{O}_{\mathrm{d}}(t)
		=\mathcal{S}_{\Delta_l}\!\left(\mathcal{O}_{\mathrm{d},0}(t)\right),
		\rho_{\mathrm{d}}(t,\cdot)
		=(\mathcal{S}_{\Delta_l})_{\#}\rho_{\mathrm{d},0}(t,\cdot),
		\label{eq:desired_density_forward_shift}
	\end{equation}
	where $(\mathcal{S}_{\Delta_l})_{\#}$ denotes the push-forward operation. The upper bound $\Delta_{\max}$ keeps the shifted support inside the virtual tube. To determine the shift, define
	\begin{equation}
		\mu(\mathbf{p})
		=\frac{\Vert\nabla_{\mathbf{p}}\phi(\mathbf{p})\Vert^2}
		{\Vert\nabla_{\mathbf{p}}\phi\Vert_{\mathcal{L}^2(\mathcal{T})}^2},
		\qquad
		r=\Vert\mathbf{v}^{\prime}_{\mathrm{c}}\Vert,
	\end{equation}
	and let
	\begin{equation}
		\mathcal{T}_{\mathrm{a}}
		=\{\mathbf{p}\in\mathcal{T}:r(\mathbf{p})>v_{\mathrm{m}}\},
		\qquad
		w(\mathbf{p})
		=\mu(\mathbf{p})\frac{\kappa_v(\mathbf{p})}{r^2(\mathbf{p})}.
	\end{equation}
	The feasible forward shift is selected to satisfy the integral acute-angle criterion
	\begin{equation}
		\mathcal{C}_{\mathrm{nt}}(\Delta_l)
		\triangleq
		\int_{\mathcal{T}_{\mathrm{a}}}
		w(\mathbf{p})
		\mathbf{v}^{\prime\mathrm{T}}_{\mathrm{n}}(\mathbf{p};\Delta_l)
		\mathbf{v}^{\prime}_{\mathrm{t}}(\mathbf{p})
		\,\mathrm{d}\mathbf{p}
		\geq0.
		\label{eq:aggregate_acute_design}
	\end{equation}
	Condition (\ref{eq:aggregate_acute_design}) requires alignment only after spatial aggregation over the active-saturation region. Therefore, it does not require the two fields to form an acute angle at every spatial location, and local counter-motion remains admissible during density redistribution. Since $\Delta_l$ is scalar, the criterion can be checked by a one-dimensional search over $[0,\Delta_{\max}]$. \emph{Proposition~\ref{prop:saturation_derivatives}} applies whenever the resulting feasible-shift set is nonempty. The choice $\Delta_l=0$ recovers the center-preserving construction when it already satisfies (\ref{eq:aggregate_acute_design}).
	
	\subsection{Nominal energy dissipation}
	For a stationary desired density and $\mathbf{v}_{\mathrm{c}}=\mathbf{v}^{\prime}_{\mathrm{n}}$, differentiating (\ref{swarmenergy}) along (\ref{swarmpde}) gives
	\begin{equation}
		\begin{split}
			\dot V_{\mathrm{m}}
			&=-\int_{\mathcal{T}}\phi
			\nabla_{\mathbf{p}}\cdot
			\left(\rho\mathbf{v}^{\prime}_{\mathrm{n}}\right)
			\,\mathrm{d}\mathbf{p}\\
			&=\int_{\mathcal{T}}
			\nabla_{\mathbf{p}}\phi^{\mathrm{T}}
			\rho\mathbf{v}^{\prime}_{\mathrm{n}}\,\mathrm{d}\mathbf{p}\\
			&=-k_{\rho}\int_{\mathcal{T}}
			\Vert\nabla_{\mathbf{p}}\phi\Vert^{2}\,\mathrm{d}\mathbf{p}.
		\end{split}
		\label{app:swarm_nominal_identity}
	\end{equation}
	The second equality follows from the divergence theorem. The boundary term vanishes because of the lateral no-flux condition and the assumed separation of the density support from the terminal cross sections. Since $\rho$ and $\rho_{\mathrm{d}}$ are normalized, $\phi$ has zero spatial mean. The Poincar\'e--Wirtinger inequality therefore gives
	\begin{equation}
		\int_{\mathcal{T}}\phi^{2}\,\mathrm{d}\mathbf{p}
		\leq C_{\mathrm{P}}^{2}
		\int_{\mathcal{T}}\Vert\nabla_{\mathbf{p}}\phi\Vert^{2}\,\mathrm{d}\mathbf{p},
	\end{equation}
	which yields (\ref{eq:swarm_nominal_energy_identity}).
	
	\subsection{Composite-field energy identity and coefficient bounds}
	For later use, define
	\begin{equation}
		A_{\phi}
		\triangleq\Vert\nabla_{\mathbf{p}}\phi\Vert_{\mathcal{L}^{2}(\mathcal{T})},
		\qquad
		v_{\mathrm{n}}\triangleq k_{\rho}A_{\phi}.
		\label{eq:swarm_regulation_intensity}
	\end{equation}
	For $A_{\phi}>0$, let
	\begin{equation}
		\begin{split}
			\bar{\kappa}_{v}
			&\triangleq
			\frac{\displaystyle
				\int_{\mathcal{T}}\kappa_v
				\Vert\nabla_{\mathbf{p}}\phi\Vert^{2}\,\mathrm{d}\mathbf{p}}
			{A_{\phi}^{2}},\\
			\beta_{\rho}
			&\triangleq
			\frac{\displaystyle
				\int_{\mathcal{T}}\kappa_v\rho
				\nabla_{\mathbf{p}}\phi^{\mathrm{T}}\mathbf{t}_{\mathrm{c}}\,\mathrm{d}\mathbf{p}}
			{A_{\phi}},
		\end{split}
		\label{eq:swarm_effective_coefficients}
	\end{equation}
	and set $\bar{\kappa}_{v}=1$ and $\beta_{\rho}=0$ when $A_{\phi}=0$. The desired-density variation residual is
	\begin{equation}
		\varepsilon_{1,\mathrm{m}}
		\triangleq
		-\int_{\mathcal{T}}
		\phi\frac{\partial\rho_{\mathrm{d}}}{\partial t}\,\mathrm{d}\mathbf{p}.
		\label{eq:swarm_reference_residual}
	\end{equation}
	These quantities satisfy
	\begin{equation}
		0<\bar{\kappa}_{v}\leq1,
		|\beta_{\rho}|\leq\Vert\rho\Vert_{\mathcal{L}^{2}(\mathcal{T})},
		|\varepsilon_{1,\mathrm{m}}|
		\leq\sqrt{2V_{\mathrm{m}}}
		\left\Vert\frac{\partial\rho_{\mathrm{d}}}{\partial t}\right\Vert_{\mathcal{L}^{2}(\mathcal{T})}.
		\label{eq:swarm_coefficient_bounds}
	\end{equation}
	
	For compactness, write $V=V_{\mathrm{m}}$. From $\phi=\rho-\rho_{\mathrm{d}}$ and (\ref{swarmpde}),
	\begin{equation}
		\begin{split}
			\dot V
			&=\int_{\mathcal{T}}\phi
			\left(\frac{\partial\rho}{\partial t}
			-\frac{\partial\rho_{\mathrm{d}}}{\partial t}\right)
			\,\mathrm{d}\mathbf{p}\\
			&=-\int_{\mathcal{T}}\phi
			\nabla_{\mathbf{p}}\cdot(\rho\mathbf{v}_{\mathrm{c}})
			\,\mathrm{d}\mathbf{p}
			-\int_{\mathcal{T}}\phi
			\frac{\partial\rho_{\mathrm{d}}}{\partial t}
			\,\mathrm{d}\mathbf{p}.
		\end{split}
		\label{app:swarmenergyidentity}
	\end{equation}
	Applying the divergence theorem under the same boundary conditions gives
	\begin{equation}
		\dot V
		=\int_{\mathcal{T}}
		\nabla_{\mathbf{p}}\phi^{\mathrm{T}}
		\rho\mathbf{v}_{\mathrm{c}}\,\mathrm{d}\mathbf{p}
		+\varepsilon_{1,\mathrm{m}}.
		\label{app:swarm_after_parts}
	\end{equation}
	Substituting (\ref{dswarmc}) and
	$\rho\mathbf{v}^{\prime}_{\mathrm{n}}=-k_{\rho}\nabla_{\mathbf{p}}\phi$ gives
	\begin{equation}
		\begin{split}
			\dot V
			&=-k_{\rho}\int_{\mathcal{T}}
			\kappa_v\Vert\nabla_{\mathbf{p}}\phi\Vert^2
			\,\mathrm{d}\mathbf{p}\\
			&\quad+v_{\mathrm{t}}
			\int_{\mathcal{T}}\kappa_v\rho
			\nabla_{\mathbf{p}}\phi^{\mathrm{T}}\mathbf{t}_{\mathrm{c}}
			\,\mathrm{d}\mathbf{p}
			+\varepsilon_{1,\mathrm{m}}.
		\end{split}
		\label{eq:swarm_exact_composite_energy}
	\end{equation}
	With the definitions in (\ref{eq:swarm_regulation_intensity}) and (\ref{eq:swarm_effective_coefficients}), (\ref{eq:swarm_exact_composite_energy}) becomes exactly (\ref{dotVswarm}). Finally, since $0<\kappa_v\leq1$ and $\Vert\mathbf{t}_{\mathrm{c}}\Vert=1$, the Cauchy--Schwarz inequality gives
	\begin{equation}
		\begin{split}
			|\beta_{\rho}|
			&\leq
			\frac{
				\Vert\kappa_v\rho\mathbf{t}_{\mathrm{c}}\Vert_{\mathcal{L}^{2}}
				\Vert\nabla_{\mathbf{p}}\phi\Vert_{\mathcal{L}^{2}}
			}{\Vert\nabla_{\mathbf{p}}\phi\Vert_{\mathcal{L}^{2}}}
			\leq\Vert\rho\Vert_{\mathcal{L}^{2}},\\
			|\varepsilon_{1,\mathrm{m}}|
			&\leq
			\Vert\phi\Vert_{\mathcal{L}^{2}}
			\left\Vert\frac{\partial\rho_{\mathrm{d}}}{\partial t}\right\Vert_{\mathcal{L}^{2}}
			=\sqrt{2V_{\mathrm{m}}}
			\left\Vert\frac{\partial\rho_{\mathrm{d}}}{\partial t}\right\Vert_{\mathcal{L}^{2}}.
		\end{split}
	\end{equation}
	These inequalities establish (\ref{eq:swarm_coefficient_bounds}) and the boundedness statements used in \emph{Proposition~\ref{prop:commonenergy}}.
	
	\section{Proof of Proposition~\ref{prop:commonenergy}}
	\label{app:proof-common-energy}
	\begin{proof}
		Equation~(\ref{dotVsatleq}) has the form (\ref{Vmodel2}) with
		$(p,q,o)=(3/2,1/2,0)$, $\lambda_1=\lambda_{\mathrm{s}}$,
		$\kappa_E=\kappa_v$, $\varepsilon=\varepsilon_{\mathrm{s}}$, and
		$\varepsilon_1=0$. Equation~(\ref{dotVswarm}) gives
		$(p,q,o)=(2,1,1)$, $\lambda_1=1/k_{\rho}$,
		$\kappa_E=\bar\kappa_v$, $\varepsilon=\beta_{\rho}/k_{\rho}$, and
		$\varepsilon_1=\varepsilon_{1,\mathrm{m}}$. Boundedness follows from
		\emph{Assumption~\ref{assumvtrack}} and (\ref{eq:swarm_coefficient_bounds}).
	\end{proof}
	
	\section{Detailed conditions and proof of Proposition~\ref{prop:saturation_derivatives}}
	\label{app:proof-saturation-derivatives}
	On the considered compact operating set, $v_{\mathrm{n}}$ is nondecreasing with $V$. For the swarm, the partial derivative with respect to $V$ is evaluated along the radial scalarization
	\begin{equation}
		\frac{\partial\mathbf{v}^{\prime}_{\mathrm{n}}}{\partial V}
		=\alpha_V\mathbf{v}^{\prime}_{\mathrm{n}},
		\qquad
		\alpha_V\geq0,
		\qquad
		\frac{\partial\mu}{\partial V}=0,
		\label{eq:swarm_radial_scalarization}
	\end{equation}
	where $\mu=\Vert\nabla_{\mathbf{p}}\phi\Vert^2/A_{\phi}^2$. This scalarization freezes only the normalized field shape during the partial differentiation; the physical density profile remains time varying through the coefficients and residuals in (\ref{Vmodel2}).
	
	At a fixed spatial location, write
	\begin{equation}
		\mathbf{v}^{\prime}_{\mathrm{n}}=a_{\mathrm{n}}\widehat{\mathbf{n}},
		\mathbf{v}^{\prime}_{\mathrm{t}}=v_{\mathrm{t}}\widehat{\mathbf{t}},
		r=\Vert\mathbf{v}^{\prime}_{\mathrm{c}}\Vert,
		\label{eq:local_component_geometry}
	\end{equation}
	with unit vectors $\widehat{\mathbf{n}}$ and $\widehat{\mathbf{t}}$. In the unsaturated region, $\kappa_v=1$, and hence
	\begin{equation}
		\frac{\partial\kappa_v}{\partial V}=0,
		\qquad
		\frac{\partial\kappa_v}{\partial v_{\mathrm{t}}}=0,
		\qquad
		\frac{\partial(\kappa_vv_{\mathrm{t}})}{\partial v_{\mathrm{t}}}=1.
		\label{eq:unsaturated_saturation_derivatives}
	\end{equation}
	In the active-saturation region, $\kappa_v=v_{\mathrm{m}}/r$, and direct differentiation gives
	\begin{equation}
		\frac{\partial\kappa_v}{\partial V}
		=-\frac{\kappa_v}{r^2}
		\mathbf{v}^{\prime\mathrm{T}}_{\mathrm{c}}
		\frac{\partial\mathbf{v}^{\prime}_{\mathrm{n}}}{\partial V},
		\qquad
		\frac{\partial\kappa_v}{\partial v_{\mathrm{t}}}
		=-\frac{\kappa_v}{r^2v_{\mathrm{t}}}
		\mathbf{v}^{\prime\mathrm{T}}_{\mathrm{c}}
		\mathbf{v}^{\prime}_{\mathrm{t}}.
		\label{eq:kappa_partial_derivatives}
	\end{equation}
	Moreover,
	\begin{equation}
		\frac{\partial(\kappa_vv_{\mathrm{t}})}{\partial v_{\mathrm{t}}}
		=\frac{\kappa_v}{r^2}
		\mathbf{v}^{\prime\mathrm{T}}_{\mathrm{c}}
		\mathbf{v}^{\prime}_{\mathrm{n}}.
		\label{eq:kappavt_derivative_identity}
	\end{equation}
	
	\begin{proof}
		For the single robot, the command construction in (\ref{singlevc}) gives
		$\widehat{\mathbf{n}}^{\mathrm{T}}\widehat{\mathbf{t}}=0$. Therefore, using
		$\mathbf{v}^{\prime}_{\mathrm{c}}
		=\mathbf{v}^{\prime}_{\mathrm{n}}
		+\mathbf{v}^{\prime}_{\mathrm{t}}$,
		\begin{equation}
			\begin{split}
				\mathbf{v}^{\prime\mathrm{T}}_{\mathrm{c}}
				\mathbf{v}^{\prime}_{\mathrm{n}}
				&=\Vert\mathbf{v}^{\prime}_{\mathrm{n}}\Vert^2,\\
				\mathbf{v}^{\prime\mathrm{T}}_{\mathrm{c}}
				\mathbf{v}^{\prime}_{\mathrm{t}}
				&=\Vert\mathbf{v}^{\prime}_{\mathrm{t}}\Vert^2.
			\end{split}
			\label{eq:single_geometry_derived}
		\end{equation}
		Since $\partial\mathbf{v}^{\prime}_{\mathrm{n}}/\partial V
		=(\partial a_{\mathrm{n}}/\partial V)\widehat{\mathbf{n}}$
		with $\partial a_{\mathrm{n}}/\partial V\geq0$, substitution of
		(\ref{eq:single_geometry_derived}) into
		(\ref{eq:kappa_partial_derivatives})--(\ref{eq:kappavt_derivative_identity}) proves (\ref{eq:saturation_monotonicity}) for the single robot. Thus, no additional non-opposition condition is imposed.
		
		For the swarm, use the active and unsaturated sets and the weight $w$ defined in \emph{Appendix~\ref{app:desired-density-design}}. Define the weighted self-products
		\begin{equation}
			\mathcal{A}_{\mathrm{n}}
			\triangleq
			\int_{\mathcal{T}_{\mathrm{a}}}
			w\Vert\mathbf{v}^{\prime}_{\mathrm{n}}\Vert^2
			\,\mathrm{d}\mathbf{p},
			\qquad
			\mathcal{A}_{\mathrm{t}}
			\triangleq
			\int_{\mathcal{T}_{\mathrm{a}}}
			w\Vert\mathbf{v}^{\prime}_{\mathrm{t}}\Vert^2
			\,\mathrm{d}\mathbf{p}.
			\label{eq:aggregate_self_products}
		\end{equation}
		By the forward-shift design (\ref{eq:aggregate_acute_design}),
		$\mathcal{C}_{\mathrm{nt}}\geq0$. Expanding the composite field inside the two weighted integrals gives
		\begin{equation}
			\begin{split}
				\int_{\mathcal{T}_{\mathrm{a}}}
				w\mathbf{v}^{\prime\mathrm{T}}_{\mathrm{c}}
				\mathbf{v}^{\prime}_{\mathrm{n}}\,\mathrm{d}\mathbf{p}
				&=\mathcal{A}_{\mathrm{n}}+\mathcal{C}_{\mathrm{nt}}\geq0,\\
				\int_{\mathcal{T}_{\mathrm{a}}}
				w\mathbf{v}^{\prime\mathrm{T}}_{\mathrm{c}}
				\mathbf{v}^{\prime}_{\mathrm{t}}\,\mathrm{d}\mathbf{p}
				&=\mathcal{A}_{\mathrm{t}}+\mathcal{C}_{\mathrm{nt}}\geq0.
			\end{split}
			\label{eq:aggregate_geometry_derived}
		\end{equation}
		Hence, the two signs needed in the saturation calculation are consequences of the integral acute-angle design; they are not imposed pointwise.
		
		Integrating (\ref{eq:kappa_partial_derivatives}) over
		$\mathcal{T}_{\mathrm{a}}$ and using
		(\ref{eq:swarm_radial_scalarization}) and
		(\ref{eq:aggregate_geometry_derived}) yields
		\begin{equation}
			\begin{split}
				\frac{\partial\bar\kappa_v}{\partial V}
				&=-\alpha_V
				\left(\mathcal{A}_{\mathrm{n}}+\mathcal{C}_{\mathrm{nt}}\right)
				\leq0,\\
				\frac{\partial\bar\kappa_v}{\partial v_{\mathrm{t}}}
				&=-\frac{1}{v_{\mathrm{t}}}
				\left(\mathcal{A}_{\mathrm{t}}+\mathcal{C}_{\mathrm{nt}}\right)
				\leq0.
			\end{split}
			\label{eq:swarm_kappa_derivatives}
		\end{equation}
		The unsaturated set contributes directly to the progression derivative. Combining (\ref{eq:unsaturated_saturation_derivatives}), (\ref{eq:kappavt_derivative_identity}), and (\ref{eq:aggregate_geometry_derived}) gives
		\begin{equation}
			\frac{\partial(\bar\kappa_vv_{\mathrm{t}})}
			{\partial v_{\mathrm{t}}}
			=\int_{\mathcal{T}_{\mathrm{a}}}\mu\,\mathrm{d}\mathbf{p}
			+\mathcal{A}_{\mathrm{n}}+\mathcal{C}_{\mathrm{nt}}
			\geq0.
			\label{eq:swarm_progression_derivative}
		\end{equation}
		This proves (\ref{eq:saturation_monotonicity}) for the swarm.
	\end{proof}
	
	The proof shows that \emph{Proposition~\ref{prop:saturation_derivatives}} is a direct consequence of the command constructions already introduced for the two tasks. In particular, the swarm result uses only the integral criterion in \emph{Appendix~\ref{app:desired-density-design}}; no pointwise acute-angle or pointwise non-opposition condition is required.
	
	\section{Proof of Theorem \ref{theoremmodelbased}} \label{prooftheoremmodelbased}
	
	\begin{proof}
		By the Lipschitz continuity of the Fr\'echet derivative, the following descent inequality holds:
		\begin{equation}
			\begin{split}
				J(v_{\mathrm{t},k+1})
				&\leq J(v_{\mathrm{t},k})
				+\langle F_{n,k},v_{\mathrm{t},k+1}-v_{\mathrm{t},k}\rangle_{(0,L)} \\
				&\quad+\frac{F_m}{2}\Vert v_{\mathrm{t},k+1}-v_{\mathrm{t},k}\Vert^2_{\mathcal{L}^2(0,L)} .
			\end{split}
		\end{equation}
		Substituting the update law $v_{\mathrm{t},k+1}=v_{\mathrm{t},k}-b_1F_{n,k}$ gives
		\begin{equation}
			J(v_{\mathrm{t},k+1})
			\leq J(v_{\mathrm{t},k})
			-\left(b_1-\frac{F_m b_1^2}{2}\right)
			\Vert F_{n,k}\Vert^2_{\mathcal{L}^2(0,L)} .
		\end{equation}
		Since $0<b_1<2/F_{m}$, there exists
		\begin{equation}
			b_1-\frac{F_{m}b_1^2}{2}>0 .
		\end{equation}
		Therefore, $\{J(v_{\mathrm{t},k})\}$ is monotonically non-increasing. Since $J$ is bounded from below, $\{J(v_{\mathrm{t},k})\}$ is convergent.
		
		Summing the above inequality from $k=0$ to infinity yields
		\begin{equation}
			\sum_{k=0}^{\infty}
			\Vert F_{n,k}\Vert^2_{\mathcal{L}^2(0,L)}
			<\infty .
		\end{equation}
		Hence,
		\begin{equation}
			\lim_{k\rightarrow\infty}
			\Vert F_{n,k}\Vert_{\mathcal{L}^2(0,L)}=0 .
		\end{equation}
		This means that the update direction vanishes asymptotically, and any accumulation point of $\{v_{\mathrm{t},k}\}$ satisfies the first-order stationarity condition. If the stationary solution in the considered neighborhood is unique, the generated sequence converges to this solution.
	\end{proof}

	\section{Proof of Theorem \ref{theoremnear}} \label{prooftheoremnear}
	Let $F_{n,k}$ denote the Fr\'echet derivative of $J$ with respect to $v_{\mathrm{t}}$ at the $k$-th iteration, and define $\zeta_k(l)\triangleq k_{\mathrm{e}}V_k(l)$ and $g_k(l)\triangleq g\left(\zeta_k(l)\right)$. By the Lipschitz continuity of the Fr\'echet derivative, the following descent inequality holds:
	\begin{equation}
		\begin{split}
			J(v_{\mathrm{t},k+1})
			&\leq J(v_{\mathrm{t},k})
			+\langle F_{n,k},v_{\mathrm{t},k+1}-v_{\mathrm{t},k}\rangle_{(0,L)}  \\
			&\quad+
			\frac{F_m}{2}
			\Vert v_{\mathrm{t},k+1}-v_{\mathrm{t},k}\Vert^2_{\mathcal{L}^2(0,L)} .
		\end{split}
	\end{equation}
	Substituting the model-free learning law (\ref{approximateILlaw}) into the above inequality gives
	\begin{equation}
		\begin{split}
			J(v_{\mathrm{t},k+1})
			&\leq
			J(v_{\mathrm{t},k})
			-b_3\langle F_{n,k},g_k\rangle_{(0,L)}\\
			&+
			\frac{F_mb_3^2}{2}
			\Vert g_k\Vert^2_{\mathcal{L}^2(0,L)} .
		\end{split}
	\end{equation}
	
	According to the sign analysis before the construction of the model-free law, $g_k$ is designed as a recorded-energy-based surrogate of the model-based descent direction. Therefore, when $g_k\not\equiv0$, the update direction satisfies the descent-consistency relation
	\begin{equation}
		\langle F_{n,k},g_k\rangle_{(0,L)}>0 .
	\end{equation}
	Equivalently, define
	\begin{equation}
		\mu_k
		=
		\frac{
			\langle F_{n,k},g_k\rangle_{(0,L)}
		}{
			\Vert g_k\Vert^2_{\mathcal{L}^2(0,L)}
		}
		>0 .
	\end{equation}
	Then,
	\begin{equation}
		J(v_{\mathrm{t},k+1})
		\leq
		J(v_{\mathrm{t},k})
		-
		\left(
		b_3\mu_k-\frac{F_{\mathrm{m}}b_3^2}{2}
		\right)
		\Vert g_k\Vert^2_{\mathcal{L}^2(0,L)} .
	\end{equation}
	Thus, for a sufficiently small learning gain satisfying
	\begin{equation}
		0<b_3<\frac{2\mu_k}{F_m},
	\end{equation}
	the coefficient in the bracket is positive, and hence
	\begin{equation}
		J(v_{\mathrm{t},k+1})\leq J(v_{\mathrm{t},k}) .
	\end{equation}
	If a fixed learning gain is used over the considered neighborhood, it is sufficient to choose
	\begin{equation}
		0<b_3<\frac{2\underline{\mu}}{F_m},
		\qquad
		\underline{\mu}
		=
		\inf_{k:g_k\not\equiv0}\mu_k>0 .
	\end{equation}
	Therefore, $\{J(v_{\mathrm{t},k})\}$ is monotonically non-increasing. Since $J$ is bounded from below, $\{J(v_{\mathrm{t},k})\}$ is convergent.
	
	Summing the descent inequality over $k$ yields
	\begin{equation}
		\sum_{k=0}^{\infty}
		\Vert g_k\Vert^2_{\mathcal{L}^2(0,L)}
		<\infty ,
	\end{equation}
	which implies
	\begin{equation}
		\lim_{k\rightarrow\infty}
		\Vert g_k\Vert_{\mathcal{L}^2(0,L)}=0 .
	\end{equation}
	
	According to the construction of $g(\cdot)$ in (\ref{ILlawestimate}), the zero set of $g$ corresponds to the admissible energy band $[\zeta_{\min},\zeta_{\max}]$. Therefore, $\Vert g_k\Vert_{\mathcal{L}^2(0,L)}\rightarrow0$ means that the model-free correction vanishes asymptotically, and the learned energy approaches the admissible band in the $\mathcal{L}^2$ sense. Since the admissible band is assumed to contain the zero point of the model-based descent direction, the obtained result can be interpreted as a neighborhood approximation of the stationary solution.

	\section{Proof of Theorem \ref{contractive mapping}} \label{prooftheoremIL}
	\begin{proof}
		For the derivative-enhanced coordinate $\zeta_{\mathrm{d},k}=k_{\mathrm{e}}V_k+k_{\mathrm{d}}\dot V_k$ and $g(\zeta)\geq k_g\zeta-\zeta_{\mathrm{th}}$,
		\begin{equation}
			\begin{split}
				v_{\mathrm{t},k+1}
				&\leq v_{\mathrm{t},k}
				-b_3k_gk_{\mathrm{e}}V_k
				-b_3k_gk_{\mathrm{d}}\dot V_k
				+b_3\zeta_{\mathrm{th}}.
			\end{split}
			\label{lawcon0}
		\end{equation}
		On the compact active-saturation set, apply the mean-value theorem to
		$\chi(v_{\mathrm{n}},v_{\mathrm{t}})=\kappa_vv_{\mathrm{n}}^{3/2}$. Since the single-robot nominal and traversal directions are orthogonal, $r^2=v_{\mathrm{n}}^2+v_{\mathrm{t}}^2$, and direct differentiation gives
		\begin{equation}
			\begin{split}
				\gamma_1
				=\frac{\partial\chi}{\partial v_{\mathrm{t}}}
				&=-\frac{\kappa_vv_{\mathrm{n}}^{3/2}v_{\mathrm{t}}}
				{v_{\mathrm{n}}^2+v_{\mathrm{t}}^2}<0,\\
				\gamma_2
				=\frac{\partial\chi}{\partial v_{\mathrm{n}}}
				&=\kappa_vv_{\mathrm{n}}^{1/2}
				\left(\frac{3}{2}-
				\frac{v_{\mathrm{n}}^2}{v_{\mathrm{n}}^2+v_{\mathrm{t}}^2}\right)>0.
			\end{split}
			\label{eq:chi_partial_derivatives}
		\end{equation}
		The second expression is exactly (\ref{gamma6definition}), while the first is positive because $v_{\mathrm{n}}^2/(v_{\mathrm{n}}^2+v_{\mathrm{t}}^2)<1$. The affine remainder and the bounded disturbance term in (\ref{dotVsatleq}) can be collected into a bounded quantity $M_k$. Substitution into (\ref{lawcon0}) yields
		\begin{equation}
			v_{\mathrm{t},k+1}
			\leq
			\left(1+b_3k_gk_{\mathrm{d}}\lambda_{\mathrm{s}}\gamma_1\right)
			v_{\mathrm{t},k}+R_k, |R_k|\leq\bar R.
			\label{lawcon}
		\end{equation}
		The stated condition implies
		$0<1+b_3k_gk_{\mathrm{d}}\lambda_{\mathrm{s}}\gamma_1<1$.
		Hence (\ref{lawcon}) is a scalar contraction with bounded input, and
		\begin{equation}
			\limsup_{k\rightarrow\infty}|v_{\mathrm{t},k}|
			\leq
			\frac{\bar R}{1-\left|1+b_3k_gk_{\mathrm{d}}\lambda_{\mathrm{s}}\gamma_1\right|}.
		\end{equation}
		Therefore $v_{\mathrm{t},k}$ is uniformly ultimately bounded.
	\end{proof}

	\bibliographystyle{SageH}
	\bibliography{journal}

\begin{thebibliography}{72}
\providecommand{\natexlab}[1]{#1}
\providecommand{\url}[1]{\texttt{#1}}
\providecommand{\urlprefix}{URL }
\expandafter\ifx\csname urlstyle\endcsname\relax
  \providecommand{\doi}[1]{DOI:\discretionary{}{}{}#1}\else
  \providecommand{\doi}{DOI:\discretionary{}{}{}\begingroup
  \urlstyle{rm}\Url}\fi

\bibitem[{Alonso-Mora et~al.(2017)Alonso-Mora, Baker and Rus}]{alonso2017multi}
Alonso-Mora J, Baker S and Rus D (2017) Multi-robot formation control and
  object transport in dynamic environments via constrained optimization.
\newblock \emph{The International Journal of Robotics Research} 36(9):
  1000--1021.

\bibitem[{Amann et~al.(1998)Amann, Owens and Rogers}]{amann1998predictive}
Amann N, Owens DH and Rogers E (1998) Predictive optimal iterative learning
  control.
\newblock \emph{International Journal of Control} 69(2): 203--226.

\bibitem[{Ames et~al.(2019)Ames, Coogan, Egerstedt, Notomista, Sreenath and
  Tabuada}]{ames2019control}
Ames AD, Coogan S, Egerstedt M, Notomista G, Sreenath K and Tabuada P (2019)
  Control barrier functions: Theory and applications.
\newblock In: \emph{2019 18th European control conference (ECC)}. IEEE, pp.
  3420--3431.

\bibitem[{Ames et~al.(2016)Ames, Xu, Grizzle and Tabuada}]{ames2016control}
Ames AD, Xu X, Grizzle JW and Tabuada P (2016) Control barrier function based
  quadratic programs for safety critical systems.
\newblock \emph{IEEE Transactions on Automatic Control} 62(8): 3861--3876.

\bibitem[{Bae et~al.(2020)Bae, Lim and Ahn}]{bae2020distributed}
Bae YB, Lim YH and Ahn HS (2020) Distributed robust adaptive gradient
  controller in distance-based formation control with exogenous disturbance.
\newblock \emph{IEEE Transactions on Automatic Control} 66(6): 2868--2874.

\bibitem[{Bristow et~al.(2006)Bristow, Tharayil and
  Alleyne}]{bristow2006survey}
Bristow DA, Tharayil M and Alleyne AG (2006) A survey of iterative learning
  control.
\newblock \emph{IEEE Control Systems Magazine} 26(3): 96--114.

\bibitem[{Chen et~al.(2020)Chen, Mei, Li and Ma}]{chen2020distributed}
Chen L, Mei J, Li C and Ma G (2020) Distributed leader--follower affine
  formation maneuver control for high-order multiagent systems.
\newblock \emph{IEEE Transactions on Automatic Control} 65(11): 4941--4948.

\bibitem[{Crespi et~al.(2008)Crespi, Galstyan and Lerman}]{crespi2008top}
Crespi V, Galstyan A and Lerman K (2008) Top-down vs bottom-up methodologies in
  multi-agent system design.
\newblock \emph{Autonomous Robots} 24: 303--313.

\bibitem[{Debnath and Mikusinski(2005)}]{debnath2005introduction}
Debnath L and Mikusinski P (2005) \emph{Introduction to Hilbert spaces with
  applications}.
\newblock Elsevier.

\bibitem[{Elamvazhuthi et~al.(2018)Elamvazhuthi, Kuiper, Kawski and
  Berman}]{elamvazhuthi2018bilinear}
Elamvazhuthi K, Kuiper H, Kawski M and Berman S (2018) Bilinear controllability
  of a class of advection--diffusion--reaction systems.
\newblock \emph{IEEE Transactions on Automatic Control} 64(6): 2282--2297.

\bibitem[{Falanga et~al.(2018)Falanga, Foehn, Lu and
  Scaramuzza}]{falanga2018pampc}
Falanga D, Foehn P, Lu P and Scaramuzza D (2018) Pampc: Perception-aware model
  predictive control for quadrotors.
\newblock In: \emph{2018 IEEE/RSJ International Conference on Intelligent
  Robots and Systems (IROS)}. IEEE, pp. 1--8.

\bibitem[{Florence et~al.(2020)Florence, Carter and
  Tedrake}]{florence2020integrated}
Florence P, Carter J and Tedrake R (2020) Integrated perception and control at
  high speed: Evaluating collision avoidance maneuvers without maps.
\newblock In: \emph{Algorithmic Foundations of Robotics XII: Proceedings of the
  Twelfth Workshop on the Algorithmic Foundations of Robotics}. Springer, pp.
  304--319.

\bibitem[{Foehn et~al.(2021)Foehn, Romero and Scaramuzza}]{foehn2021time}
Foehn P, Romero A and Scaramuzza D (2021) Time-optimal planning for quadrotor
  waypoint flight.
\newblock \emph{Science Robotics} 6(56): eabh1221.

\bibitem[{Fox et~al.(1997)Fox, Burgard and Thrun}]{fox1997dynamic}
Fox D, Burgard W and Thrun S (1997) The dynamic window approach to collision
  avoidance.
\newblock \emph{IEEE Robotics \& Automation Magazine} 4(1): 23--33.

\bibitem[{Gao et~al.(2026)Gao, Shen and Tayebi}]{gao2026robust}
Gao S, Shen D and Tayebi A (2026) Robust adaptive learning control for a class
  of nonaffine nonlinear systems.
\newblock \emph{IEEE Transactions on Automatic Control} 71(7): 4450--4464.

\bibitem[{Gao et~al.(2025)Gao, Bai and Quan}]{gao2025distributed}
Gao Y, Bai C and Quan Q (2025) Distributed and differentiable vector field
  control within a curved virtual tube for a robotic swarm under field-of-view
  constraints.
\newblock \emph{IEEE Transactions on Automatic Control} 70(8): 5190--5205.

\bibitem[{Glotfelter et~al.(2017)Glotfelter, Cort{\'e}s and
  Egerstedt}]{glotfelter2017nonsmooth}
Glotfelter P, Cort{\'e}s J and Egerstedt M (2017) Nonsmooth barrier functions
  with applications to multi-robot systems.
\newblock \emph{IEEE Control Systems Letters} 1(2): 310--315.

\bibitem[{Gon{\c{c}}alves et~al.(2020)Gon{\c{c}}alves, Adorno, Crosnier and
  Fraisse}]{gonccalves2020stable}
Gon{\c{c}}alves VM, Adorno BV, Crosnier A and Fraisse P (2020) Stable-by-design
  kinematic control based on optimization.
\newblock \emph{IEEE Transactions on Robotics} 36(3): 644--656.

\bibitem[{Goncalves et~al.(2010)Goncalves, Pimenta, Maia, Dutra and
  Pereira}]{goncalves2010vector}
Goncalves VM, Pimenta LC, Maia CA, Dutra BC and Pereira GA (2010) Vector fields
  for robot navigation along time-varying curves in $ n $-dimensions.
\newblock \emph{IEEE Transactions on Robotics} 26(4): 647--659.

\bibitem[{Han et~al.(2021)Han, Wang, Pan, Lin, Xu and Gao}]{han2021fast}
Han Z, Wang Z, Pan N, Lin Y, Xu C and Gao F (2021) Fast-racing: An open-source
  strong baseline for ${SE}$(3) planning in autonomous drone racing.
\newblock \emph{IEEE Robotics and Automation Letters} 6(4): 8631--8638.

\bibitem[{{\.I}{\c{s}}leyen et~al.(2022){\.I}{\c{s}}leyen, van~de Wouw and
  Arslan}]{icsleyen2022low}
{\.I}{\c{s}}leyen A, van~de Wouw N and Arslan {\"O} (2022) From low to high
  order motion planners: Safe robot navigation using motion prediction and
  reference governor.
\newblock \emph{IEEE Robotics and Automation Letters} 7(4): 9715--9722.

\bibitem[{Janssens et~al.(2012)Janssens, Pipeleers and
  Swevers}]{janssens2012data}
Janssens P, Pipeleers G and Swevers J (2012) A data-driven constrained
  norm-optimal iterative learning control framework for lti systems.
\newblock \emph{IEEE Transactions on Control Systems Technology} 21(2):
  546--551.

\bibitem[{Karaman and Frazzoli(2011)}]{karaman2011sampling}
Karaman S and Frazzoli E (2011) Sampling-based algorithms for optimal motion
  planning.
\newblock \emph{The International Journal of Robotics Research} 30(7):
  846--894.

\bibitem[{Khatib(1986)}]{khatib1986real}
Khatib O (1986) Real-time obstacle avoidance for manipulators and mobile
  robots.
\newblock \emph{The International Journal of Robotics Research} 5(1): 90--98.

\bibitem[{Krishnan and Martinez(2018)}]{krishnan2018distributed}
Krishnan V and Martinez S (2018) Distributed control for spatial
  self-organization of multi-agent swarms.
\newblock \emph{SIAM Journal on Control and Optimization} 56(5): 3642--3667.

\bibitem[{Liniger et~al.(2015)Liniger, Domahidi and
  Morari}]{liniger2015optimization}
Liniger A, Domahidi A and Morari M (2015) Optimization-based autonomous racing
  of 1: 43 scale rc cars.
\newblock \emph{Optimal Control Applications and Methods} 36(5): 628--647.

\bibitem[{Lv et~al.(2025)Lv, Gao and Quan}]{lv2025high}
Lv S, Gao Y and Quan Q (2025) High-efficiency vector field by time-optimal
  spatial iterative learning.
\newblock \emph{IEEE Transactions on Robotics} 41: 5624--5644.

\bibitem[{Lv et~al.(2026)Lv, Mao, Min, Hong, Liu and Quan}]{Lv2026time}
Lv S, Mao P, Min C, Hong L, Liu Y and Quan Q (2026) Time-optimal iterative
  learning planning for lagrangian systems and its application to quadcopters.
\newblock \emph{IEEE Transactions on Automation Science and Engineering} 23:
  8556--8570.

\bibitem[{Lv et~al.(2024)Lv, Mao and Quan}]{lv2024mean}
Lv S, Mao P and Quan Q (2024) Mean-field based time-optimal spatial iterative
  learning within a virtual tube.
\newblock \emph{IEEE Control Systems Letters} 8: 2021--2026.

\bibitem[{Mao et~al.(2024)Mao, Fu and Quan}]{mao2023optimal}
Mao P, Fu R and Quan Q (2024) {Optimal virtual tube planning and control for
  swarm robotics}.
\newblock \emph{The International Journal of Robotics Research} 43(5):
  602--627.

\bibitem[{Mao et~al.(2025)Mao, Lv and Quan}]{mao2025tube}
Mao P, Lv S and Quan Q (2025) Tube {RRT}*: Efficient homotopic path planning
  for swarm robotics passing-through large-scale obstacle environments.
\newblock \emph{IEEE Robotics and Automation Letters} 10(3): 2247--2254.

\bibitem[{Mellinger and Kumar(2011)}]{mellinger2011minimum}
Mellinger D and Kumar V (2011) Minimum snap trajectory generation and control
  for quadrotors.
\newblock In: \emph{2011 IEEE International Conference on Robotics and
  Automation (ICRA)}. IEEE, pp. 2520--2525.

\bibitem[{Mellinger et~al.(2012)Mellinger, Michael and
  Kumar}]{mellinger2012trajectory}
Mellinger D, Michael N and Kumar V (2012) Trajectory generation and control for
  precise aggressive maneuvers with quadrotors.
\newblock \emph{The International Journal of Robotics Research} 31(5):
  664--674.

\bibitem[{Mueller et~al.(2015)Mueller, Hehn and
  D'Andrea}]{mueller2015computationally}
Mueller MW, Hehn M and D'Andrea R (2015) A computationally efficient motion
  primitive for quadrocopter trajectory generation.
\newblock \emph{IEEE Transactions on Robotics} 31(6): 1294--1310.

\bibitem[{Nguyen et~al.(2021)Nguyen, Kamel, Alexis and
  Siegwart}]{nguyen2021model}
Nguyen H, Kamel M, Alexis K and Siegwart R (2021) Model predictive control for
  micro aerial vehicles: A survey.
\newblock In: \emph{2021 European Control Conference (ECC)}. IEEE, pp.
  1556--1563.

\bibitem[{Orthey et~al.(2023)Orthey, Chamzas and Kavraki}]{orthey2023sampling}
Orthey A, Chamzas C and Kavraki LE (2023) Sampling-based motion planning: A
  comparative review.
\newblock \emph{Annual Review of Control, Robotics, and Autonomous Systems} 7.

\bibitem[{Owens(2016)}]{owens2016norm}
Owens DH (2016) Norm optimal iterative learning control.
\newblock \emph{Iterative Learning Control: An Optimization Paradigm} :
  233--276.

\bibitem[{Park et~al.(2025)Park, Lee, Jang and Kim}]{park2025decentralized}
Park J, Lee Y, Jang I and Kim HJ (2025) Decentralized trajectory planning for
  quadrotor swarm in cluttered environments with goal convergence guarantee.
\newblock \emph{The International Journal of Robotics Research} 44(8):
  1336--1359.

\bibitem[{Purwin and D’Andrea(2011)}]{purwin2011performing}
Purwin O and D’Andrea R (2011) Performing and extending aggressive maneuvers
  using iterative learning control.
\newblock \emph{Robotics and Autonomous Systems} 59(1): 1--11.

\bibitem[{Quan et~al.(2023)Quan, Gao and Bai}]{quan2023distributed}
Quan Q, Gao Y and Bai C (2023) Distributed control for a robotic swarm to pass
  through a curve virtual tube.
\newblock \emph{Robotics and Autonomous Systems} 162: 104368.

\bibitem[{Quan et~al.(2025)Quan, Wang and Gao}]{quan2025lifting}
Quan Q, Wang S and Gao W (2025) Lifting-wing quadcopter modeling and unified
  control.
\newblock \emph{Journal of Guidance, Control, and Dynamics} 48(3): 689--699.

\bibitem[{Rafaj{\l}owicz and Rafaj{\l}owicz(2018)}]{rafajlowicz2018iterative}
Rafaj{\l}owicz E and Rafaj{\l}owicz W (2018) Iterative learning in optimal
  control of linear dynamic processes.
\newblock \emph{International Journal of Control} 91(7): 1522--1540.

\bibitem[{Ratcliffe et~al.(2006)Ratcliffe, Lewin, Rogers, Hatonen and
  Owens}]{ratcliffe2006norm}
Ratcliffe JD, Lewin PL, Rogers E, Hatonen JJ and Owens DH (2006) Norm-optimal
  iterative learning control applied to gantry robots for automation
  applications.
\newblock \emph{IEEE Transactions on Robotics} 22(6): 1303--1307.

\bibitem[{Rezende et~al.(2021)Rezende, Goncalves and
  Pimenta}]{rezende2021constructive}
Rezende AM, Goncalves VM and Pimenta LC (2021) Constructive time-varying vector
  fields for robot navigation.
\newblock \emph{IEEE Transactions on Robotics} 38(2): 852--867.

\bibitem[{Richter et~al.(2016)Richter, Bry and Roy}]{richter2016polynomial}
Richter C, Bry A and Roy N (2016) Polynomial trajectory planning for aggressive
  quadrotor flight in dense indoor environments.
\newblock In: \emph{Robotics Research: The 16th International Symposium ISRR}.
  Springer, pp. 649--666.

\bibitem[{Romero et~al.(2022)Romero, Sun, Foehn and
  Scaramuzza}]{romero2022model}
Romero A, Sun S, Foehn P and Scaramuzza D (2022) Model predictive contouring
  control for time-optimal quadrotor flight.
\newblock \emph{IEEE Transactions on Robotics} 38(6): 3340--3356.

\bibitem[{Rousseas et~al.(2024)Rousseas, Bechlioulis and
  Kyriakopoulos}]{rousseas2024reactive}
Rousseas P, Bechlioulis C and Kyriakopoulos K (2024) Reactive optimal motion
  planning to anywhere in the presence of moving obstacles.
\newblock \emph{The International Journal of Robotics Research} 43(13):
  2027--2048.

\bibitem[{Rub{\'\i} et~al.(2020)Rub{\'\i}, P{\'e}rez and
  Morcego}]{rubi2020survey}
Rub{\'\i} B, P{\'e}rez R and Morcego B (2020) A survey of path following
  control strategies for {UAV}s focused on quadrotors.
\newblock \emph{Journal of Intelligent \& Robotic Systems} 98(2): 241--265.

\bibitem[{Safadi et~al.(2023)Safadi, Fu, Quan and
  Haddad}]{safadi2023macroscopic}
Safadi Y, Fu R, Quan Q and Haddad J (2023) Macroscopic fundamental diagrams for
  low-altitude air city transport.
\newblock \emph{Transportation Research Part C: Emerging Technologies} 152:
  104141.

\bibitem[{Singamaneni et~al.(2024)Singamaneni, Bachiller-Burgos, Manso,
  Garrell, Sanfeliu, Spalanzani and Alami}]{singamaneni2024survey}
Singamaneni PT, Bachiller-Burgos P, Manso LJ, Garrell A, Sanfeliu A, Spalanzani
  A and Alami R (2024) A survey on socially aware robot navigation: Taxonomy
  and future challenges.
\newblock \emph{The International Journal of Robotics Research} 43(10):
  1533--1572.

\bibitem[{Sinigaglia et~al.(2022)Sinigaglia, Manzoni and
  Braghin}]{sinigaglia2022density}
Sinigaglia C, Manzoni A and Braghin F (2022) Density control of large-scale
  particles swarm through pde-constrained optimization.
\newblock \emph{IEEE Transactions on Robotics} 38(6): 3530--3549.

\bibitem[{Song et~al.(2023)Song, Romero, M{\"u}ller, Koltun and
  Scaramuzza}]{song2023reaching}
Song Y, Romero A, M{\"u}ller M, Koltun V and Scaramuzza D (2023) Reaching the
  limit in autonomous racing: Optimal control versus reinforcement learning.
\newblock \emph{Science Robotics} 8(82): eadg1462.

\bibitem[{Sun and Wang(2002)}]{sun2002iterative}
Sun M and Wang D (2002) Iterative learning control with initial rectifying
  action.
\newblock \emph{Automatica} 38(7): 1177--1182.

\bibitem[{Sun et~al.(2022)Sun, Romero, Foehn, Kaufmann and
  Scaramuzza}]{sun2022comparative}
Sun S, Romero A, Foehn P, Kaufmann E and Scaramuzza D (2022) A comparative
  study of nonlinear { } and differential-flatness-based control for quadrotor
  agile flight.
\newblock \emph{IEEE Transactions on Robotics} 38(6): 3357--3373.

\bibitem[{Teissing et~al.(2024)Teissing, Novosad, Penicka and
  Saska}]{teissing2024real}
Teissing K, Novosad M, Penicka R and Saska M (2024) Real-time planning of
  minimum-time trajectories for agile uav flight.
\newblock \emph{IEEE Robotics and Automation Letters} 9(11): 10351--10358.

\bibitem[{Tordesillas and How(2021)}]{tordesillas2021mader}
Tordesillas J and How JP (2021) {MADER}: Trajectory planner in multiagent and
  dynamic environments.
\newblock \emph{IEEE Transactions on Robotics} 38(1): 463--476.

\bibitem[{Tordesillas et~al.(2021)Tordesillas, Lopez, Everett and
  How}]{tordesillas2021faster}
Tordesillas J, Lopez BT, Everett M and How JP (2021) Faster: Fast and safe
  trajectory planner for navigation in unknown environments.
\newblock \emph{IEEE Transactions on Robotics} 38(2): 922--938.

\bibitem[{Toumieh and Lambert(2022)}]{toumieh2022decentralized}
Toumieh C and Lambert A (2022) Decentralized multi-agent planning using model
  predictive control and time-aware safe corridors.
\newblock \emph{IEEE Robotics and Automation Letters} 7(4): 11110--11117.

\bibitem[{V{\'a}s{\'a}rhelyi et~al.(2018)V{\'a}s{\'a}rhelyi, Vir{\'a}gh,
  Somorjai, Nepusz, Eiben and Vicsek}]{vasarhelyi2018optimized}
V{\'a}s{\'a}rhelyi G, Vir{\'a}gh C, Somorjai G, Nepusz T, Eiben AE and Vicsek T
  (2018) Optimized flocking of autonomous drones in confined environments.
\newblock \emph{Science Robotics} 3(20): eaat3536.

\bibitem[{Webb and Van Den~Berg(2013)}]{webb2013kinodynamic}
Webb DJ and Van Den~Berg J (2013) Kinodynamic {RRT}*: Asymptotically optimal
  motion planning for robots with linear dynamics.
\newblock In: \emph{2013 IEEE International Conference on Robotics and
  Automation (ICRA)}. IEEE, pp. 5054--5061.

\bibitem[{Wiedemann et~al.(2025)Wiedemann, Scheffler, Shutin and
  Lilienthal}]{Wiedemann2025physics}
Wiedemann T, Scheffler M, Shutin D and Lilienthal AJ (2025) Physics-informed
  robotic airflow exploration and mapping with a swarm of mobile robots.
\newblock \emph{The International Journal of Robotics Research} .

\bibitem[{Williams et~al.(2017)Williams, Aldrich and
  Theodorou}]{williams2017model}
Williams G, Aldrich A and Theodorou EA (2017) Model predictive path integral
  control: From theory to parallel computation.
\newblock \emph{Journal of Guidance, Control, and Dynamics} 40(2): 344--357.

\bibitem[{Xu(2011)}]{xu2011survey}
Xu JX (2011) A survey on iterative learning control for nonlinear systems.
\newblock \emph{International Journal of Control} 84(7): 1275--1294.

\bibitem[{Xu and Huang(2008)}]{xu2008spatial}
Xu JX and Huang D (2008) Spatial periodic adaptive control for rotary machine
  systems.
\newblock \emph{IEEE Transactions on Automatic Control} 53(10): 2402--2408.

\bibitem[{Xu et~al.(2022)Xu, Cai, He, Lin and Zhang}]{xu2022fast}
Xu W, Cai Y, He D, Lin J and Zhang F (2022) Fast-lio2: Fast direct
  lidar-inertial odometry.
\newblock \emph{IEEE Transactions on Robotics} 38(4): 2053--2073.

\bibitem[{Zhang et~al.(2022)Zhang, Chu and Shu}]{zhang2022model}
Zhang Y, Chu B and Shu Z (2022) Model-free predictive optimal iterative
  learning control using reinforcement learning.
\newblock In: \emph{2022 American Control Conference (ACC)}. IEEE, pp.
  3279--3284.

\bibitem[{Zhang and Shen(2026)}]{zhang2026user}
Zhang Z and Shen D (2026) User preferences-based incompatible multiobjective
  iterative learning tracking control.
\newblock \emph{IEEE Transactions on Automatic Control} 71(1): 427--442.

\bibitem[{Zhao et~al.(2019)Zhao, Li and Ding}]{zhao2019bearing}
Zhao S, Li Z and Ding Z (2019) Bearing-only formation tracking control of
  multiagent systems.
\newblock \emph{IEEE Transactions on Automatic Control} 64(11): 4541--4554.

\bibitem[{Zheng et~al.(2021)Zheng, Han and Lin}]{zheng2021transporting}
Zheng T, Han Q and Lin H (2021) Transporting robotic swarms via mean-field
  feedback control.
\newblock \emph{IEEE Transactions on Automatic Control} 67(8): 4170--4177.

\bibitem[{Zhong et~al.(2025)Zhong, Wang and Sun}]{Zhong2026initial}
Zhong G, Wang L and Sun M (2025) An initial-rectifying-constraint optimization
  scheme for repetitive motion planning by using prescribed-time zeroing neural
  networks.
\newblock \emph{IEEE Transactions on Automation Science and Engineering} 22:
  11087--11098.

\bibitem[{Zhou et~al.(2020)Zhou, Wang, Ye, Xu and Gao}]{zhou2020ego}
Zhou X, Wang Z, Ye H, Xu C and Gao F (2020) Ego-planner: An esdf-free
  gradient-based local planner for quadrotors.
\newblock \emph{IEEE Robotics and Automation Letters} 6(2): 478--485.

\bibitem[{Zhou et~al.(2022)Zhou, Wen, Wang, Gao, Li, Wang, Yang, Lu, Cao, Xu
  et~al.}]{zhou2022swarm}
Zhou X, Wen X, Wang Z, Gao Y, Li H, Wang Q, Yang T, Lu H, Cao Y, Xu C et~al.
  (2022) Swarm of micro flying robots in the wild.
\newblock \emph{Science Robotics} 7(66): eabm5954.

\end{thebibliography}
	
\end{document}